\documentclass[10pt,twocolumn,letterpaper]{article}

\usepackage[pagenumbers]{cvpr}

\usepackage{graphicx}
\usepackage{amsmath}
\usepackage{amssymb}
\usepackage{booktabs}
\usepackage{times}
\usepackage{xcolor}
\usepackage{epsfig}
\usepackage{colortbl}
\usepackage{tabulary}
\usepackage{etoolbox}
\usepackage{multirow}
\usepackage[accsupp]{axessibility}

\usepackage[pagebackref,breaklinks,colorlinks]{hyperref}

\usepackage[capitalize]{cleveref}
\crefname{section}{Sec.}{Secs.}
\Crefname{section}{Section}{Sections}
\Crefname{table}{Table}{Tables}
\crefname{table}{Tab.}{Tabs.}

\def\cvprPaperID{4692}
\def\confName{CVPR}
\def\confYear{2022}

\begin{document}

\title{Focal Length and Object Pose Estimation via Render and Compare}
\author{
	Georgy Ponimatkin\textsuperscript{1,2}
	\quad\quad\quad
	Yann Labb\'{e}\textsuperscript{3}
	\quad\quad\quad
	Bryan Russell\textsuperscript{4}
	\\
	Mathieu Aubry\textsuperscript{1}
	\quad\quad\quad
	Josef Sivic\textsuperscript{2}
	\\
	\small{$^1$LIGM, École des Ponts, Univ Gustave Eiffel, CNRS \quad $^2$CIIRC CTU \quad $^3$ENS/Inria \quad $^4$Adobe Research}
	\\
	\small{\texttt{georgy.ponimatkin@enpc.fr}}
	\\
	\small{\url{https://ponimatkin.github.io/focalpose}}
}

\newcommand{\edit}[1]{\textcolor{black}{#1}}

\renewcommand{\paragraph}[1]{\smallskip \noindent \textbf{#1}}

\definecolor{Gray}{gray}{0.85}
\newcolumntype{a}{>{\columncolor{Gray}}c}
\newcolumntype{b}{>{\columncolor{white}}c}
\newcommand{\rulesep}{\unskip\ \vrule\ }

\maketitle

\begin{abstract}
We introduce FocalPose, a neural \textit{render-and-compare} method for jointly estimating the camera-object 6D pose and camera focal length given a single RGB input image depicting a known object.  
The contributions of this work are twofold. 
First, we derive a focal length update rule that extends an existing state-of-the-art render-and-compare  6D pose estimator to address the joint estimation task.  
Second, we investigate several different loss functions for jointly estimating the object pose and focal length. We find that a combination of direct focal length regression with a reprojection loss disentangling the contribution of translation, rotation, and focal length leads to improved results. 
We show results on three challenging benchmark datasets that depict known 3D models in uncontrolled settings. We demonstrate that our focal length and 6D pose estimates have lower error than the  existing state-of-the-art methods.

\end{abstract}

\section{Introduction}
The projection of a 3D object into an image depends not only on the object's relative pose to the camera, but also on the camera's intrinsic parameters. 
While it is possible to capture objects in a controlled environment where the camera's intrinsic parameters are known (\eg, a calibrated camera on a robot), for many ``in-the-wild" images we do not have control over the capture process and these parameters are unknown, \eg, Internet pictures or archival photographs.

\begin{figure}
    \centering
    \begin{minipage}{0.230\textwidth}
            \centering
            \includegraphics[width=\textwidth]{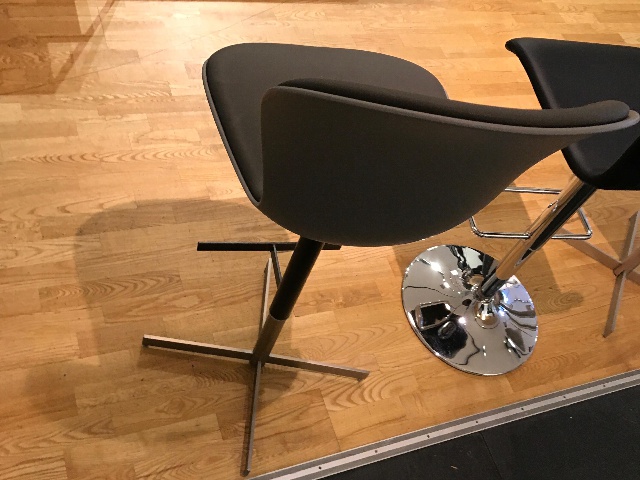}
        \end{minipage}
        \begin{minipage}{0.230\textwidth}
            \centering
            \includegraphics[width=\textwidth]{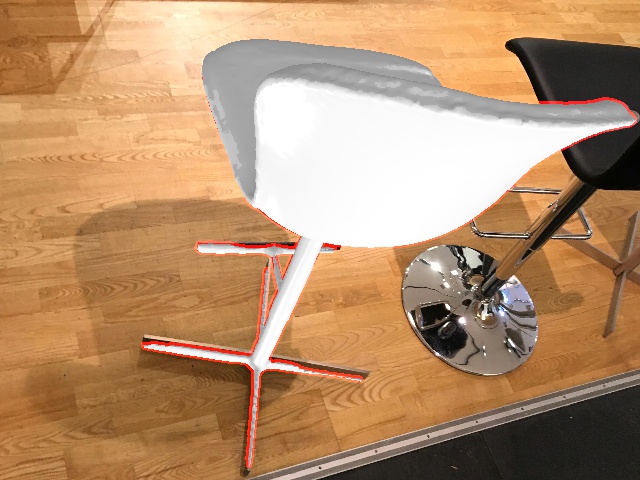}
        \end{minipage} \\[0.5mm]
        \hspace{0.125mm}
           \begin{minipage}{0.230\textwidth}
            \centering
            \includegraphics[width=\textwidth]{figures/teaser_2_input.jpeg}
        \end{minipage}
        \begin{minipage}{0.230\textwidth}
            \centering
            \includegraphics[width=\textwidth]{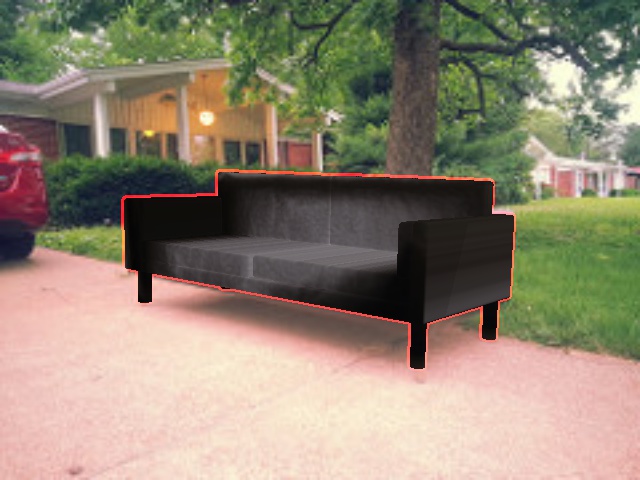}
        \end{minipage}
        \vspace{1mm}
    \caption{
        Given a single input photograph ({\bf left}) and a known 3D model, our approach accurately estimates the 6D camera-object pose together with the focal length of the camera ({\bf right}), here shown by overlaying the aligned 3D model over the input image. Our approach handles a large range of focal lengths and the resulting perspective effects. 
    }
    \vspace{-5mm}
    \label{fig:teaser}
\end{figure}

Given an input image, we seek to retrieve a 3D model of a depicted object from a model library and estimate the relative camera-object 6D pose jointly with the camera's focal length (depicted in Figure~\ref{fig:teaser}).
This problem has its origins in the early days of computer vision~\cite{Lowe1999-bf, Lowe1987-yf,Roberts1963-ck} and has important modern-day applications in augmented reality and computer graphics, such as applying in situ object overlays or editing the position of an object via 3D compositing in uncontrolled 
consumer-captured images.

The problem of 6D object pose estimation in an uncalibrated setting is, by its nature, challenging. First, it is difficult to distinguish subtle changes of the camera's focal length from changes in an object's depth. 
Second, including the camera's focal length increases the number of parameters that must be estimated and hence increases the optimization complexity. 
Finally, ``in-the-wild'' consumer-captured images may depict large appearance variation for a particular object instance in the model library. 
Variation may be due to differences in illumination and the depicted object having slightly different, non-identical shapes or surface appearance in different real-world instance captures. 
For example, consider different instances of the same car model that have a similar overall shape but may have different color, wear and tear, or customizable features (\eg, additional headlights, alloy wheels, or a spoiler). 

Previous approaches for this task primarily rely on establishing \edit{local} 2D-3D correspondences between an image and a 3D model using either hand-crafted~\cite{Aubry14,bay2006surf,Collet2010-zj,Collet2011-lj,Hinterstoisser2011-es,Lowe1999-bf} or CNN features~\cite{grabner2019gp2c,hu2019segmentation,Kehl2017-ek,Park2019-od,pavlakos20176,peng2019pvnet,Rad2017-de,song2020hybridpose,Tekin2017-hp,Tremblay2018-bd,Xiang2018-dv,zakharov2019dpod}, followed by robust camera pose estimation using PnP~\cite{lepetit2009epnp}. 
These approaches often fail on scenes with large textureless areas where local correspondences cannot be reliably established. 
In contrast, the recent best-performing 6D object pose estimation methods are based on the render-and-compare strategy~\cite{labbe2020cosypose,li2018deepim,manhardt2018deep,oberweger2019generalized,zakharov2019dpod}, which 
performs
a dense alignment \edit{over all pixels} of rendered views of the 3D model to its depiction in the input image.
However, \edit{all prior} render-and-compare methods fall short of handling the aforementioned desired uncontrolled, uncalibrated setting as they assume a controlled environment where the camera intrinsic \edit{parameters are fixed and} known {\it a priori}. \edit{Also, these prior methods typically operate}
over only a handful of known objects. 

To address these challenges, we build on the strengths of render and compare and extend it to handle our desired uncontrolled, uncalibrated setting. We introduce FocalPose, a novel render-and-compare approach for jointly estimating an object's 6D pose and camera focal length based on a monocular image input. Our contributions are twofold. 
\edit{First, we extend a recent state-of-the-art~\cite{hodan2020bop} method for 6D pose estimation (CosyPose~\cite{labbe2020cosypose}) by deriving and integrating focal length update rules in a differentiable manner, which allows our method to overcome the added complexity of including the focal length. Second, we investigate several different loss functions for jointly estimating object pose and focal length. We find that a combination of direct focal length regression with a reprojection loss disentangling the contribution of translation, rotation, and focal length leads to the best performance and allows our method to distinguish subtle differences due to the focal length and the object's depth.}
We apply our method to three real-world consumer-captured image datasets with varying camera focal lengths  and show that our focal length and 6D pose estimates have lower error compared to the state of the art.
\edit{As an added benefit, our work is the first render-and-compare method applied to a large collection of 3D meshes (20-200 meshes for Pix3D~\cite{pix3d}, $\sim150$ for the car datasets~\cite{wang20183d}).}

\section{Related Work}

\paragraph{6D pose estimation of rigid objects from RGB images.} This task is one of the oldest problems in computer vision~\cite{Roberts1963-ck,Lowe1987-yf,Lowe1999-bf} and has been successfully approached by estimating the pose from 2D-3D correspondences obtained via local invariant features~\cite{Lowe1999-bf,bay2006surf,Collet2010-zj,Collet2011-lj}, or by template-matching~\cite{Hinterstoisser2011-es}. Both of these strategies rely on shallow hand-designed image features and have been revisited with learnable deep convolutional neural networks (CNNs) \cite{Rad2017-de,Tremblay2018-bd,Kehl2017-ek,Tekin2017-hp,peng2019pvnet,pavlakos20176,hu2019segmentation,Xiang2018-dv,Park2019-od,song2020hybridpose,zakharov2019dpod}. The best-performing  methods for 6D pose estimation from RGB images are now based on variants of the deep {\em render-and-compare} strategy \cite{li2018deepim,manhardt2018deep,oberweger2019generalized,labbe2020cosypose,zakharov2019dpod}. However,  these methods assume the full perspective camera model is known so that the object can be rendered and compared with the input image. We build on the state-of-the-art render-and-compare approach of Labb\'{e} \etal~\cite{labbe2020cosypose} and extend it to the ``in-the-wild" uncontrolled set-up where the focal length of the camera is not known and has to be estimated together with the object's 6D pose directly from the input image.

\paragraph{Camera calibration.}
Camera calibration techniques ~\cite{faugeras1993three,andrew2001multiple,nakano2016versatile,penate2013exhaustive,zheng2014general,tsai1987versatile,dubska2014fully,szeliski2010computer} recover the camera model (intrinsic parameters) and its pose (extrinsic parameters) jointly. A limitation is that they require estimating 2D-3D correspondences in multiple images using structured object patterns~\cite{Forsyth12,Hartley2004,szeliski2010computer,tsai1987versatile}, identifying specific image elements such as lines or vanishing points~\cite{dubska2014fully,chen2004camera,szeliski2010computer} or structured features (\eg, human face landmarks~\cite{burgos2014distance}). These requirements limit their applicability to unconstrained images where these structures are not present.  Other works~\cite{workman2015deepfocal} have considered in-the-wild images, but only focus on recovering the focal length of the camera. In contrast, our approach recovers both components of the camera calibration (focal length and 6D camera pose) given a single image of a known object.

\begin{figure*}[th!]
  \centering
    \begin{minipage}{0.56\textwidth}
            \centering
            \includegraphics[width=\textwidth]{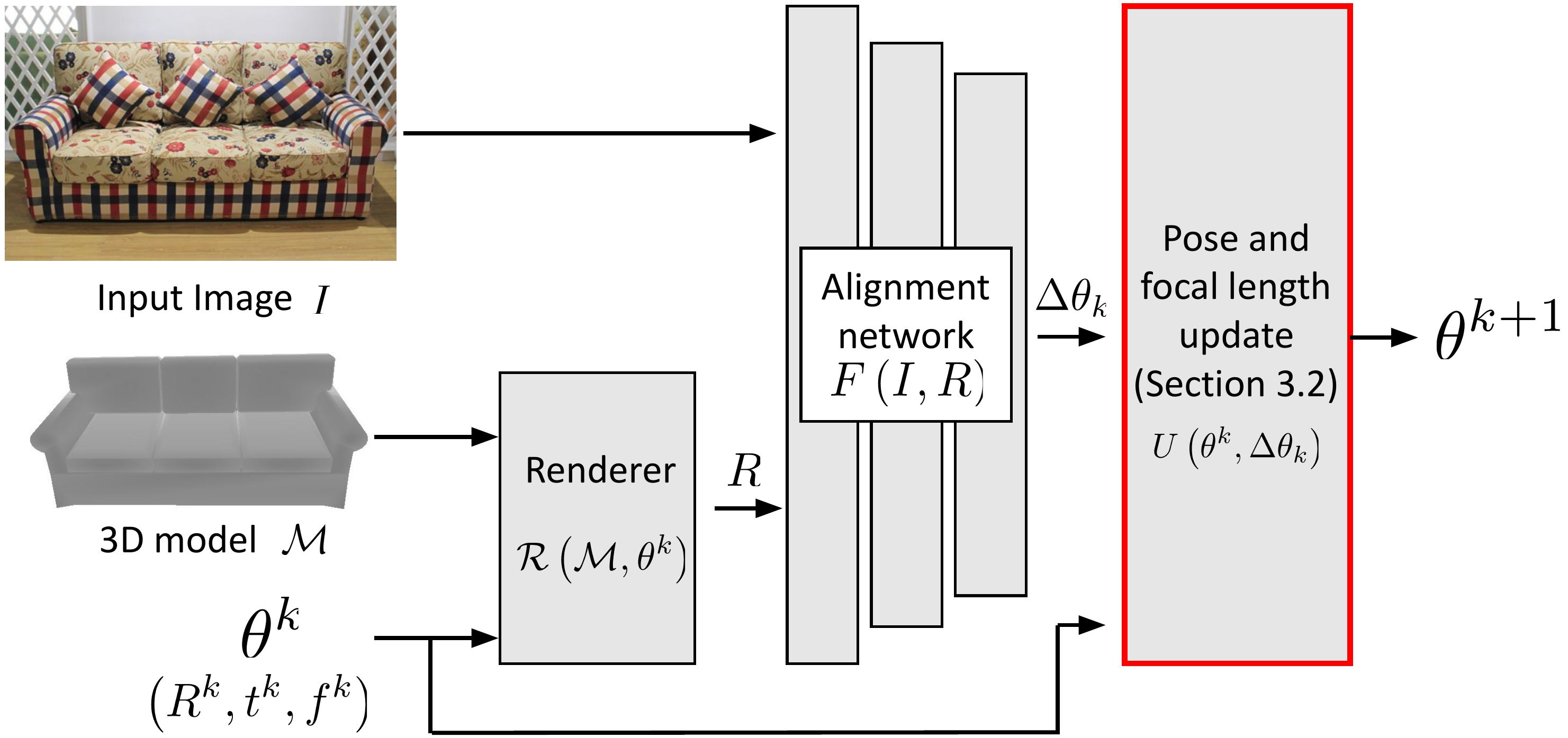}\\
    \small{(a)}
        \end{minipage}
        \rulesep
        \begin{minipage}{0.39\textwidth}
            \centering
            \includegraphics[width=\textwidth]{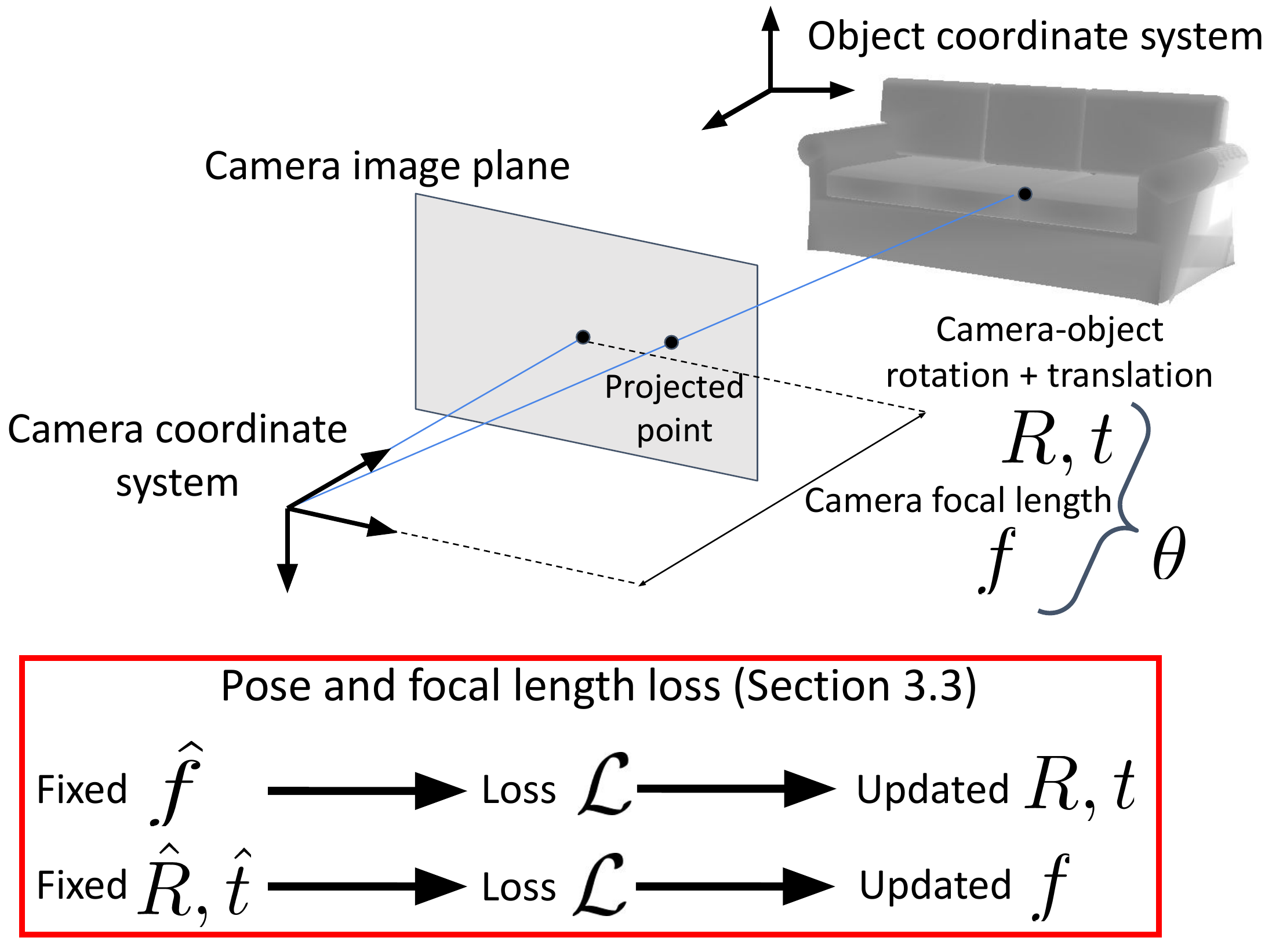}\\
    \small{(b)}
        \end{minipage} \\
  \vspace*{-3mm}
  \caption{\small {\bf FocalPose overview.} {\bf (a)}  Given a single  in-the-wild RGB input image $I$ of a known object 3D model $\mathcal{M}$, parameters $\theta^k$ composed of focal length $f^k$ and the object 6D pose (3D translation $t^k$ and 3D rotation $R^k$) are iteratively updated using our render-and-compare approach. Rendering $R$, together with the input image $I$, are given to a deep neural network $F$ that predicts update $\Delta \theta_k$, which is then converted into parameter update $\theta^{k+1}$ using a non-linear update rule $U$. {\bf (b)} Illustration of the camera-object setup with parameters $\theta$ composed of 3D translation $t$, 3D rotation $R$ and focal length $f$. The alignment network is trained using a novel pose and focal length loss that disentangles the focal length and pose updates. The two main contributions of this work are highlighted by red boxes in the figure. 
}
      
    \vspace*{-3mm}
  \label{fig:method_overview}
\end{figure*}

\paragraph{Joint 6D pose and focal length estimation from a single in-the-wild image.}
The prior work closest to our approach establishes point correspondences, followed by robust fitting of the camera model~\cite{wang20183d,grabner2019gp2c,han2020gcvnet}.
Wang \etal~\cite{wang20183d} uses Faster R-CNN with a scalar regression head and L1 loss for estimating the focal length, and the 6D pose is estimated by predicting 2D-3D correspondences followed by PnP.

GP2C~\cite{grabner2019gp2c} extends this approach via a two-step procedure that predicts initial 2D-3D correspondences and focal length with a similar direct regression, followed by applying a PnPf solver to refine jointly the 6D pose and the focal length. The model cannot be trained end-to-end as it relies on a separate non-differentiable optimizer.
GCVNet~\cite{han2020gcvnet} uses an approximation of the PnPf solver for differentiability, but its results are limited by this approximation.
In contrast, our work builds on the success of the recent render-and-compare methods~\cite{labbe2020cosypose,li2018deepim} for 6D rigid pose estimation. Our 6D pose and focal length updates are learned end-to-end using our novel focal length update parameterization coupled with a disentangled training loss. Our approach produces lower-error focal length and pose estimates compared to the two-step approach of GP2C~\cite{grabner2019gp2c} and the prior one-shot end-to-end approaches~\cite{wang20183d,han2020gcvnet}.

\section{Approach}  

Our goal is to estimate the 6D pose of objects in a photograph taken with unknown focal length.
To achieve this goal, we use a render-and-compare strategy where we estimate jointly the camera focal length with the 6D pose.
We assume knowledge of a database of 3D models that may appear in the image, but our results show that the approach is effective even if the 3D models are only approximate.

\subsection{Approach Overview}
\label{approach-intro}

The first step of our approach, illustrated in Fig.~\ref{fig:method_overview}, identifies the object location in the input image and retrieves a 3D model from the database that matches the depicted object instance. We use an object detector~\cite{he_2017_iccv} trained on real images of these known objects. At test time, we run this detector on the test image to obtain a 2D bounding box of the object and its corresponding 3D model $\mathcal{M}$.
We describe a {\em render and compare} approach, which iteratively estimates the focal length and 6D pose of the identified object. We denote the current estimate of focal length and 6D pose at iteration $k$ jointly as $\theta^k$.
The object model is first {\em rendered} using the current estimates $\theta^k$ into an image $\mathcal{R}(\mathcal{M},\theta^k)$ using a renderer $\mathcal{R}$. 
The rendering $\mathcal{R}(\mathcal{M},\theta^{k})$ and observed input image $I$  are given to a deep neural network $F$ which predicts the pose and focal length update $\Delta \theta_k$:
\begin{equation}
    \Delta\theta_k=F(I,\mathcal{R}(\mathcal{M},\theta^k )).
    \label{eq:network-predictions}
\end{equation}
The intuition is that the neural network {\em compares} the input image $I$ with the rendering $\mathcal{R}(\mathcal{M},\theta^k)$ and based on their (potentially subtle) differences predicts the update in the rendering parameters $\Delta\theta_k$. 
The pose and focal length updates $\Delta\theta_k$ are designed to be, as much as possible, free of non-linearities and thus easy to predict by the neural network $F$. The pose and focal length at the next iteration $k+1$ is then computed by a non-linear update rule $U$:
\begin{equation}
    \theta^{k+1}=U(\theta^k,\Delta\theta_k),
    \label{eq:update-rule}
\end{equation}
where $\theta^{k}$ is the current estimate of the pose and focal length, $\Delta\theta_k$ is the prediction by the network $F$ given by eq.~\eqref{eq:network-predictions}, and $\theta^{k+1}$ are the updated pose and focal length.
Note that $U$ is not learnt but derived from the 3D to 2D projection model and takes into account the nonlinearities of the imaging process.
The neural network $F$ is trained in such a way that the updated pose and focal length  $\theta^{k+1}$ are progressively closer to their ground truth. Our approach is summarized in Fig.~\ref{fig:method_overview}

\paragraph{Discussion.} Existing render-and-compare estimators~\cite{labbe2020cosypose,li2018deepim} require knowledge of the camera intrinsic parameters.
In our scenario, the problem is more challenging because the rendering also depends on the unknown focal length. 
We address this challenge by proposing an update rule for the focal length as well as a modification of the update rules for 6D pose parameters accounting for the unknown focal length (Sec~\ref{approach-inference}). We then introduce a novel loss function adapted for joint focal length and 6D pose estimation, which disentangles the  effects of the pose and focal length updates for better end-to-end training of the network (Sec.~\ref{approach-training}). Please see the \textbf{supp.\ materials} for details of our implementation, $\theta^{0}$ parameter initialization, and our training data.

\subsection{Update rules with focal length estimation}
\label{approach-inference}
The standard render-and-compare approach to 6D pose estimation~\cite{li2018deepim,labbe2020cosypose} considers only translation $t^{k}$ and rotation $R^k$ as parameters $\theta^k$. We additionally estimate the focal length $f^k$ as an unknown, and thus need to build an appropriate rule $U$ (as defined in eq.~\eqref{eq:update-rule}) for updating jointly all parameters. In detail, we assume a pinhole camera model with focal length $f_{x}^{k}=f_{y}^{k}=f^k$ in which the optical center is set at the center of the image. We define the 6D pose of the object with respect to the camera by a 3D rotation $R^k$ and a 3D translation  $t^k=[x^k,y^k,z^k]$. Next, we describe our updates for focal length and 6D pose. 

\paragraph{Focal length update.} To build an appropriate focal length update rule, we take into account the fact that it should remain strictly positive throughout the update iterations. We consider update rules that are multiplicative, \ie, they scale an initial guess $f^0$ by a sequence of multiplications. 
Let $f^{k}$ be the current estimate of the focal length at iteration $k$ and $v_{f}^{k}$ be the focal length update predicted by the network $F$ (see eq.~\eqref{eq:network-predictions}). We define the updated focal length $f^{k+1}$ as the multiplication,
\begin{equation}
\label{eq:f_update}
    f^{k+1} = e^{v_f^{k}}f^{k}.
\end{equation}
The sequence of multiplicative updates can be written as $f^{k+1} = e^{\sum_{i = 1}^{k} v_f^k}f^0$, where $f^0$ is the initial focal length and $v_{f}^i, i=0,\ldots,k-1$ are the individual updates. An alternative to the above strategy would be enforcing positivity of the focal length update via a sigmoid function instead of an exponential function. We found the exponential and sigmoid functions to behave similarly, but the sigmoid update requires setting an additional scale parameter. Hence, we opted for the simpler exponential updates as described in eq.~\eqref{eq:f_update}.

\paragraph{6D pose update.}
For the update of the 6D pose, we build on the update rule introduced in DeepIM~\cite{li2018deepim} that disentangles 3D rotation and 3D translation updates.
In more detail, the network $F$ is trained to predict a translation of the projected object center into the image $[v_{x}^{k},v_{y}^{k}]$ (measured in pixels), and a ratio $v_{z}^{k}$ of the camera-to-object depth between the observed and the rendered image. 
The 3D translation of the object is then updated from the quantities  $[v_{x}^{k},v_{y}^{k}, v_{z}^{k}]$ predicted by network $F$, taking into account the nonlinear projection equations derived from the camera model. In~\cite{li2018deepim} the focal length is known and fixed.
In our scenario the focal length is not fixed and we replace the known fixed focal length with the \textit{predicted} focal length $f^{k+1}$.
In detail, the updated 3D translation  $[x^{k+1},y^{k+1},z^{k+1}]$ of the object with respect to the camera is obtained as :
\begin{align}
\label{eq:new-rules-x}
x^{k+1} &= \left(\frac{v_x^{k}}{f^{k+1}} + \frac{x^k}{z^k}\right)z^{k+1}\\
\label{eq:new-rules-y}
y^{k+1} &= \left(\frac{v_y^{k}}{f^{k+1}} + \frac{y^k}{z^k}\right)z^{k+1}\\
\label{eq:new-rules-z}
z^{k+1} &= v_z^{k} z^k,
\end{align}
where $[v_{x}^{k},v_{y}^{k},v_{z}^{k}]$ are the object translation updates predicted by network $F$ as part of $\Delta\theta$ (eq.~\ref{eq:network-predictions}), $[x^{k},y^{k},z^{k}]$ is the 3D translation vector of the relative camera-object pose at iteration $k$, $[x^{k+1},y^{k+1},z^{k+1}]$ is the new updated 3D translation vector, and $f^{k+1}$ is the updated focal length of the camera given by eq.~\eqref{eq:f_update}.

To obtain the update of the rotation component of the object pose we use directly the prediction of the alignment network $F$ in a multiplicative update, which does not depend on the focal length. In particular, we parametrize the rotation update using two 3-vectors $v_{R,1}^{k}$, $v_{R,2}^{k}$ that define the rotation matrix $R(v_{R,1}^{k},v_{R,2}^{k})$ by Gram-Schmidt orthogonalization as described in~\cite{Zhou2018-eg}. This parametrization was found to work well for different prediction tasks~\cite{Zhou2018-eg} including 6D object pose estimation~\cite{labbe2020cosypose}. The resulting update rule is then written as 
\begin{equation}
\label{eq:R_update}
R^{k+1} = R(v_{R,1}^{k},v_{R,2}^{k}) R^{k},
\end{equation}
where $R^{k+1}$ is the new updated object rotation, $R^{k}$ is the current object rotation, and 
  $R(v_{R,1}^{k},v_{R,2}^{k})$ is the rotation matrix obtained by Gram-Schmidt orthogonalization from the two 3-vectors $v_{R,1}^{k}$, $v_{R,2}^{k}$ predicted by the alignment network $F$ as part of $\Delta\theta_k$.
Note that this rotation update is similar to the one used in DeepIM~\cite{li2018deepim}.

\subsection{Pose and focal length training loss}
\label{approach-training}
We now present our network training loss\edit{, where we assume the training data consist of image and aligned model pairs. Note that a training pair may be a real image with a manually aligned model or a rendered image of a model under a specified 6D pose and focal length.} 
Given input parameters $\theta^{k}$, 
the output parameters $\theta^{k+1}$ are fully defined by the network outputs $\Delta\theta$ given by eq.~\eqref{eq:network-predictions} and the differentiable update rules described by eqs.~\eqref{eq:f_update}-\eqref{eq:R_update} in the previous section. In the following, we consider a single network iteration and denote $\theta=\{R, t, f\}$ as the estimated parameters.
For jointly learning to estimate the 6D pose and the focal length, we use the following loss that penalizes errors in the output 6D pose predictions $(R,t)$ and the estimated focal length $f$:
\begin{equation}
\begin{split}
\mathcal{L}(\theta, \hat{\theta}) &= \mathcal{L}_{pose}((R, t), (\hat{R}, \hat{t})) \\ &+ \alpha \mathcal{L}_{focal}((R, t, f), (\hat{R}, \hat{t}, \hat{f})),
\end{split}
\label{eq:full-loss}
\end{equation}
where $\theta=\{R,t,f\}$ are the estimated pose and focal length parameters, $\hat{\theta}=\{\hat{R},\hat{t},\hat{f}\}$ are the ground truth pose and focal length parameters, $\mathcal{L}_{{pose}}$ is a loss that penalizes errors in the 6D pose estimate, $\mathcal{L}_{focal}$ is our novel loss function that jointly takes into account the errors in the focal length and the 6D predicted pose, and $\alpha$ is a scalar hyper-parameter. 
\edit{This loss is written for a single instance, but our model is trained to minimize the average loss over all training images.} 
We now describe the individual losses \edit{$\mathcal{L}_{focal}$ and $\mathcal{L}_{pose}$}.

\paragraph{Focal length loss.}
We use the following focal length loss:
\begin{equation}
    \mathcal{L}_{focal} = \beta \mathcal{L}_{H} (f, \hat{f}) + \mathcal{L}_{DR} ((R, t, f), (\hat{R}, \hat{t}, \hat{f})),
\end{equation}
where  $\mathcal{L}_{H}$ is Huber regression loss, $\mathcal{L}_{DR}$ is disentangled reprojection loss and $\beta$ is a scalar hyper-parameter. The individual terms are explained next.
The Huber regression loss $\mathcal{L}_{H}$ measures the errors between the estimated and the ground truth focal length using a logarithmic parametrization of the focal length following the recommendations from Grabner \etal~\cite{grabner2019gp2c} for better training:
\begin{equation}
\begin{split}
    \mathcal{L}_{H}(f,\hat{f}) &= ||\log(f) - \log(\hat{f})||_{H},
\end{split}
\end{equation}
where again $\hat{f}$ is the ground truth focal length and $f$ is the focal length estimated by our model.

While using only the loss $\mathcal{L}_{H}$ is possible for training our model, we found better results are obtained by also considering the 2D errors of the \edit{projected} 3D model in the image using the current estimates of the focal length and object 6D pose. We first define the reprojection error:
\begin{equation}
  \begin{split}
  \mathcal{L}_{proj.}((R, t, f), (\hat{R}, \hat{t}, \hat{f})) = \\
  \sum_{p \in \mathcal{M}}||\pi{\left(K(f), R, t, p\right)} - \pi{\left(K(\hat{f}), \hat{R}, \hat{t}, p\right)}||_1,
    \label{eq:proj}
\end{split}
\end{equation}
where $K(f)$ is the intrinsic camera matrix of our camera model with focal length $f$, $p\in\mathcal{M}$ are 3D points sampled on the object model, $\pi(K(f), R, t, p)$ is the \edit{projection} of a 3D point $p$ using the current estimates of all the parameters, and $\pi(K(\hat{f}), \hat{R}, \hat{t}, p)$ is the \edit{projection} of the same 3D point $p$ using ground truth parameters. This loss can be seen as the counterpart of the pose loss $\mathcal{L}_{pose}$ \edit{(defined below)}: instead of penalizing errors in 3D space, it penalizes reprojection errors in the image while also taking into account the estimated focal length $f$. However, this loss does not disentangle the effects of the pose and focal length predictions. We thus introduce our disentangled reprojection loss:
\begin{align}
\label{eq:DR_1}
  \mathcal{L}_{DR} &=  \frac{1}{2} \mathcal{L}_{proj}((R, t, \hat{f}), (\hat{R}, \hat{t}, \hat{f})) \\
  &+ \frac{1}{2} \mathcal{L}_{proj}((\hat{R}, \hat{t}, f), (\hat{R}, \hat{t}, \hat{f})),
   \label{eq:DR_2}
\end{align}
where each term separately measures the 2D reprojection errors resulting from errors in the 6D pose (the first term) and in the focal length (the second term). This disentanglement leads to faster convergence and better model accuracy, as we show in our ablation results. 

\paragraph{6D pose loss.} For $\mathcal{L}_{pose}$ (in equation \eqref{eq:full-loss}), we build on the loss used in CosyPose~\cite{labbe2020cosypose}. This loss is based on the point-matching loss~\cite{Xiang2018-dv,li2018deepim} that measures the error between the alignment of the points on the 3D model $\mathcal{M}$ transformed with the predicted pose $(R,t)$ and the ground truth pose $(\hat{R},\hat{t})$. CosyPose~\cite{labbe2020cosypose} extends this loss to take into account object symmetries, and uses the disentanglement ideas of~\cite{Simonelli2019-da} to separate the influence of translation errors along the camera axis, image plane, and rotations.
In our approach, we do not consider object symmetries as they are nontrivial to obtain for 3D models in the wild considered in this work.
In detail, for the pose loss
we utilize the following distance metric between two poses specified by $\{R_1, t_1\}$ and $\{R_2, t_2\}$:
\begin{equation}
    D(\{R_1, t_1\},\{R_2, t_2\}) = \frac{1}{|\mathcal{M}|} \sum_{p \in \mathcal{M}} ||(R_1p + t_1) - (R_2p - t_2)||_1,
    \label{eq:distance-pose}
\end{equation}
where $||\cdot||_1$ denotes $L_1$ norm, $R_i$ is a rotation matrix, $t_i$ is a translation vector and $p \in \mathcal{M}$ is a point sampled from the mesh $\mathcal{M}$. Following~\cite{labbe2020cosypose}, we disentangle the pose loss as
\begin{align}
    \begin{split}
        \mathcal{L}_{pose} &= D(U(\theta^k, \{v_x^k, v_y^k, \hat{v}_z^k, \hat{R}^k, \hat{v}_f^k\}), \hat{R}, \hat{t})    \\
            &+ D(U(\theta^k, \{\hat{v}_x^k, \hat{v}_y^k, v_z^k, \hat{R}^k, \hat{v}_f^k\}), \hat{R}, \hat{t}) \\
            &+ D(U(\theta^k, \{\hat{v}_x^k, \hat{v}_y^k, \hat{v}_z^k, R^k, \hat{v}_f^k\}), \hat{R}, \hat{t}),
    \end{split}
    \label{eq:disent-pose-loss}
\end{align}
where $\theta^k$ are the pose and focal length parameters at iteration $k$, $\hat{R}$ is a ground truth rotation, $\hat{t}$ is a ground truth translation, $D$ is a distance defined by Eq.~\eqref{eq:distance-pose} and $U$ is an update function defined by \eqref{eq:update-rule}. The main idea of this loss is to separate the influence of translation errors in the $x-y$ plane, depth alignment errors along the $z$ axis, and rotation errors. In Eq.~\eqref{eq:disent-pose-loss} the terms $\{v_x^k, v_y^k, \hat{v}_z^k, \hat{R}^k, v_f^k\}$, $\{\hat{v}_x^k, \hat{v}_y^k, v_z^k, \hat{R}^k, \hat{v}_f^k\}$ and $\{\hat{v}_x^k, \hat{v}_y^k, \hat{v}_z^k, R^k, \hat{v}_f^k\}$  represent the necessary updates that lead to such loss disentanglement. Here $[v_x^k, v_y^k, v_z^k]$ are translation updates at iteration $k$ as predicted by the network $F$, $R^k$ is a rotation update at iteration $k$ predicted by network $F$ and $v_f^k$ is a focal length update at iteration $k$. $\hat{v}_i^k$ and $\hat{R}^k$ then represent the updates needed to transform the current parameters to the ground truth values, which leads to the disentanglement along each of the dimensions. The first term in Eq.~\eqref{eq:disent-pose-loss} leads to the disentanglement along the $x-y$ axis, since this term provides the gradients resulting from the $x-y$ alignment errors. Analogously, the second and third terms provide gradients that arise from depth and rotation alignment errors.

\section{Experiments}
We evaluate our method for focal length and 6D pose estimation on three challenging benchmarks: the Pix3D~\cite{pix3d}, CompCars~\cite{wang20183d} and
StanfordCars~\cite{wang20183d} datasets.
In the remainder of this section we first introduce the benchmark datasets and give details of the full pose estimation pipeline.  
Then, in Sec.~\ref{sec:ablation} we present the ablation of the main components of the proposed loss function.
In Sec.~\ref{sec:sota} we compare our method with the state of the art~\cite{wang2018improving,grabner2019gp2c,han2020gcvnet} addressing the same task.
Finally, in Sec.~\ref{sec:limitations} we discuss the main limitations of our approach. 

\paragraph{Datasets \edit{and evaluation criteria}.} We consider three real-world in-the-wild datasets depicting objects with known 3D models annotated with ground truth focal length and 6D pose of the object. Following \edit{Grabner \etal}~\cite{grabner2019gp2c}, we consider the {\em bed}, {\em chair}, {\em sofa}, and {\em table} classes in the Pix3D dataset~\cite{pix3d}. The images for each object class are considered as separate datasets. The Stanford cars and CompCars datasets~\cite{wang20183d} contain images of  different car instances. Note that for the Pix3D chair images and both cars datasets, there are hundreds of different object instances in the dataset, which makes the task of recognizing the object instance challenging. 
We use the standard set of evaluation criteria used by prior work~\cite{wang2018improving,grabner2019gp2c,han2020gcvnet} that include the detection accuracy and several 6D pose metrics.
The results are reported as median errors (smaller is better) \edit{between the prediction and ground truth} (\eg, the $MedErr_R$ is the median rotation error) and accuracies (higher is better), which report the percentage of images with an error smaller than a certain threshold (\eg, $Acc_{R\frac{\pi}{6}}$ reports the percentage of test images with the rotation error smaller than $\frac{\pi}{6}$).    
\edit{See the supplemental for a detailed description of all the evaluation criteria.}

\paragraph{The complete pose estimation pipeline.}
The first step of our pipeline returns bounding box coordinates for depicted model instances in the input image via a Mask R-CNN detector. One detector is trained for each object class. For each detected instance, we crop the input image given the bounding box and apply an instance classifier to obtain which 3D model instance to align. In our case we finetune the DINO model \cite{caron2021emerging} as the instance classifier. We align the 3D model instance corresponding to the top classifier score. The classifier achieves top-1 retrieval accuracy of 62.1\% for Pix3D, 71.2\% for Stanford Cars and 79.0\% for CompCars datasets. Next, we estimate the coarse 6D pose and focal length using the full image, bounding box, and retrieved 3D model instance. Finally, the refiner FocalPose model iteratively refines the estimates for $N$ iterations given the coarse estimates.

\subsection{Loss ablation study}
\label{sec:ablation}

In this section we ablate the different components of our proposed loss function.
We train the coarse and refinement networks with the three different losses introduced in Section~\ref{approach-training}. We report the results in Table~\ref{tab:loss_ablation}. First, our solution (c.)\ combining the Huber regression loss with the 2D reprojection error taking into account the object 3D model and its 6D pose results in significantly lower errors than simply using the regression loss (a.)\ used in Grabner \etal~\cite{grabner2019gp2c}. 
Second, our new loss (c.), which disentangles the effects of focal length and pose, results in lower median errors compared to the standard reprojection loss that does not disentangle pose and focal length (b.).

\begin{table}[t]
    \centering
    \small
    \setlength{\tabcolsep}{3.2pt}
    \begin{tabular}{l|c|c|c}
    \toprule
    Loss & $MedErr_R$ & $MedErr_t \cdot 10$ & $MedErr_f \cdot 10$ \\
    \midrule
        a. $\mathcal{L}_H$ & 6.61 & 1.51 &  4.17 \\
		b. $\mathcal{L}_H + \mathcal{L}_{proj}$& 3.28 & 1.42 & 1.45 \\
		c. $\mathcal{L}_H + \mathcal{L}_{DR}$ & {\bf 2.98} & {\bf 1.29} & {\bf 1.36} \\
		\bottomrule
    \end{tabular}
    \vspace*{-2mm}
    \caption{{\bf Training loss ablation on Pix3D sofa.} The median alignment errors for refinement models trained using different loss functions. Our proposed combination of Huber regression loss with a disentangled reprojection loss (c.) performs best.  
    \vspace*{-2mm}
    }
    \label{tab:loss_ablation}
\end{table}

\subsection{Comparison to the state-of-the-art}
\label{sec:sota}
Below we report the results of our approach on the three different datasets and compare with other methods for 6D object pose and focal length estimation~\cite{wang2018improving,grabner2019gp2c,han2020gcvnet}. 

\paragraph{Pix3D dataset.}
We report the average for the four classes (bed, chair, sofa, table) in Table~\ref{table:pix3d} (top). 
The per-class results are in the {\bf supplementary material}. 
On average over all classes, our method significantly outperforms the other methods in 5 out of the 8 metrics. In particular, we see a clear improvement in the estimated focal length (almost 11\% relative reduction in the median focal length error, from 0.172 to 0.155). We see also a clear improvement in the estimated 3D translation (20\% relative reduction in the median 3D translation error, from 0.185 to 0.148). Please note that the 3D translation is related to the focal length because of the focal length/depth ambiguity. These improvements are significant and validate the contribution of our method. 

\paragraph{CompCars and Stanford cars.} A similar pattern of results is shown in Table~\ref{table:pix3d} (middle, bottom) also  for the CompCars and Stanford cars datasets that contain hundreds of different car models. Our approach obtains the best results in 4 (CompCars) and 5 (Stanford cars) of the 8 reported metrics. In particular, our method significantly improves the focal length estimates (11\% relative reduction on CompCars and 54\% relative reduction on Stanford cars) and the 3D translation estimates (10\% relative reduction on CompCars and 52\% relative reduction on StanfordCars). Again, these improvements are significant and validate the contribution of our method.

\definecolor{lightgreen}{RGB}{200,240,217}
\definecolor{lightred}{RGB}{240,200,200}
\begin{table*}[t]
	\centering
	\small
	\setlength{\tabcolsep}{3.2pt}
	\begin{tabular}{lccc||cc|a|c|a|cc}
		\toprule
		\multicolumn{3}{c}{}&\multicolumn{1}{c}{\bf Detection}&\multicolumn{2}{c}{\bf Rotation}&\multicolumn{1}{c}{\bf Translation}&\multicolumn{1}{c}{\bf Pose}&\multicolumn{1}{c}{\bf Focal}&\multicolumn{2}{c}{\bf Projection}\\

		\cmidrule(lr){4-4}\cmidrule(lr){5-6}\cmidrule(lr){7-7}\cmidrule(lr){8-8}\cmidrule(lr){9-9}\cmidrule(lr){10-11}
		\multirow{2}{*}{Method}&\multicolumn{1}{c}{\multirow{2}{*}{Dataset}}&&\multicolumn{1}{c}{\multirow{2}{*}{$Acc_{D_{0.5}}$}}&\multicolumn{1}{c}{$MedErr_R$}&\multicolumn{1}{c}{\multirow{2}{*}{$Acc_{R\frac{\pi}{6}}$}}&\multicolumn{1}{c}{$MedErr_{t}$}&\multicolumn{1}{c}{$MedErr_{R,t}$}&\multicolumn{1}{c}{$MedErr_f$}&\multicolumn{1}{c}{$MedErr_{P}$}&\multicolumn{1}{c}{\multirow{2}{*}{$Acc_{P_{0.1}}$}}\\
		&&\multicolumn{1}{c}{}&\multicolumn{1}{c}{}&\multicolumn{1}{c}{$\cdot1$}&\multicolumn{1}{c}{}&\multicolumn{1}{c}{$\cdot10^{1}$}&\multicolumn{1}{c}{$\cdot10^{1}$}&\multicolumn{1}{c}{$\cdot10^{1}$}&\multicolumn{1}{c}{$\cdot10^{2}$}\\
		\midrule
		\midrule
		\cite{wang20183d}&\multirow{5}{*}{Pix3D}&&96.0\%&7.25&87.8\%&2.52&1.76&2.41&6.33&71.5\%\\
		\cite{grabner2019gp2c}-LF&&&96.2\%&6.92&88.4\%&1.85&1.30&1.72&3.85&85.5\%\\
		\cite{grabner2019gp2c}-BB&&&\bf97.7\%&6.89&\bf{90.8\%}&1.94&1.30&1.75&3.66&\bf{88.0\%}\\
		Ours &&& 95.5\% & \textbf{4.92} & 84.1\% & \textbf{1.49} & \textbf{1.09} & \textbf{1.53} & \textbf{2.97} & 79.2\%\\
		\midrule
	    \midrule
		\cite{wang20183d}&\multirow{6}{*}{CompCars}&&\bf98.9\%&5.24&97.6\%&3.30&2.35&3.23&7.85&73.7\%\\
		\cite{grabner2019gp2c}-LF&&&98.8\%&5.23&97.9\%&2.61&1.86&2.97&4.21&95.1\%\\
		\cite{grabner2019gp2c}-BB&&&\bf98.9\%&4.87&98.1\%&2.55&1.84&2.95&3.87&\bf{95.7\%}\\
		\cite{han2020gcvnet}-TwoStep&&&-&4.37&98.1\%&3.22&1.90&3.79&4.54&90.2\%\\
		\cite{han2020gcvnet}-GCVNet&&&-&\bf{3.99}&\bf{98.4\%}&3.18&1.89&3.76&4.31&90.5\%\\
	
		Ours  &&& 98.2\% & \textbf{3.99} & 98.4\% & \textbf{2.35} & \textbf{1.67} & \textbf{2.65} & \textbf{2.95} & 93.0\%\\
		\midrule
		\midrule
		\cite{wang20183d}&\multirow{6}{*}{Stanford}&&99.6\%&5.43&98.0\%&2.33&1.80&2.34&7.46&76.4\%\\
		\cite{grabner2019gp2c}-LF&&&\bf99.6\%&5.38&\bf{98.3\%}&1.93&1.51&2.01&3.72&96.2\%\\
		\cite{grabner2019gp2c}-BB&&&\bf99.6\%&5.24&\bf{98.3\%}&1.92&1.47&2.07&3.25&\bf{96.5\%}\\
		\cite{han2020gcvnet}-TwoStep&&&-&5.09&97.5\%&2.29&1.52&2.52&3.78&93.6\%\\
		\cite{han2020gcvnet}-GCVNet&&&-&4.92&97.5\%&2.20&1.46&2.43&3.65&94.6\%\\
		Ours  &&& 99.5\% & \textbf{4.44} & 95.1\% & \textbf{1.00} & \textbf{0.84} & \textbf{1.09} & \textbf{2.55} & 93.8\%\\
		\bottomrule
	\end{tabular}
	\vspace*{-2mm}
	\caption{{\bf Comparison with the state of the art for 6D pose and focal length prediction} on the Pix3D, CompCars and Stanford cars datasets. {\bf Bold} denotes the best result among directly comparable methods.
	Our approach outperforms other competing methods in 4/5 out the 8 reported metrics on all three datasets. Clear improvements (ranging from 10\% to 50\% relative reduction in the median error) are obtained in the focal length (``Focal'') and 3D translation (``Translation'') estimates (shaded columns) on all three datasets validating our approach and demonstrating our method deals well with the focal length/depth ambiguity.
	}
	\vspace*{-2mm}
	\label{table:pix3d}
\end{table*}

\begin{figure}[t]
    \centering
    \small{a}
        \begin{minipage}{0.31\columnwidth}
            {\small Input image}
            \centering
            \includegraphics[width=\textwidth]{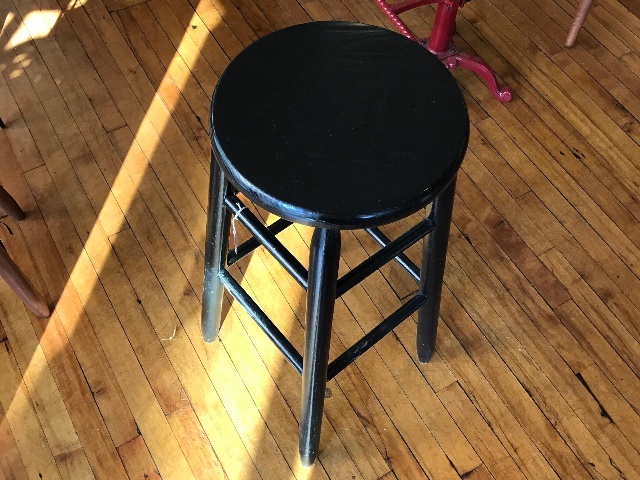}
        \end{minipage}
        \begin{minipage}{0.31\columnwidth}
            {\small Ground truth}
            \centering
            \includegraphics[width=\textwidth]{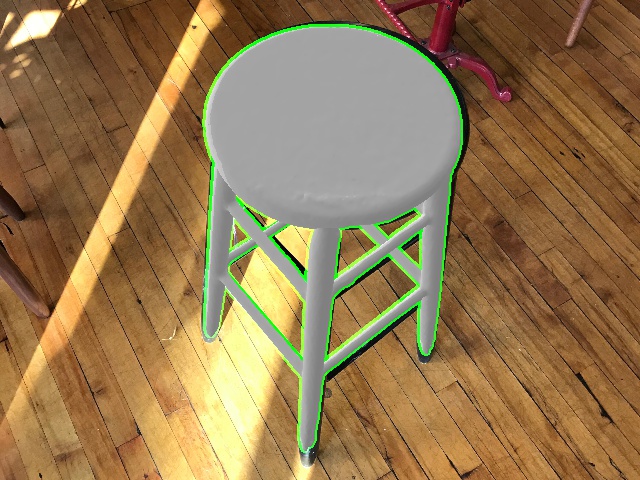}
        \end{minipage}
        \begin{minipage}{0.31\columnwidth}
        {\small Our prediction}
            \centering
            \includegraphics[width=\textwidth]{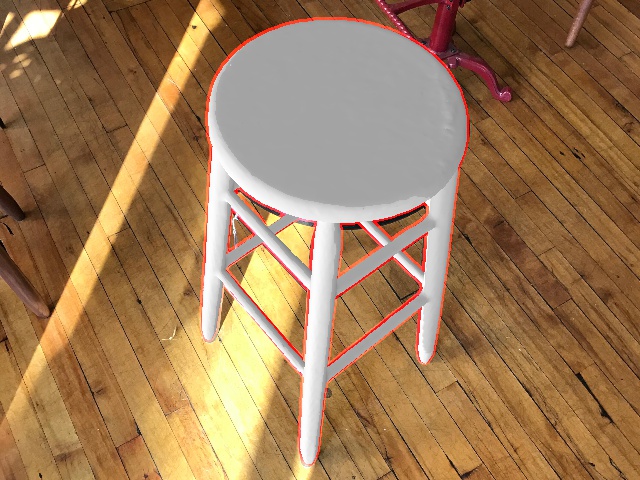}
        \end{minipage}\\[1mm]
        \hspace{0.15mm}
    \small{b}
        \begin{minipage}{0.31\columnwidth}
            \centering
            \includegraphics[width=\textwidth]{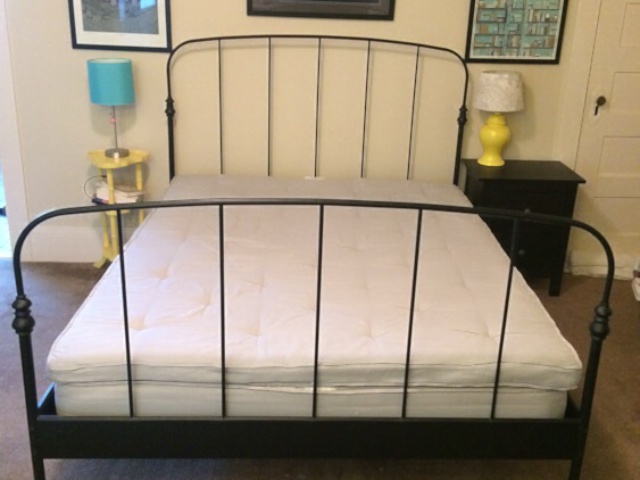}
        \end{minipage}
        \begin{minipage}{0.31\columnwidth}
            \centering
            \includegraphics[width=\textwidth]{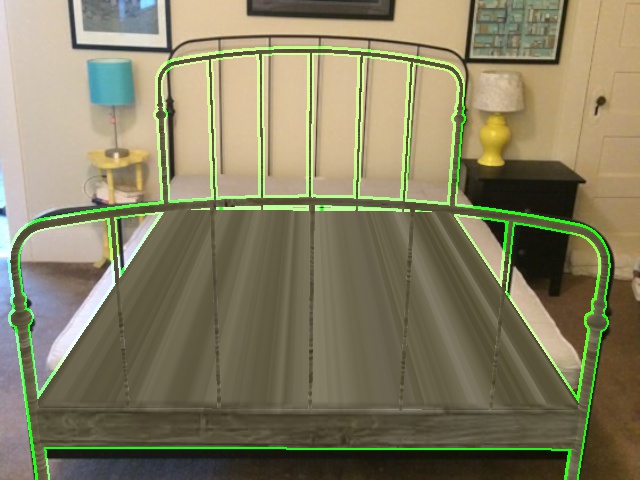}
        \end{minipage}
        \begin{minipage}{0.31\columnwidth}
            \centering
            \includegraphics[width=\textwidth]{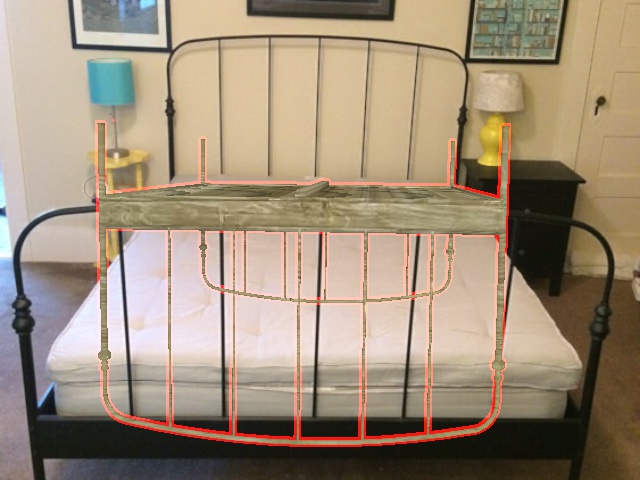}
        \end{minipage}\\[1mm]
    \small{c}
        \begin{minipage}{0.31\columnwidth}
            \centering
            \includegraphics[width=\textwidth]{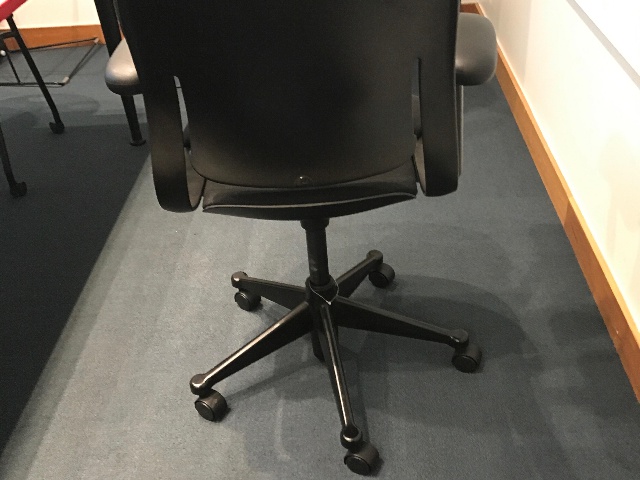}
        \end{minipage}
        \begin{minipage}{0.31\columnwidth}
            \centering
            \includegraphics[width=\textwidth]{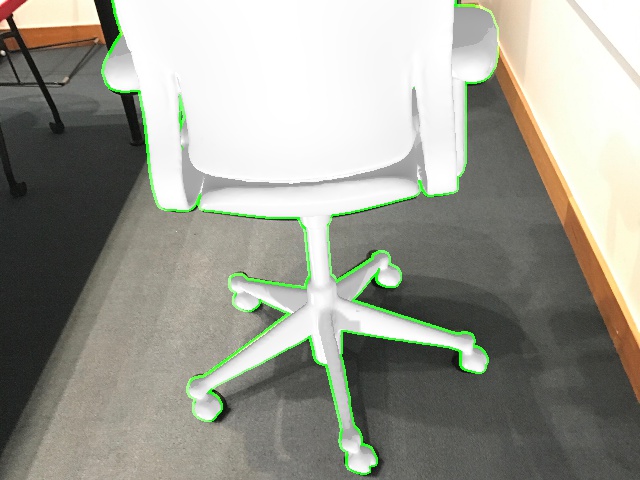}
        \end{minipage}
        \begin{minipage}{0.31\columnwidth}
            \centering
            \includegraphics[width=\textwidth]{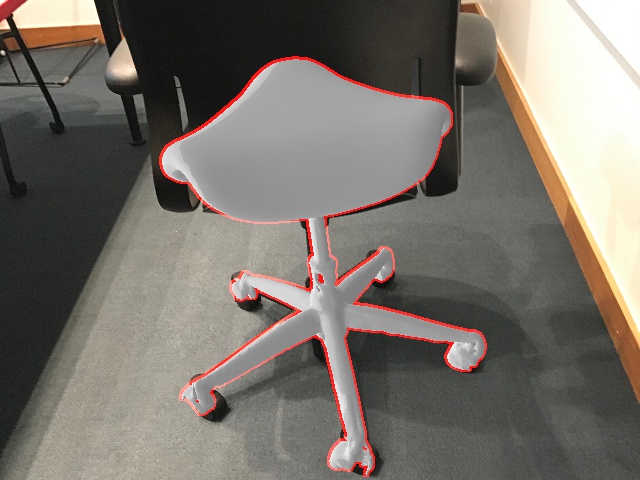}
        \end{minipage}
        \hspace{0.15mm}\\[1mm]
        \vspace*{-3mm}
        \caption{{\bf Main failure modes} are: (a) symmetric objects, (b) local minima, and (c) incorrect 3D models identified by the object detector.}
    \label{fig:failure_modes}
\end{figure}

\begin{figure}[t]
    \centering
    \small{1}
        \begin{minipage}{0.31\columnwidth}
            {\small Input image}
            \centering
            \includegraphics[width=\textwidth]{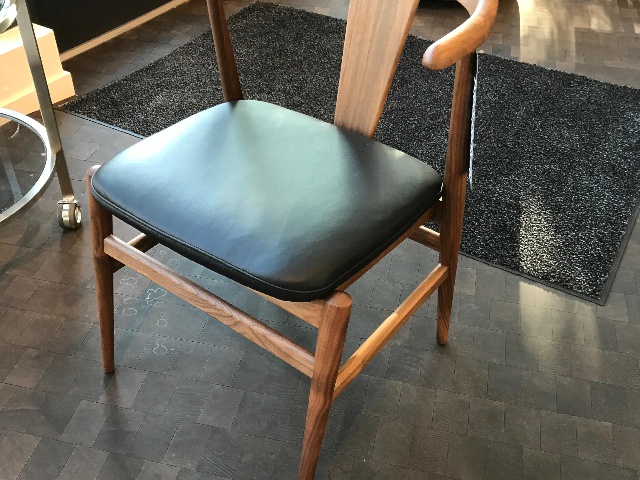}
        \end{minipage}
        \begin{minipage}{0.31\columnwidth}
            {\small Ground truth}
            \centering
            \includegraphics[width=\textwidth]{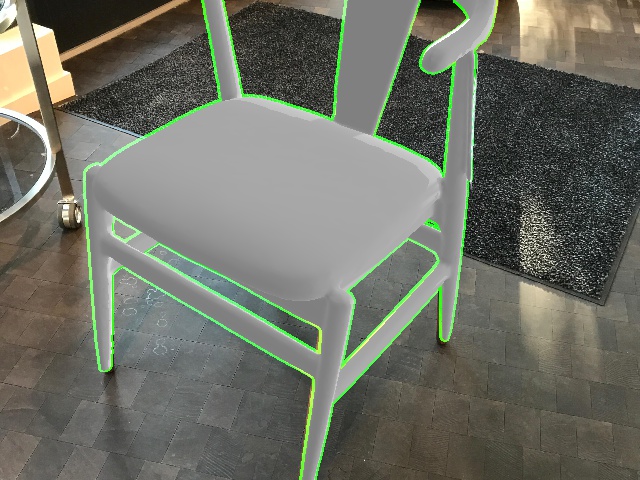}
        \end{minipage}
        \begin{minipage}{0.31\columnwidth}
            {\small Our prediction}
            \centering
            \includegraphics[width=\textwidth]{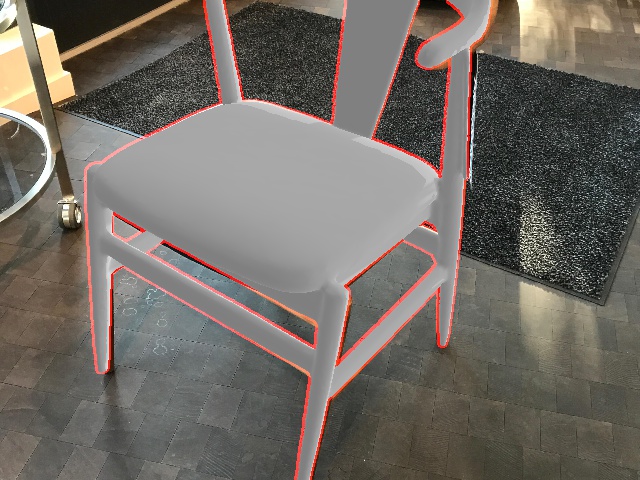}
        \end{minipage}\\[0.5mm]
    \small{2}
        \begin{minipage}{0.31\columnwidth}
            \centering
            \includegraphics[width=\textwidth]{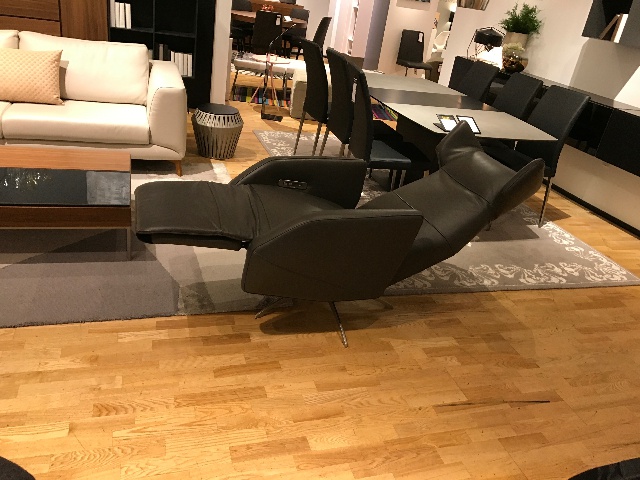}
        \end{minipage}
        \begin{minipage}{0.31\columnwidth}
            \centering
            \includegraphics[width=\textwidth]{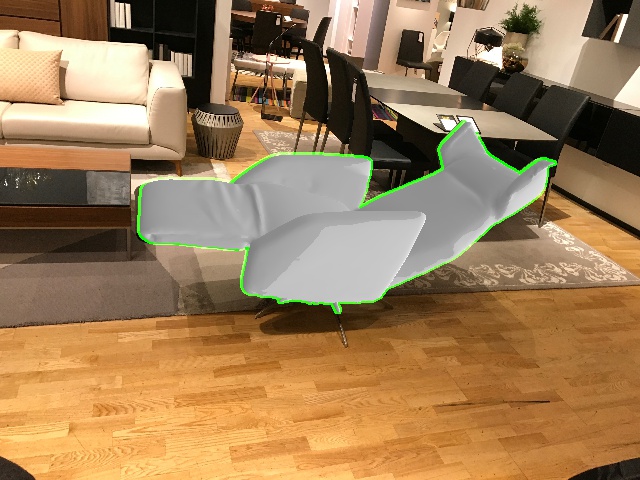}
        \end{minipage}
        \begin{minipage}{0.31\columnwidth}
            \centering
            \includegraphics[width=\textwidth]{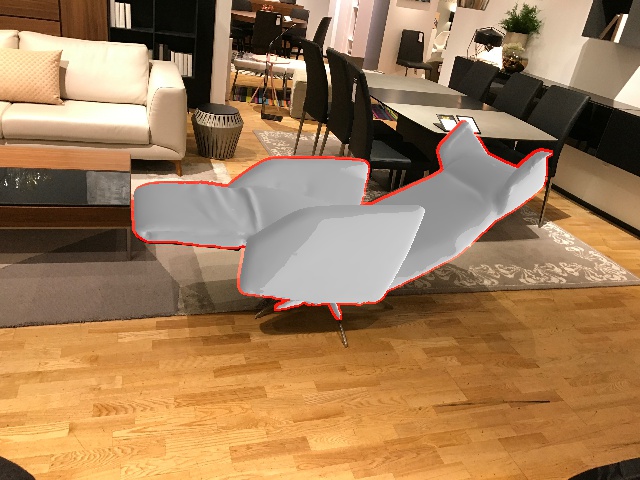}
        \end{minipage}\\[0.5mm]
    \small{3}
        \begin{minipage}{0.31\columnwidth}
            \centering
            \includegraphics[width=\textwidth]{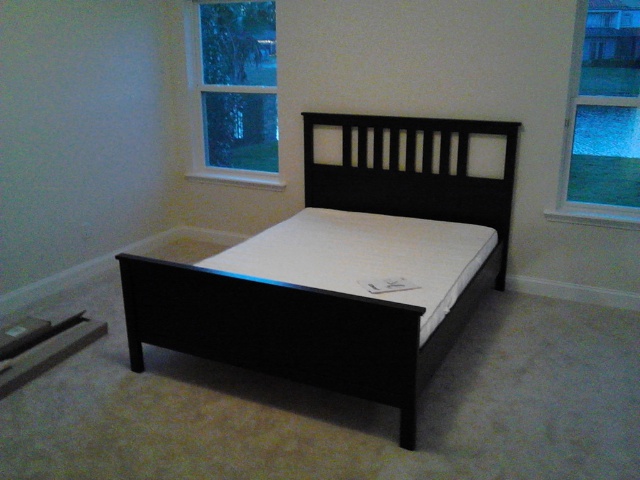}
        \end{minipage}
        \begin{minipage}{0.31\columnwidth}
            \centering
            \includegraphics[width=\textwidth]{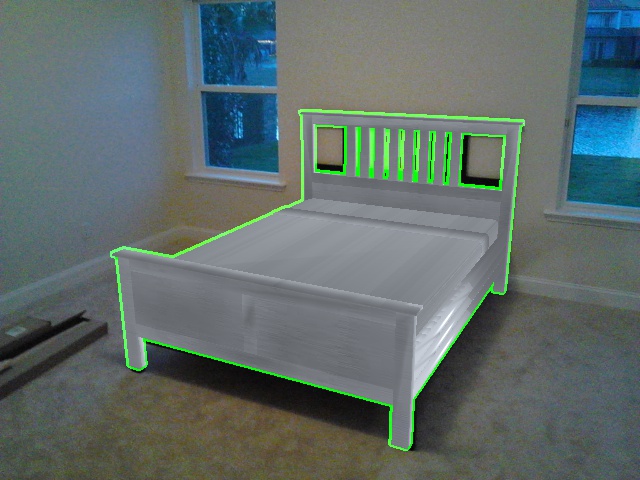}
        \end{minipage}
        \begin{minipage}{0.31\columnwidth}
            \centering
            \includegraphics[width=\textwidth]{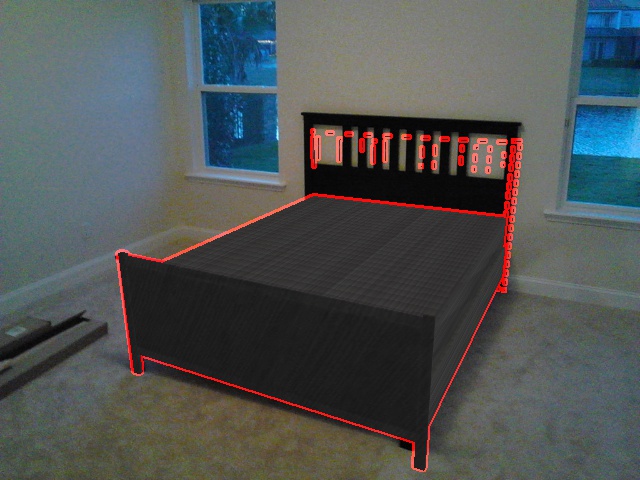}
        \end{minipage}\\[0.5mm]
    \small{4}
        \begin{minipage}{0.31\columnwidth}
            \centering
            \includegraphics[width=\textwidth]{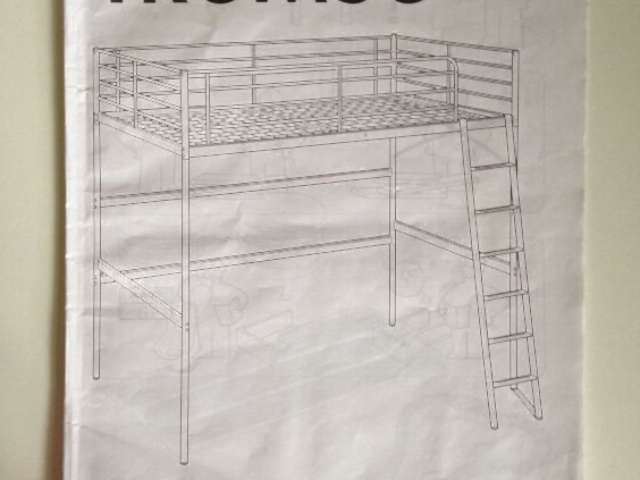}
        \end{minipage}
        \begin{minipage}{0.31\columnwidth}
            \centering
            \includegraphics[width=\textwidth]{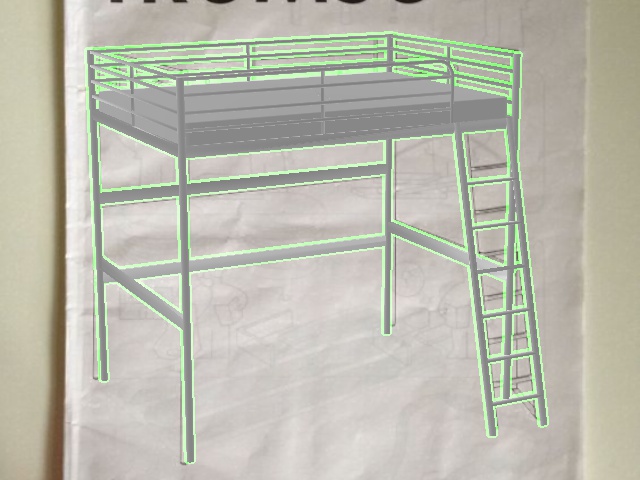}
        \end{minipage}
        \begin{minipage}{0.31\columnwidth}
            \centering
            \includegraphics[width=\textwidth]{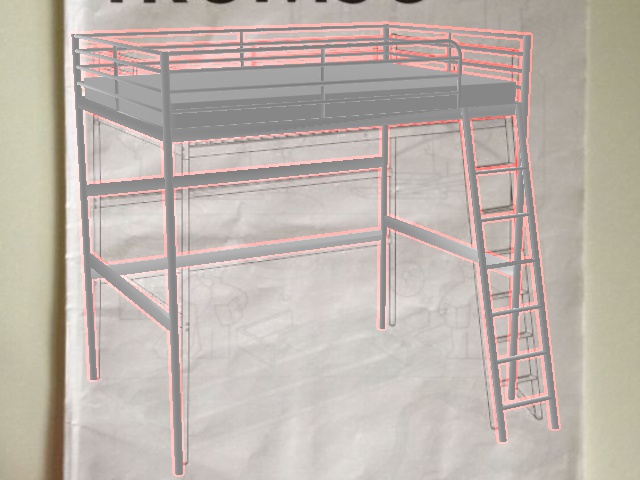}
        \end{minipage}\\[0.5mm]

    \small{5}
        \begin{minipage}{0.31\columnwidth}
            \centering
            \includegraphics[width=\textwidth]{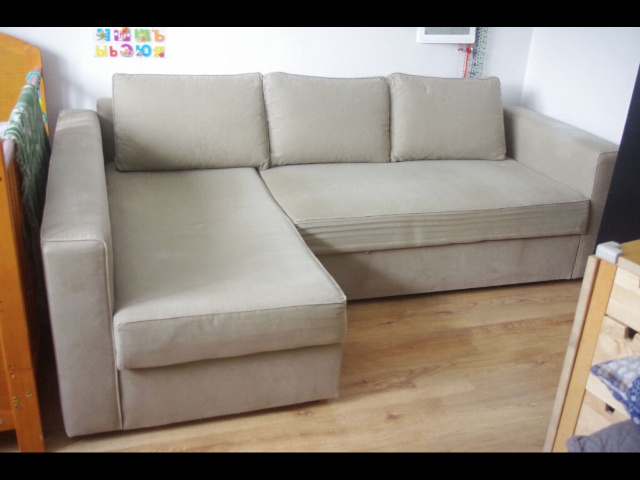}
        \end{minipage}
        \begin{minipage}{0.31\columnwidth}
            \centering
            \includegraphics[width=\textwidth]{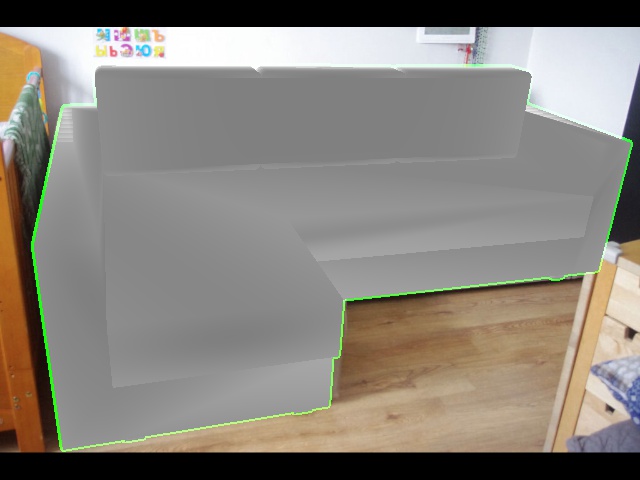}
        \end{minipage}
        \begin{minipage}{0.31\columnwidth}
            \centering
            \includegraphics[width=\textwidth]{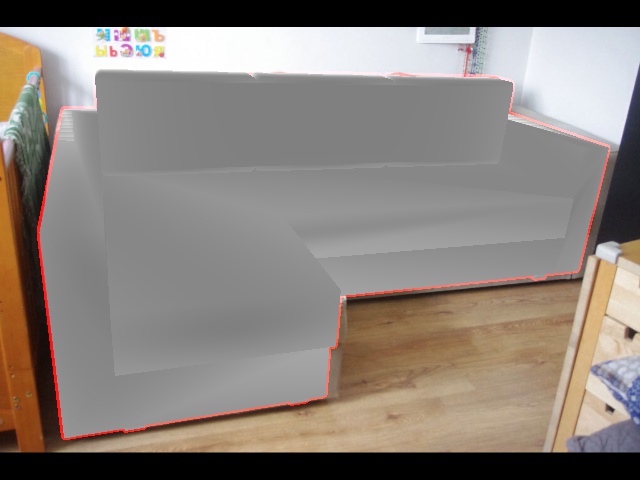}
        \end{minipage}\\[0.5mm]

    \small{6}
        \begin{minipage}{0.31\columnwidth}
            \centering
            \includegraphics[width=\textwidth]{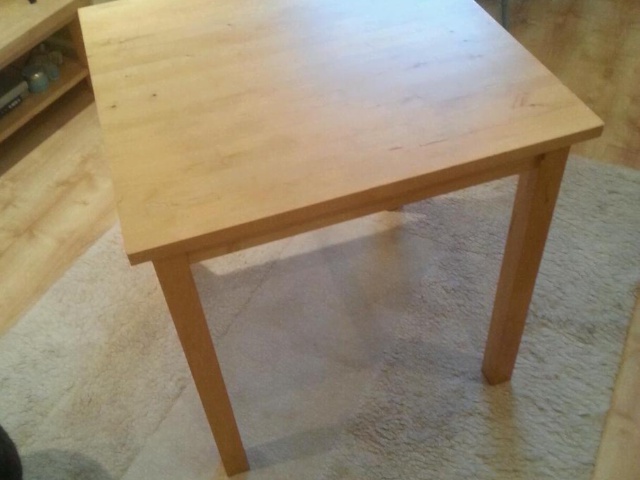}
        \end{minipage}
        \begin{minipage}{0.31\columnwidth}
            \centering
            \includegraphics[width=\textwidth]{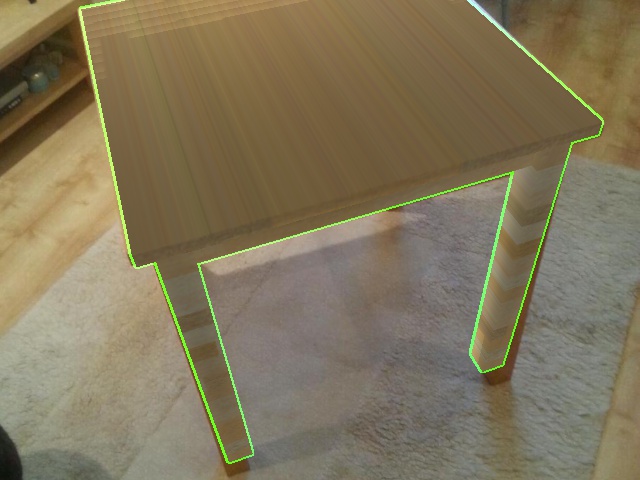}
        \end{minipage}
        \begin{minipage}{0.31\columnwidth}
            \centering
            \includegraphics[width=\textwidth]{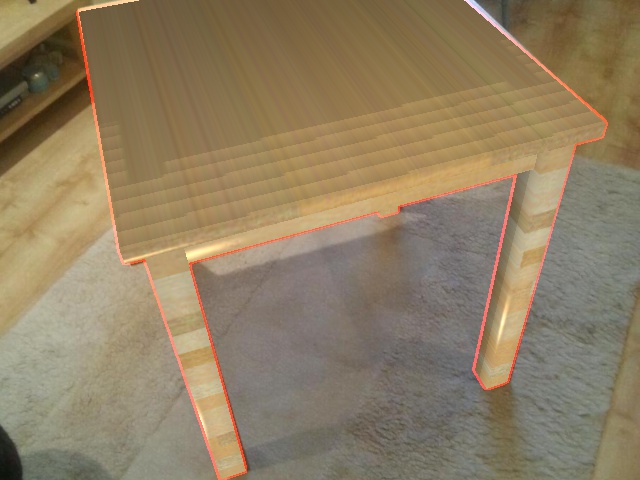}
        \end{minipage}\\[0.5mm]

    \small{7}
        \begin{minipage}{0.31\columnwidth}
            \centering
            \includegraphics[width=\textwidth]{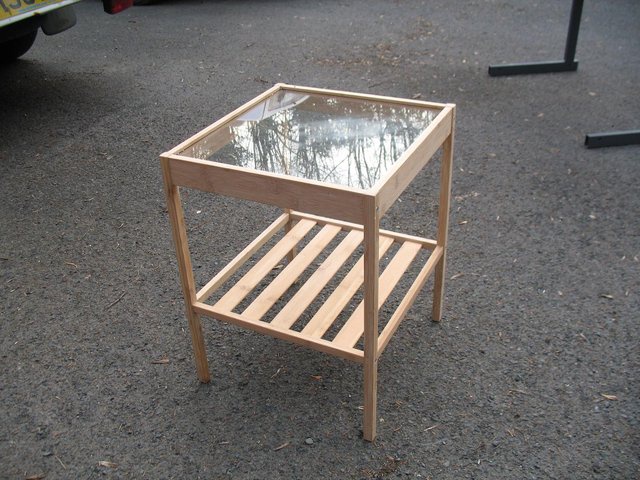}
        \end{minipage}
        \begin{minipage}{0.31\columnwidth}
            \centering
            \includegraphics[width=\textwidth]{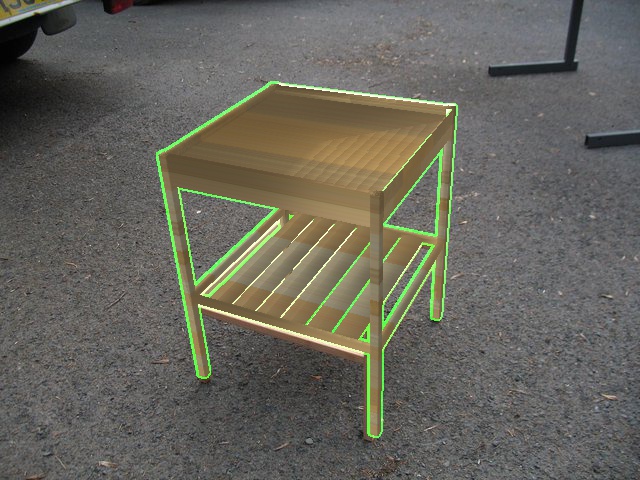}
        \end{minipage}
        \begin{minipage}{0.31\columnwidth}
            \centering
            \includegraphics[width=\textwidth]{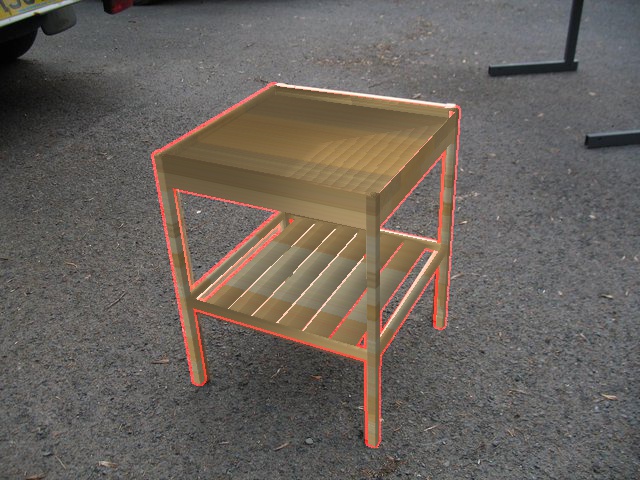}
        \end{minipage}\\[0.5mm]
        
    \small{8}
    \begin{minipage}{0.31\columnwidth}
            \centering
            \includegraphics[width=\textwidth]{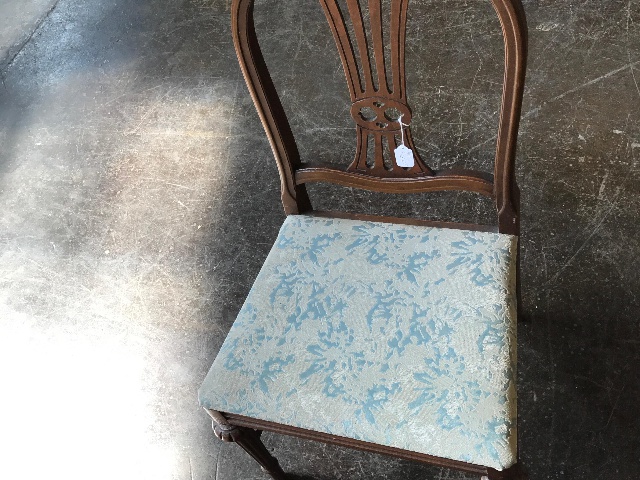}
        \end{minipage}
        \begin{minipage}{0.31\columnwidth}
            \centering
            \includegraphics[width=\textwidth]{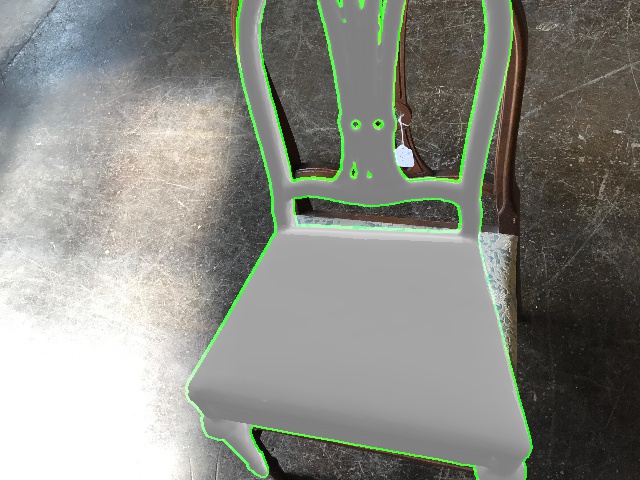}
        \end{minipage}
        \begin{minipage}{0.31\columnwidth}
            \centering
            \includegraphics[width=\textwidth]{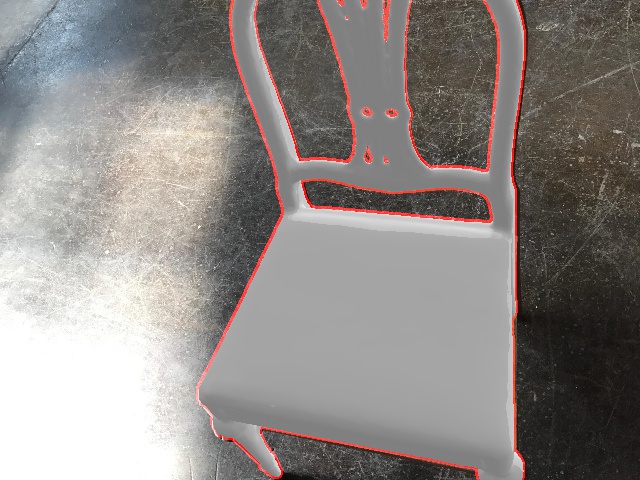}
        \end{minipage}\\[0.5mm]
    
    \small{9}
    \begin{minipage}{0.31\columnwidth}
            \centering
            \includegraphics[width=\textwidth]{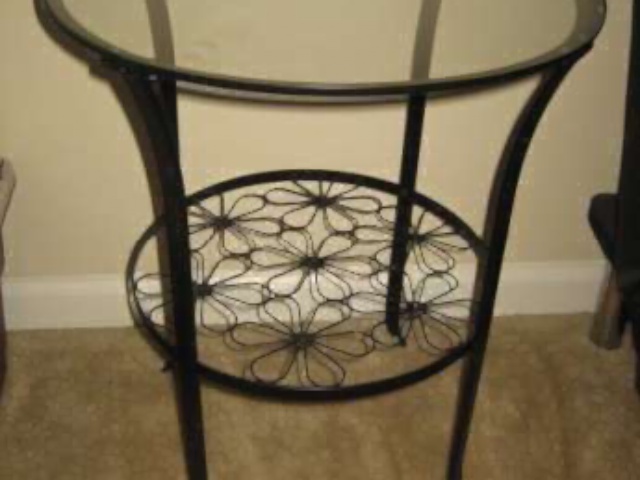}
        \end{minipage}
        \begin{minipage}{0.31\columnwidth}
            \centering
            \includegraphics[width=\textwidth]{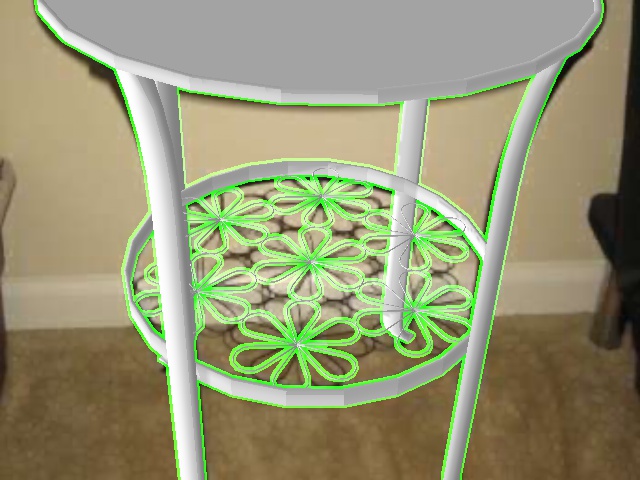}
        \end{minipage}
        \begin{minipage}{0.31\columnwidth}
            \centering
            \includegraphics[width=\textwidth]{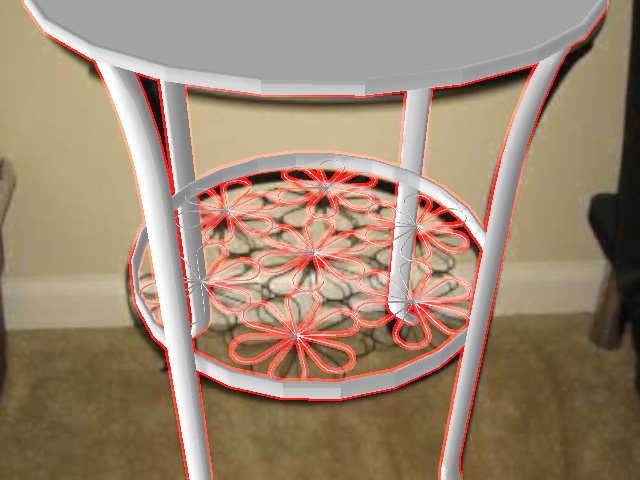}
        \end{minipage}\\[0.5mm]

    \vspace*{-2.5mm}
    \caption{{\bf Pix3D qualitative results.} 
    For each example (each row), we show the input image (left), ground truth focal length and pose annotation (center) and our prediction (right). We overlay a rendering of the detected 3D model with the jointly estimated 6D pose and focal length. Notice how our method produces precise alignments for truncated objects (rows 1, 2, 8, 9) and handles large perspective effects (rows 3, 5, 6). Notice also that in row 8 our prediction is better than the manually annotated ground truth. 
    }
    \label{fig:qual_examples}
    \vspace*{-5mm}
\end{figure}

\begin{figure}[t]
    \centering
    \small{1}
        \begin{minipage}{0.31\columnwidth}
            {\small Input image}
            \centering
            \includegraphics[width=\textwidth]{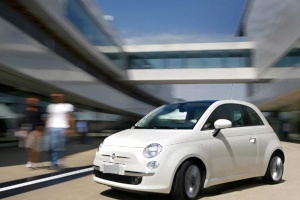}
        \end{minipage}
        \begin{minipage}{0.31\columnwidth}
            {\small Ground truth}
            \centering
            \includegraphics[width=\textwidth]{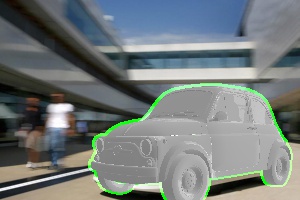}
        \end{minipage}
        \begin{minipage}{0.31\columnwidth}
            {\small Our prediction}
            \centering
            \includegraphics[width=\textwidth]{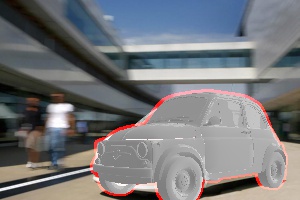}
        \end{minipage}\\[0.5mm]
    
    \small{2}
    \begin{minipage}{0.31\columnwidth}
            \centering
            \includegraphics[width=\textwidth]{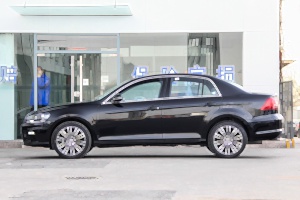}
        \end{minipage}
        \begin{minipage}{0.31\columnwidth}
            \centering
            \includegraphics[width=\textwidth]{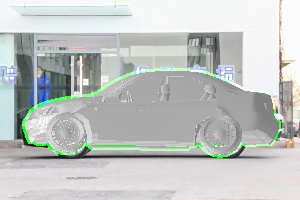}
        \end{minipage}
        \begin{minipage}{0.31\columnwidth}
            \centering
            \includegraphics[width=\textwidth]{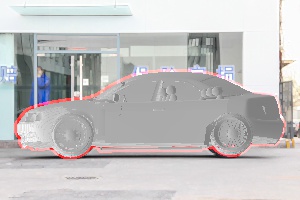}
        \end{minipage}\\[0.5mm]
    \small{3}
    \begin{minipage}{0.31\columnwidth}
            \centering
            \includegraphics[width=\textwidth]{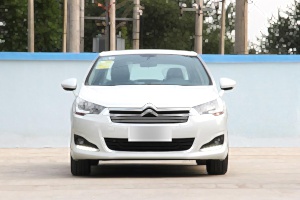}
        \end{minipage}
        \begin{minipage}{0.31\columnwidth}
            \centering
            \includegraphics[width=\textwidth]{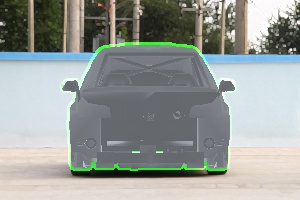}
        \end{minipage}
        \begin{minipage}{0.31\columnwidth}
            \centering
            \includegraphics[width=\textwidth]{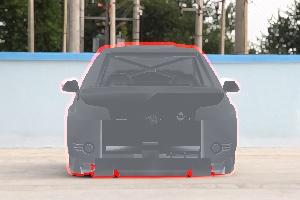}
        \end{minipage}\\[0.5mm]
    \small{4}    
    \begin{minipage}{0.31\columnwidth}
            \centering
            \includegraphics[width=\textwidth]{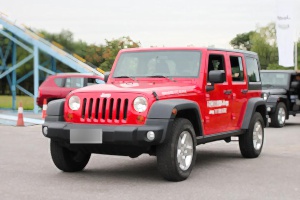}
        \end{minipage}
        \begin{minipage}{0.31\columnwidth}
            \centering
            \includegraphics[width=\textwidth]{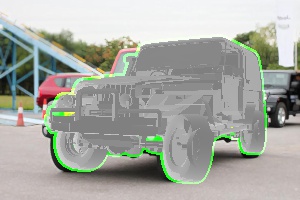}
        \end{minipage}
        \begin{minipage}{0.31\columnwidth}
            \centering
            \includegraphics[width=\textwidth]{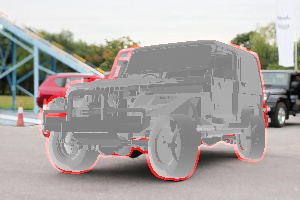}
        \end{minipage}\\[0.5mm]

    \small{5}
        \begin{minipage}{0.31\columnwidth}
            \centering
            \includegraphics[width=\textwidth]{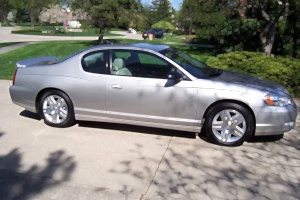}
        \end{minipage}
        \begin{minipage}{0.31\columnwidth}
            \centering
            \includegraphics[width=\textwidth]{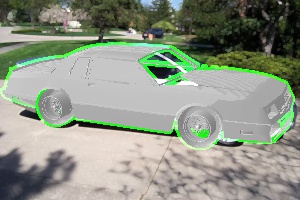}
        \end{minipage}
        \begin{minipage}{0.31\columnwidth}
            \centering
            \includegraphics[width=\textwidth]{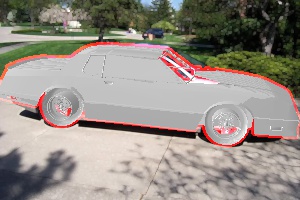}
        \end{minipage}\\[0.5mm]
     \small{6}
        \begin{minipage}{0.31\columnwidth}
            \centering
            \includegraphics[width=\textwidth]{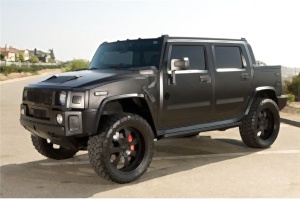}
        \end{minipage}
        \begin{minipage}{0.31\columnwidth}
            \centering
            \includegraphics[width=\textwidth]{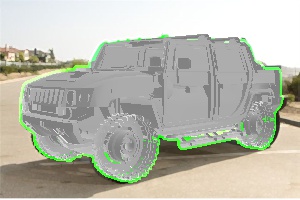}
        \end{minipage}
        \begin{minipage}{0.31\columnwidth}
            \centering
            \includegraphics[width=\textwidth]{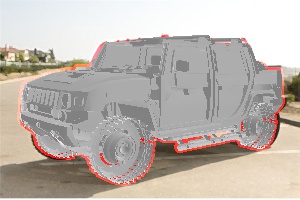}
        \end{minipage}\\[0.5mm]
    
    \small{7}
    \begin{minipage}{0.31\columnwidth}
            \centering
            \includegraphics[width=\textwidth]{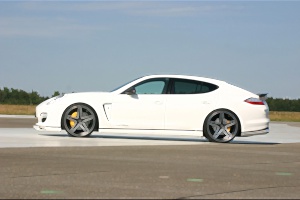}
        \end{minipage}
        \begin{minipage}{0.31\columnwidth}
            \centering
            \includegraphics[width=\textwidth]{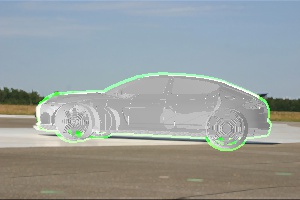}
        \end{minipage}
        \begin{minipage}{0.31\columnwidth}
            \centering
            \includegraphics[width=\textwidth]{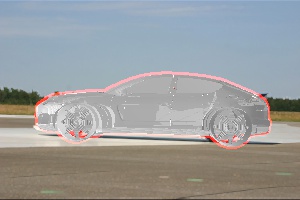}
        \end{minipage}\\[0.5mm]

    \small{8}    
    \begin{minipage}{0.31\columnwidth}
            \centering
            \includegraphics[width=\textwidth]{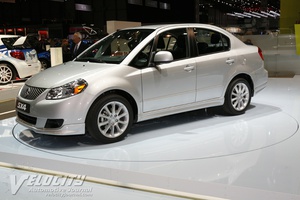}
        \end{minipage}
        \begin{minipage}{0.31\columnwidth}
            \centering
            \includegraphics[width=\textwidth]{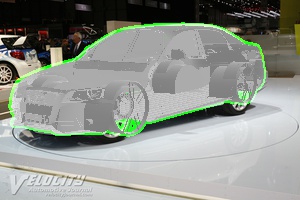}
        \end{minipage}
        \begin{minipage}{0.31\columnwidth}
            \centering
            \includegraphics[width=\textwidth]{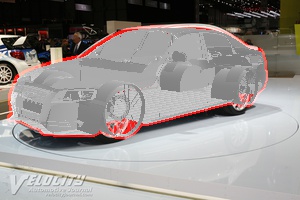}
        \end{minipage}\\[0.5mm]

    \vspace*{-2mm}
    \caption{{Example qualitative results on the {\bf CompCars} (rows 1-4) and {\bf Stanford cars} (rows 5-8) datasets.} 
    }
    \label{fig:qual_examples_stanford}
    \vspace*{-3mm}
\end{figure}

\paragraph{Qualitative results.} We report examples of qualitative results for our method on the four classes of the Pix3D dataset in Fig.~\ref{fig:qual_examples} and qualitative results on Stanford cars and CompCars datasets in Fig.~\ref{fig:qual_examples_stanford}. Please note that the renderings of the predictions (taking into account focal length and object 6D pose) show precise alignment with the observed image for in-the-wild photographs. Notably, these qualitative results demonstrate the robustness of our approach to large object truncation and strong perspective effects. 
Please see the \textbf{supplementary material} for additional qualitative results and comparisons.

\subsection{Limitations}
\label{sec:limitations}
There are three main failure modes of our approach, illustrated in Fig.~\ref{fig:failure_modes}.
First, we observe high rotation errors for \textit{symmetric objects} such as tables or stools, where the correct orientation is ambiguous. Please note that none of the used evaluation criteria
take into account the symmetries of objects. Second, our iterative alignment procedure can get stuck into a \textit{local minima} where the predicted object model in the predicted configuration is reasonably aligned but the errors are still high, \eg, because the object is flipped upside down. This \edit{failure} could be mitigated by running our approach from multiple initializations or running our refinement network on better coarse estimates. Finally, we observe that in some situations the 3D model retrieved by our pipeline is incorrect. 
These failure modes lead to large errors, which explains the lower accuracies measured by the $Acc_{R\frac{\pi}{6}}$ and $Acc_{P_{0.1}}$ metrics. 
Nevertheless, our approach achieves significantly lower median errors (5 out of the 8 reported metrics) compared to the current state-of-the-art methods, which demonstrates the high precision of our approach outside of these failure modes. 

\noindent {\bf Broader impact.} Our work has the potential to positively impact practical applications in augmented reality and robotics, among them overlaying artistic effects on viewed objects or for a robotic assistant that can manipulate real-world objects. However, our work could also potentially be used as a component for 3D-assisted manipulation of an image or video via object compositing to create misinformation.

\section{Conclusion}

We have demonstrated successful joint estimation of camera-object 6D pose and camera focal length given a single still image. Key to our success was our extension of render and compare that incorporated the estimated focal length in the iterative update rules and a disentangled loss for training. We have shown that our approach produces lower-error focal length and pose estimates compared to prior art.

Our approach can be extended to other camera intrinsic parameters besides focal length, including different forms of camera distortions, provided they can be reliably rendered.
This work opens up the possibility of downstream applications in augmented reality/computer graphics and reasoning over ``in-the-wild'' articulated and interacted objects in video.  

\paragraph{Acknowledgements.}
This work was partly supported by the European Regional Development Fund under the project IMPACT (reg. no. CZ.02.1.01/0.0/0.0/15\_003/0000468), the Ministry of Education, Youth and Sports of the Czech Republic through the e-INFRA CZ (ID:90140), the French government under management of Agence Nationale de la Recherche as part of the ``Investissements d'avenir'' program, reference ANR19-P3IA-0001 (PRAIRIE 3IA Institute).

\newpage

{\small
\bibliographystyle{ieee_fullname}
\bibliography{main}}

\newpage
\clearpage
\appendix

\section{Supplementary material -- Overview}

The supplementary material is organized as follows. In Sec.~\ref{approach-details} we provide the implementation details for our approach. 
In Sec.~\ref{sec:metrics} we give the details of the evaluation metrics used in the main paper and here.
In Sec.~\ref{per-class-pix3d} we provide the per class results for the Pix3D dataset. In Sec.~\ref{synth-data-ablation} we provide an ablation of the ratio between real data and sythetic datasets. In Sec.~\ref{refiner-iterations} we show detailed results on benefits of multiple refiner iterations. Sec.~\ref{detailed-results} provides additional quantitative evaluation of the performance of our model. Finally, in Sec.~\ref{qualitative-results} we provide more qualitative results on Pix3D, Stanford Cars, and CompCars datasets.

\subsection{Implementation details and training data}
\label{approach-details}
We base our implementation on the render and compare approach of CosyPose~\cite{labbe2020cosypose} for 6D object pose estimation. We recall the main implementation details, explain the main differences with~\cite{labbe2020cosypose}, and give the details of our training data.

\paragraph{Network architecture.} The architecture of the network $F$ (equation (1) in the main paper) relies on a ResNet-50~\cite{he2016deep} backbone, followed by average pooling and a linear layer for predicting the update $\Delta \theta$. The first input block in the backbone is inflated from 3 to 6 channels, to allow for the input of the merged RGB input image and the RGB rendered view. The (cropped) input image and rendering are resized to the input resolution: $640\times640$ for Pix3D dataset and $300\times200$ for StanfordCars and CompCars datasets.

\paragraph{Initialization.} In all experiments, we set the initial focal length $f^{0} = 600$ pixels, which we found experimentally to be a good initial value for all datasets. This focal length could also be initializatied using an EXIF file, or using a coarse estimate directly predicted by a different method.
The initialization of the 6D object pose $T^0$ follows~\cite{labbe2020cosypose} but relies on the initial focal length $f^{0}$ instead of using the ground truth focal length for computing an approximation of the object 3D translation. The initial depth of the object is set to $z = 1$ m, and the $x-y$ components of the 3D translation are derived analytically by computing the 3D position of the object center that reprojects to the center of the 2D detection, assuming the camera projection model defined by $f^{0}$. The initial object rotation is set to the identity: $R^0=I_{3}$.

\paragraph{Coarse estimate and refinement.} 
We follow CosyPose~\cite{labbe2020cosypose} and use two separate networks for coarse initialization and iterative refinement. The coarse network corrects the largest errors (between the observed state and the fixed initialization $\theta^{0}$) during the first iteration $k=1$. A separate refinement network iteratively refines the estimates by correcting smaller errors. The refinement network runs for multiple iterations, we run $K$ iterations of the refinement network at test time in our experiments, with $K=15$ on Pix3D and $K=55$ on the Stanford cars/CompCars datasets in our experiments.

\paragraph{Training input error distribution.} We use the same network architecture defined above for the coarse and refinement networks, but both are trained with different error distributions to simulate what each network is going to see at test time. During training, the initialization of the coarse network is the same as the one used at test time and described in the previous paragraph. Simulating the error distribution of the refinement network is more complicated as its input is not fixed and depends on the coarse estimate. To simulate the errors in focal length which the refinement network will see, we sample the focal length $f^{k}$ from a gaussian distribution centered on the ground truth $f^{gt}$, with variance $0.15 f^{gt}$. The error of the input pose given to the refiner is sampled from  
a Gaussian with standard deviation of $1cm$ around the $x-y$ components of translation, $5$ cm for the depth, and noise is added to the ground truth rotation matrix using three Euler angles sampled from Gaussian distributions with variance of $15^\circ$.

\paragraph{Training data.} For training our coarse and refinement networks, we use the same training images. They consist of both real training images (of the Pix3D, CompCars/Stanford Cars datasets) and one million synthetic images that are generated for each dataset using the following procedure. For each image, we sample a random object instance, sample its rotation uniformly in the quaternion space, and sample its 3D position within a box of 15 cm size. We add random textures to the object and to the background. The camera-to-object distance is sampled within the interval (0.8, 3.0) meters for the Stanford/Comp cars datasets, and (0.8, 2.4) meters for Pix3D. The focal length is sampled within (200, 1000) pixels, which covers the range of focal lengths from all datasets. While sampling each minibatch during training, one of the real images is sampled with probability $0.5\%$ while the synthetic images are sampled with probability $99.5\%$ and account for most images in each minibatch. Following~\cite{labbe2020cosypose}, we also use data augmentation to increase the number of training images. Data augmentation includes adding blur, contrast, brightness, color, and sharpness image filters to the image, and replacing the background with an image from the Pascal VOC dataset with probability 0.3.

\paragraph{Training procedure.} The coarse and refinement networks are initialized using classification network pretrained on ImageNet, and are trained using the same procedure as in~\cite{labbe2020cosypose}. Training is performed on 40 NVIDIA A100 GPUs using a global batch size of 1280. The average training time for one coarse/refiner model is around 5 hours. Each network is trained for $10$M iterations using the Adam optimizer~\cite{DBLP:journals/corr/KingmaB14} with a learning rate of $3\times 10^{-4}$. We use a linear warmup of the learning rate during the first $700$K iterations and decrease it to $3\times 10^{-5}$ after $7$M iterations. During inference, the network can process 32 $640\times 640$ pixel resolution images in approximately 10 seconds. This time includes coarse estimation and 15 refiner iterations.

\paragraph{2D detection and instance-recognition.} We use Mask R-CNN~\cite{he_2017_iccv} for predicting a 2D bounding box of the object of interest and identifying the object instance that is rendered during the alignment. 

The Mask R-CNN is based on a ResNet-50~\cite{he2016deep} feature pyramid (FPN) backbone~\cite{lin2017feature}. The network is initialized from a network trained on MS COCO, and the first ten convolutional layers remain fixed during training. This detector is trained using only the data provided by the Pix3D and Stanford/Comp cars datasets.

\paragraph{Cropping strategy.} The images from the datasets are center cropped to $640 \times 640$px for Pix3D and $300 \times 200$px for Stanford cars and CompCars. The input image is padded to conserve the input aspect ratio.

The second cropping happens before the input to the network itself. Let us call $(x_c, y_c)$ the 2D coordinates resulting from the projection of the 3D object center by the camera with intrinsic parameter matrix $K$ and $[x_1, y_1, x_2, y_2]$ the coordinates of the bounding box provided by external means (for example, the Mask R-CNN detector), where $x_1$ is the lower-left coordinate, $x_2$ is the lower-right coordinate, $y_1$ is the upper-left coordinate and $y_2$ is the upper-right coordinate of the provided bounding box. Then we define 
\begin{equation}
    x_{dist} = \max(|x_1 - x_c|, |x_2 - x_c|),
\end{equation}
\begin{equation}
    y_{dist} = \max(|y_1 - y_c|, |y_2 - y_c|).
\end{equation}
Then, the cropped image width and height are given by
\begin{equation}
    w = \max(x_{dist}, y_{dist}/r)\cdot 2\lambda,
\end{equation}
\begin{equation}
    h = \max(x_{dist}/r, y_{dist})\cdot 2\lambda,
\end{equation}
where $r$ is the aspect ratio of the input image and $\lambda = 1.4$ is a parameter controlling the enlargement of the input image to capture the whole object. This value was chosen following~\cite{li2018deepim}.

\paragraph{Loss weights.} We utilize $\alpha = 10^{-2}$ and $\beta = 1$ as weights for the losses given by equations (8) and (9) in the main paper.

\subsection{Evaluation criteria}
\label{sec:metrics}

We now recall the metrics presented in~\cite{wang20183d}, commonly~\cite{wang20183d,grabner2019gp2c,han2020gcvnet} used on these datasets and also used in this work. 

\paragraph{Detection metric.} We report the detection accuracy $Acc_{D_{0.5}}$ which corresponds to the percentage of images for which the intersection over union between the ground truth and predicted 2D bounding box is larger than $0.5$. 

Note that an incorrect object prediction is not penalized by this metric as our method can predict the focal length and object 6D pose even if the model is only approximate as long as it belongs to the correct category for which the 3D models are approximately aligned, similar to~\cite{wang20183d,grabner2019gp2c,han2020gcvnet}.

\paragraph{6D pose metrics.} We report the point matching error $e_{R,t}$ that measures the error between the 3D points of the object model transformed with the ground truth and with the estimated 6D pose with respect to the camera:
\begin{equation}
\label{eq:errorRt}
  e_{R,t} = \frac{d_{bbox}}{d_{img}} \underset{p \in \mathcal{M}^{\star}}{\mathrm{avg}} \frac{||(R p + t) -  (\hat{R} p + \hat{t})||_{2}}{||\hat{t}||_{2}},
\end{equation}
where $d_{{bbox}}$ is the diagonal of the ground truth 2D bounding box, $d_{img}$ is the diagonal of the image, $\mathcal{M}^{\star}$ is the 3D model of the ground truth object instance, $(R, t)$ is the predicted 6D pose and $(\hat{R}, \hat{t})$ is the ground truth 6D pose. Note that the point error in 3D (the numerator of~\eqref{eq:errorRt})  is normalized by the ground truth object-to-camera distance $||\hat{t}||_{2}$ and multiplied by the relative size of the object in the image $\frac{d_{bbox}}{d_{img}}$~\cite{grabner2019gp2c}.

Following~\cite{grabner2019gp2c}, we also use metrics that evaluate separately the quality of the estimated 3D translation and rotation. We use the rotation error computed using the geometric distance between the predicted rotation $R$ and the ground truth rotation $\hat{R}$ $e_{R}=\frac{||\mathrm{log}(\hat{R}^T R)||_{F}}{\sqrt{2}}$, and the normalized translation error
$e_{t}=\frac{||t - \hat{t}||_{2}}{||\hat{t}||_{2}}$, where $t$ is the predicted translation and $\hat{t}$ is the ground truth translation. For all the errors, we report the median value (denoted as $MedErr_{Rt}$, $MedErr_{R}$, $MedErr_{t}$, respectively). Following~\cite{grabner2019gp2c}, for the rotation error we also report the percentage of images with $e_{R} \leq \frac{\pi}{6}$ denoted as $Acc_{R\frac{\pi}{6}}$.

\paragraph{Focal length and reprojection metrics.} Following~\cite{grabner2019gp2c}, we report the relative focal length error $e_{f}=\frac{|\hat{f} - f|}{\hat{f}}$ between the estimated focal length $f$ and the ground truth focal length $\hat{f}$. We also report the reprojection error $e_{P}$ which is similar to the error of 6D pose (eq.~\eqref{eq:errorRt} but reprojects the 3D points into the image, also taking into account the estimated focal length $f$:
\begin{equation}
  e_{P} = \underset{p \in \mathcal{M}^{\star}}{\mathrm{avg}} \frac{||\pi(R, t, f, p) - \pi(\hat{R}, \hat{t}, \hat{f}, p)||_{2}}{d_{bbox}},
\end{equation}
where $p$ are the 3D points of the object model $\mathcal{M}^{\star}$ of the ground truth object instance, $\pi(K(f), R, t, p)$ is the reprojection of a 3D point $p$ using the estimated parameters, and $\pi(K(\hat{f}), \hat{R}, \hat{t}, p)$ is the reprojection of the same 3D point $p$ using ground truth parameters, and $d_{bbox}$ is the diagonal of the ground truth 2D bounding box. We report the median value of the reprojection error $MedErr_{P}$ and the percentage of images where the reprojection error is below $0.1$ of the image, $Acc_{P_{0.1}}$

\subsection{Per class results on the Pix3D dataset} 
\label{per-class-pix3d}
In Tab.~\ref{table:pix3d-perclass} we show the performance of our FocalPose approach on individual Pix3D classes. 
For \textbf{bed}, \textbf{chair} and \textbf{sofa} our algorithm clearly outperforms the prior methods on the five out of eight reported metrics. In particular, we see a clear improvement in the estimated focal length and 3D translation, which validates the contribution of our work. For tables, our approach improves only two out of the eight metrics. We believe this could be attributed to the fact that tables are often symmetric, which makes the 6D object pose estimation approach hard and often ambiguous, as discussed in the main paper. Object symmetries are one of the main failure models of our approach. The overall difficulty of the \textbf{table} class is clearly visible from the significantly worse results for all the tested methods on this class.

\subsection{Training data ablation}
\label{synth-data-ablation}
Manually annotating real in-the-wild images~\cite{wang20183d,wang2018improving} with the focal length and 6D pose is difficult because it requires significant effort and the ambiguities can be hard to resolve.

This setting results in relatively few available training images. Moreover, the annotations are often of poor quality as has been also discussed in Sec.~\ref{sec:sota} and illustrated in Fig.~\ref{fig:qual_examples} (row 8) in the main paper. Using synthetic data allows generating many images with accurate annotations. 

In Tab.~\ref{tab:data_ablation}, we report the results of our coarse model trained with only real data, only synthetic data, or a mix of synthetic and real data in each mini-batch (the fraction of real data in the mixed-data mini-batch is indicated in the table row). Using (exact) synthetic data in addition to a small number of (human-labeled) real images in each mini-batch yields the lowest median error.

\subsection{Multiple refiner iterations}
\label{refiner-iterations}
\begin{figure*}[t]
    \centering
    \small{1}
        \begin{minipage}{0.675\columnwidth}
            \textbf{Focal length error}
            \centering
            \includegraphics[width=\textwidth]{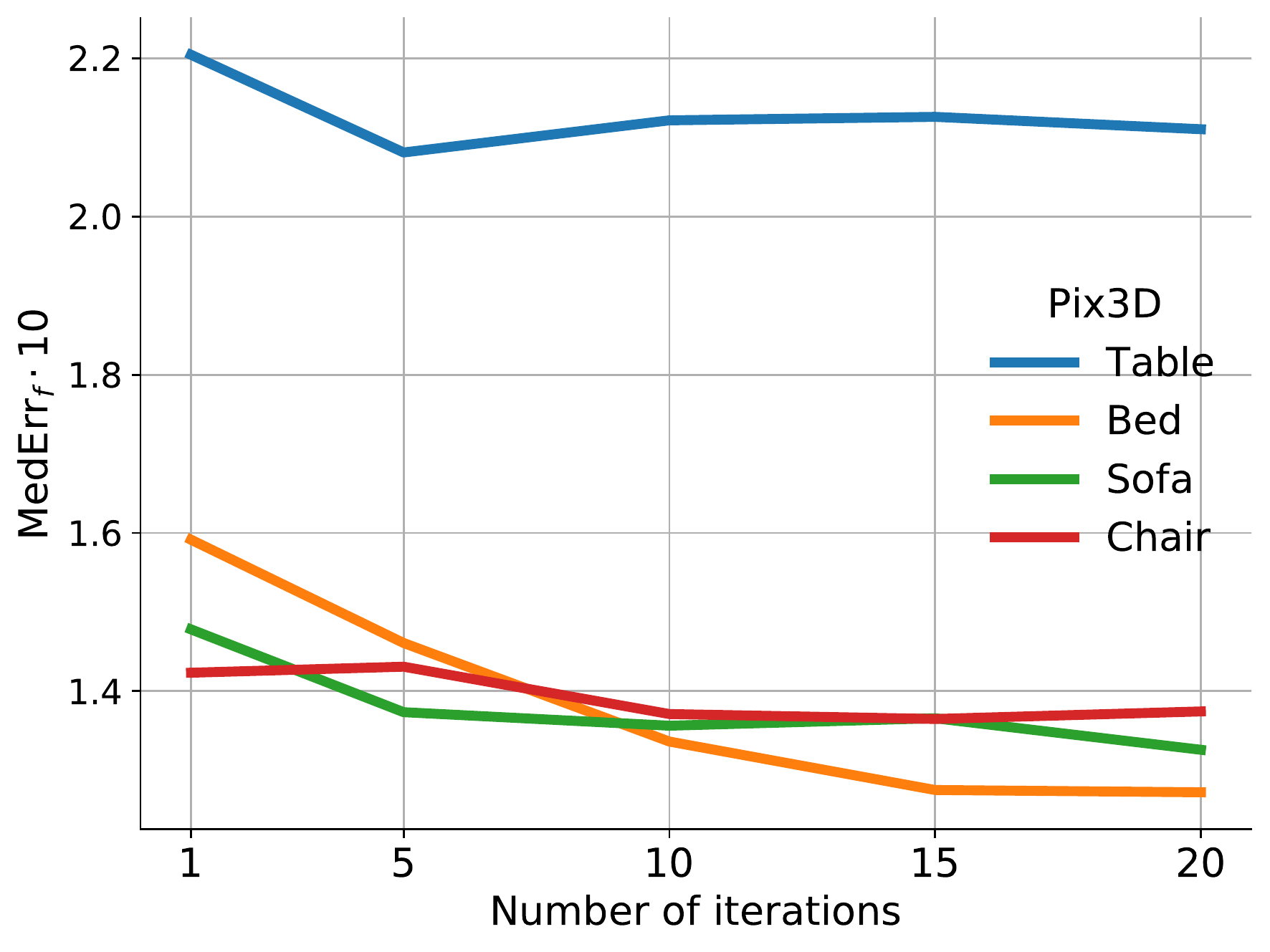}
        \end{minipage}
        \begin{minipage}{0.675\columnwidth}
            \textbf{Translation error}
            \centering
            \includegraphics[width=\textwidth]{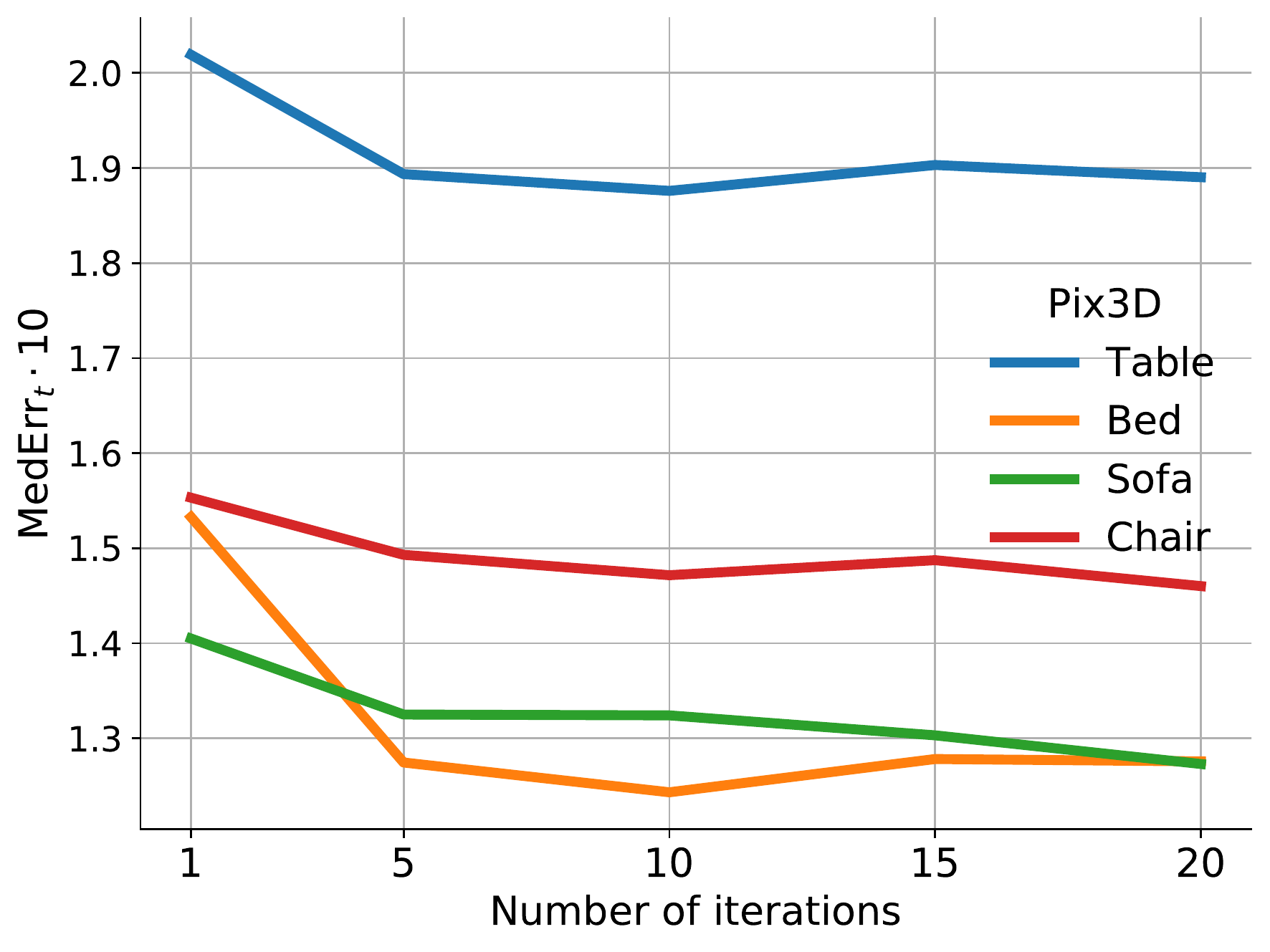}
        \end{minipage}
        \begin{minipage}{0.675\columnwidth}
            \textbf{Rotation error}
            \centering
            \includegraphics[width=\textwidth]{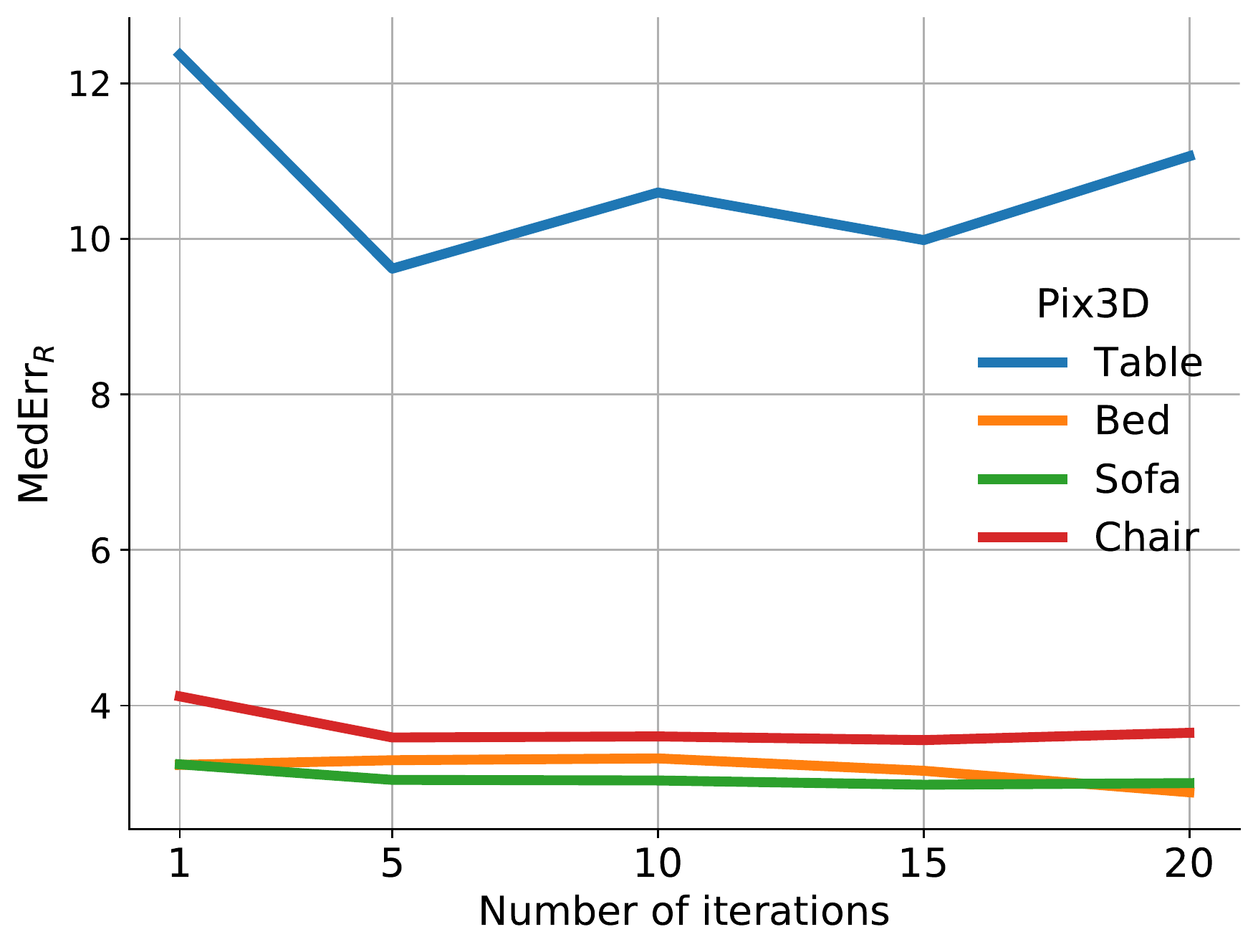}
        \end{minipage}\\[1mm]
    
    \caption{{Evolution of errors with an increasing number of refiner iterations at inference time for different object classes on the Pix3D dataset}. } 
    \label{fig:iter_errs}
    \vspace*{-5mm}
\end{figure*}

Finally, in Figure~\ref{fig:iter_errs}, we show how the model performance evolves with an increasing number of refiner iterations at inference time. Two effects can be observed. First, the translation and focal length errors tend to go down with the number of iterations and they empirically reach a fixed error value. On the other hand, we observe that the rotation errors can increase with the number of iterations, which can be seen for the Pix3D table class.

We believe this finding could be attributed to the fact that our refiner model is trained only for one iteration.  These results can be potentially improved by increasing the number of refiner iterations during training at the cost of additional compute.

\begin{table}[t]
    \centering
    \small
    \setlength{\tabcolsep}{3.2pt}
    \begin{tabular}{l|c|c|c}
    \toprule
    Dataset & $MedErr_R$ & $MedErr_t \cdot 10$ & $MedErr_f \cdot 10$ \\
    \midrule
         Synth only & 5.44 & 2.18 & 2.04 \\
		 Synth + Real 0.5\% & \textbf{2.98} & \textbf{1.29} & \textbf{1.36} \\
		 Synth + Real 5\% & 3.08 & 1.33 & 1.40 \\
		 Real only & 4.13 & 1.92 & 1.91 \\
		\bottomrule
    \end{tabular}
    \vspace*{-3mm}
    \caption{{\bf Ablation for combining real and synthetic training data on Pix3D sofa dataset.} Mix of mostly synthetic data with a small number of real images in each mini-batch performs best. }
    \vspace*{-4mm}
    \label{tab:data_ablation}
\end{table}

\subsection{Detailed results}
\label{detailed-results}
To show fine-grained information about the errors of our model, we provide a set of histograms and plots that are complementary to the results in Table~\ref{table:pix3d} in the main paper and Table~\ref{table:pix3d-perclass} in this supplementary material. 

\begin{figure*}[t]
    \centering
    \small{1}
        \begin{minipage}{0.8\columnwidth}
            \textbf{Projection error histograms}
            \centering
            \includegraphics[width=\textwidth]{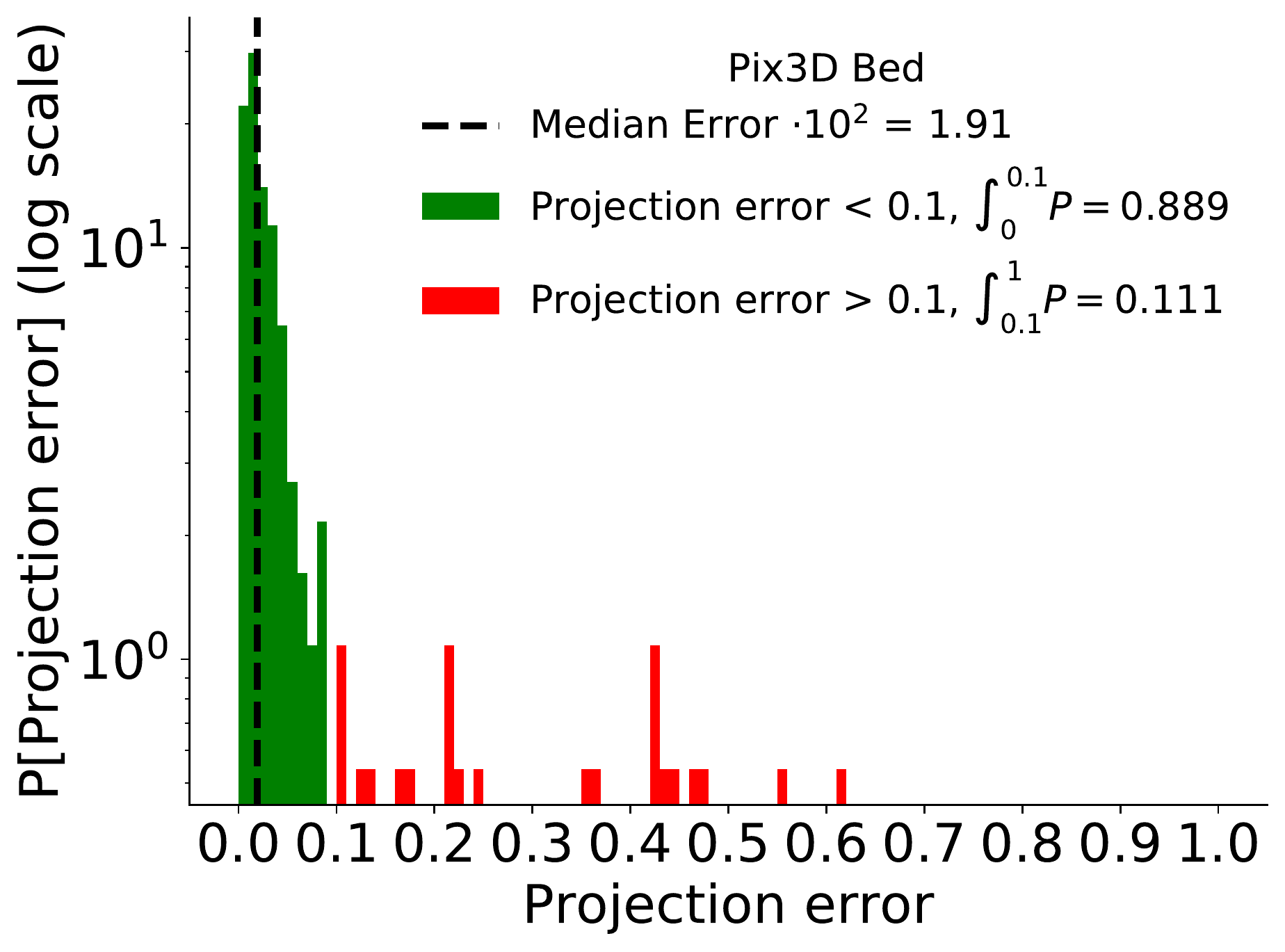}
        \end{minipage}
        \begin{minipage}{0.8\columnwidth}
            \textbf{Rotation error histograms}
            \centering
            \includegraphics[width=\textwidth]{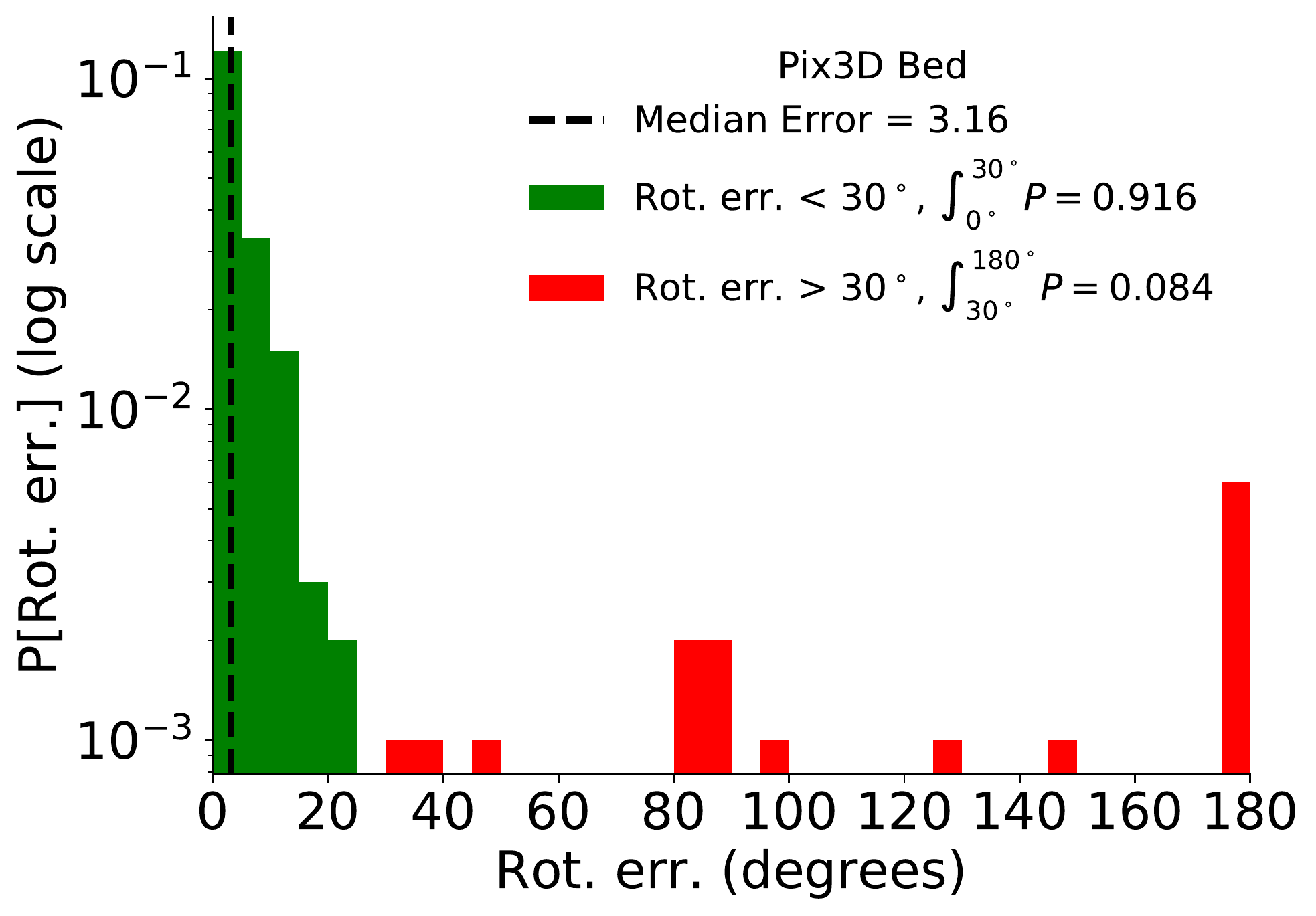}
        \end{minipage}\\[1mm]
    \small{2}
        \begin{minipage}{0.8\columnwidth}
            \centering
            \includegraphics[width=\textwidth]{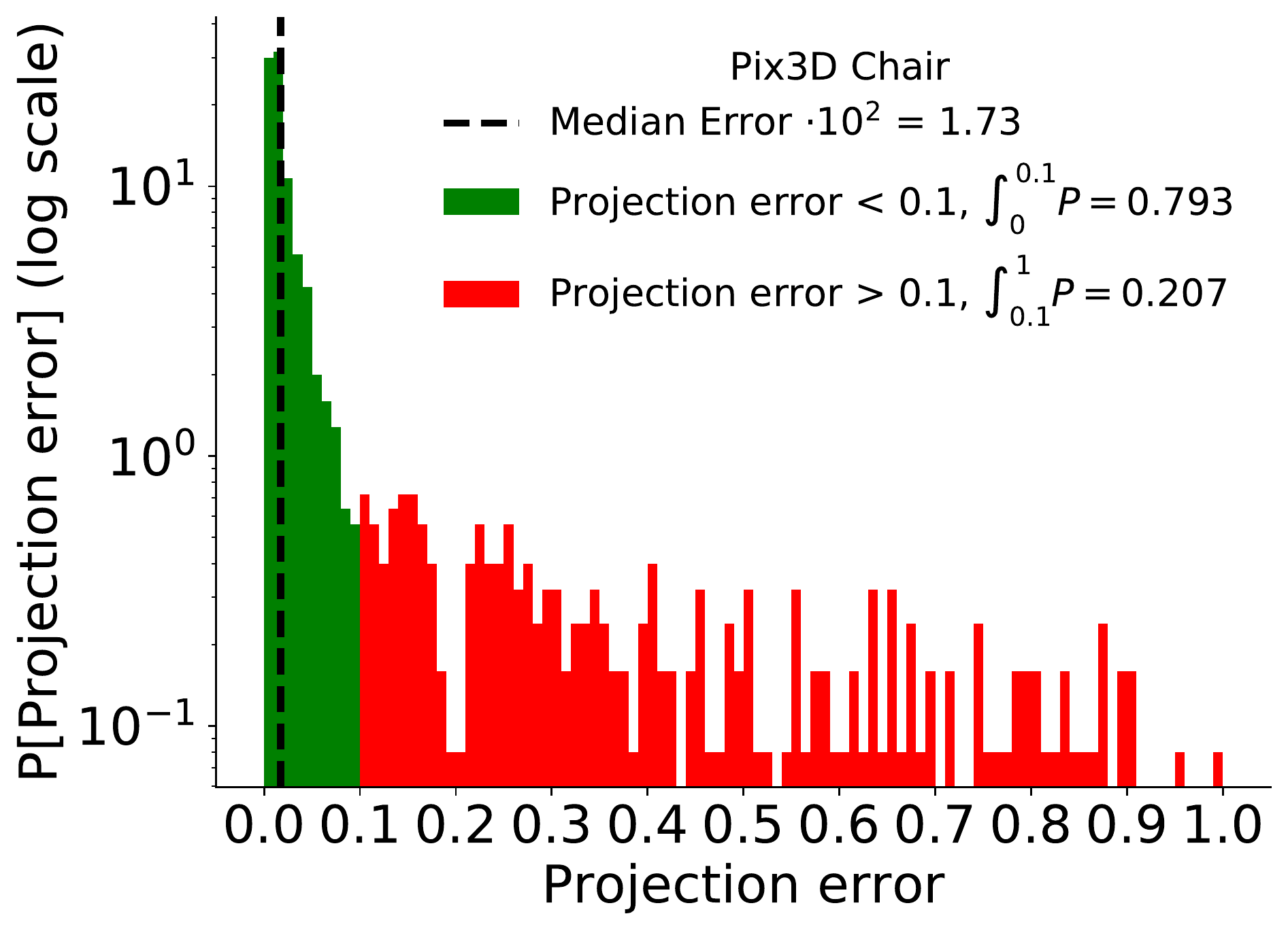}
        \end{minipage}
        \begin{minipage}{0.8\columnwidth}
            \centering
            \includegraphics[width=\textwidth]{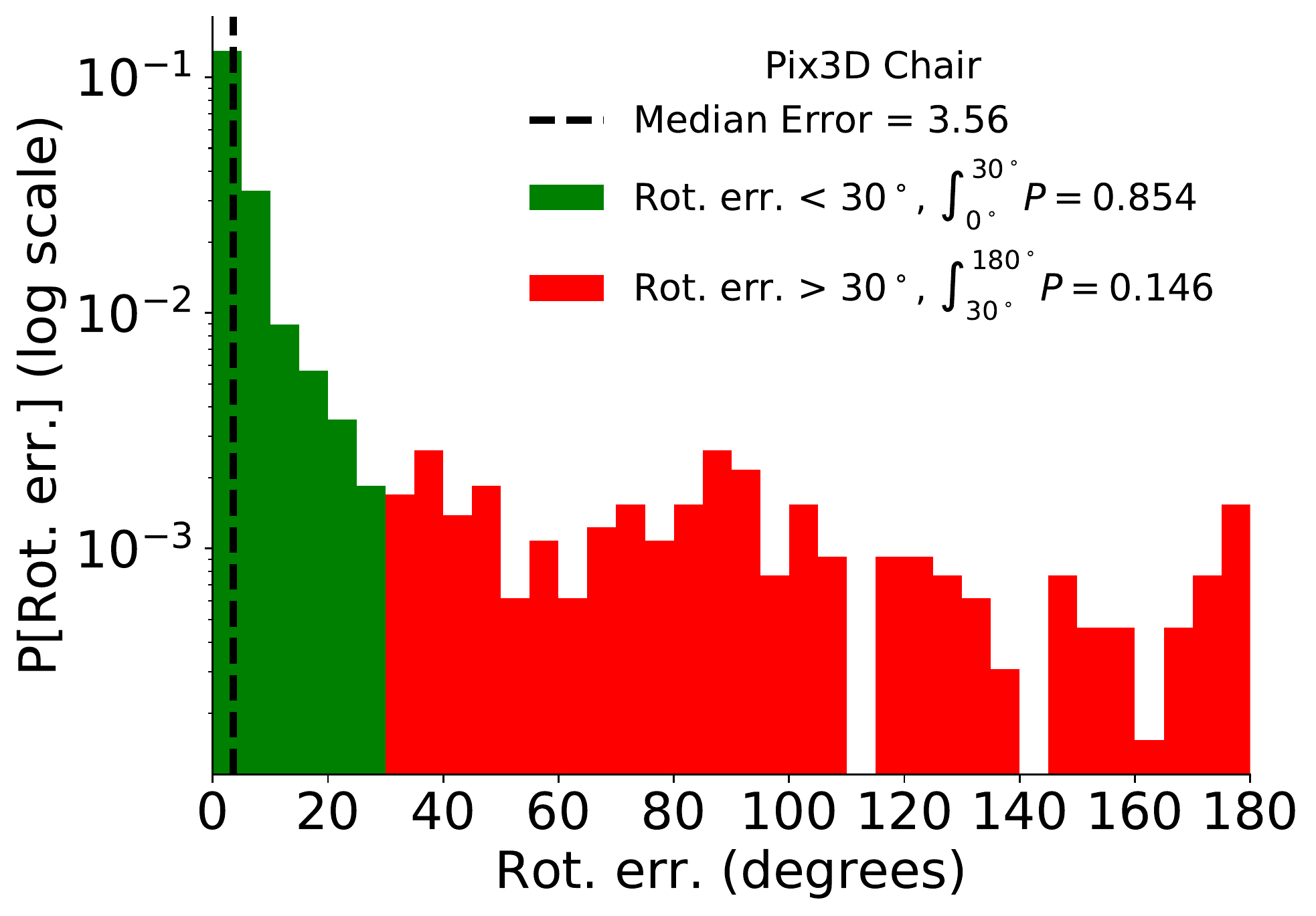}
        \end{minipage}\\[1mm]
     \small{3}
        \begin{minipage}{0.8\columnwidth}
            \centering
            \includegraphics[width=\textwidth]{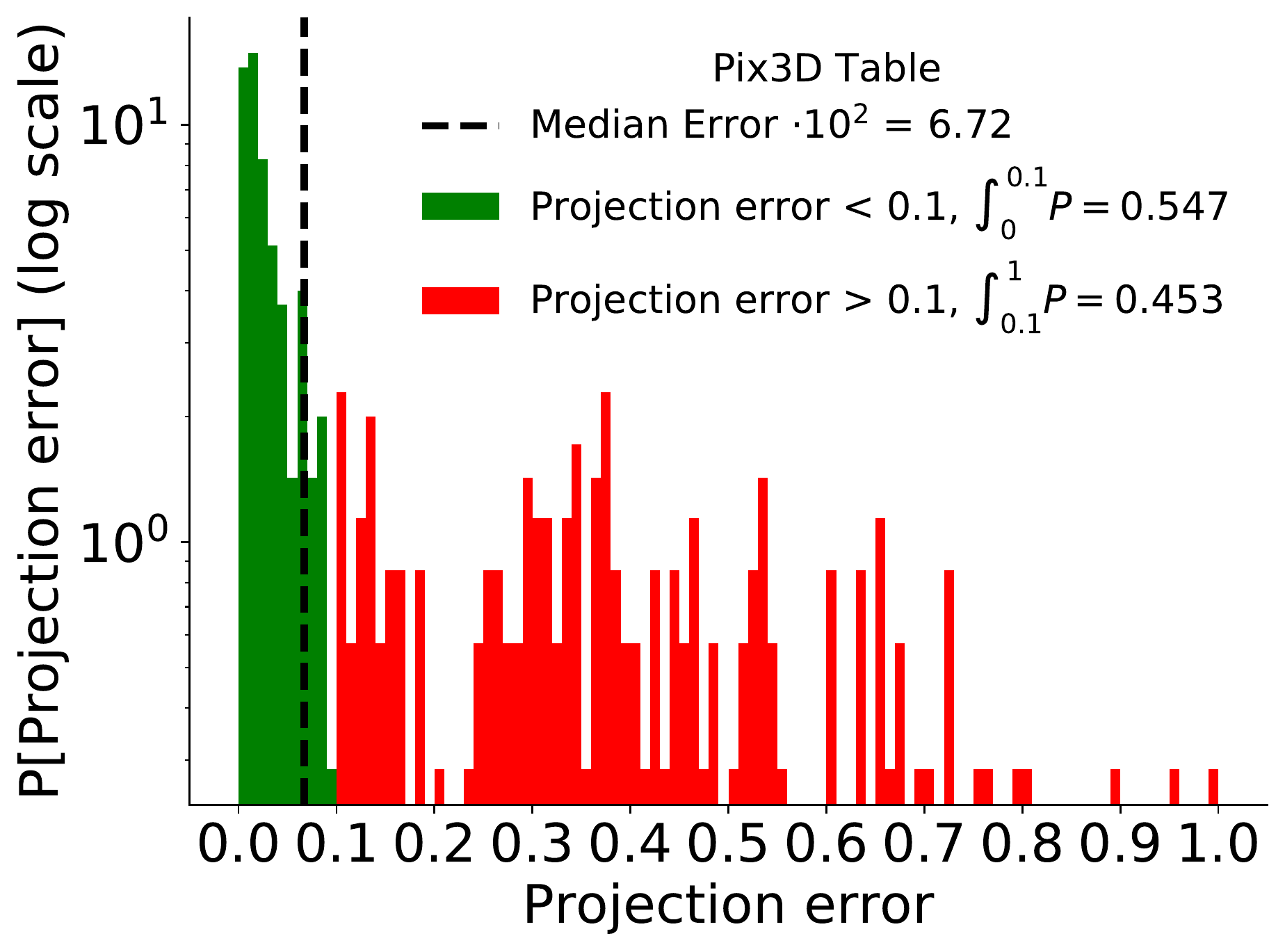}
        \end{minipage}
        \begin{minipage}{0.8\columnwidth}
            \centering
            \includegraphics[width=\textwidth]{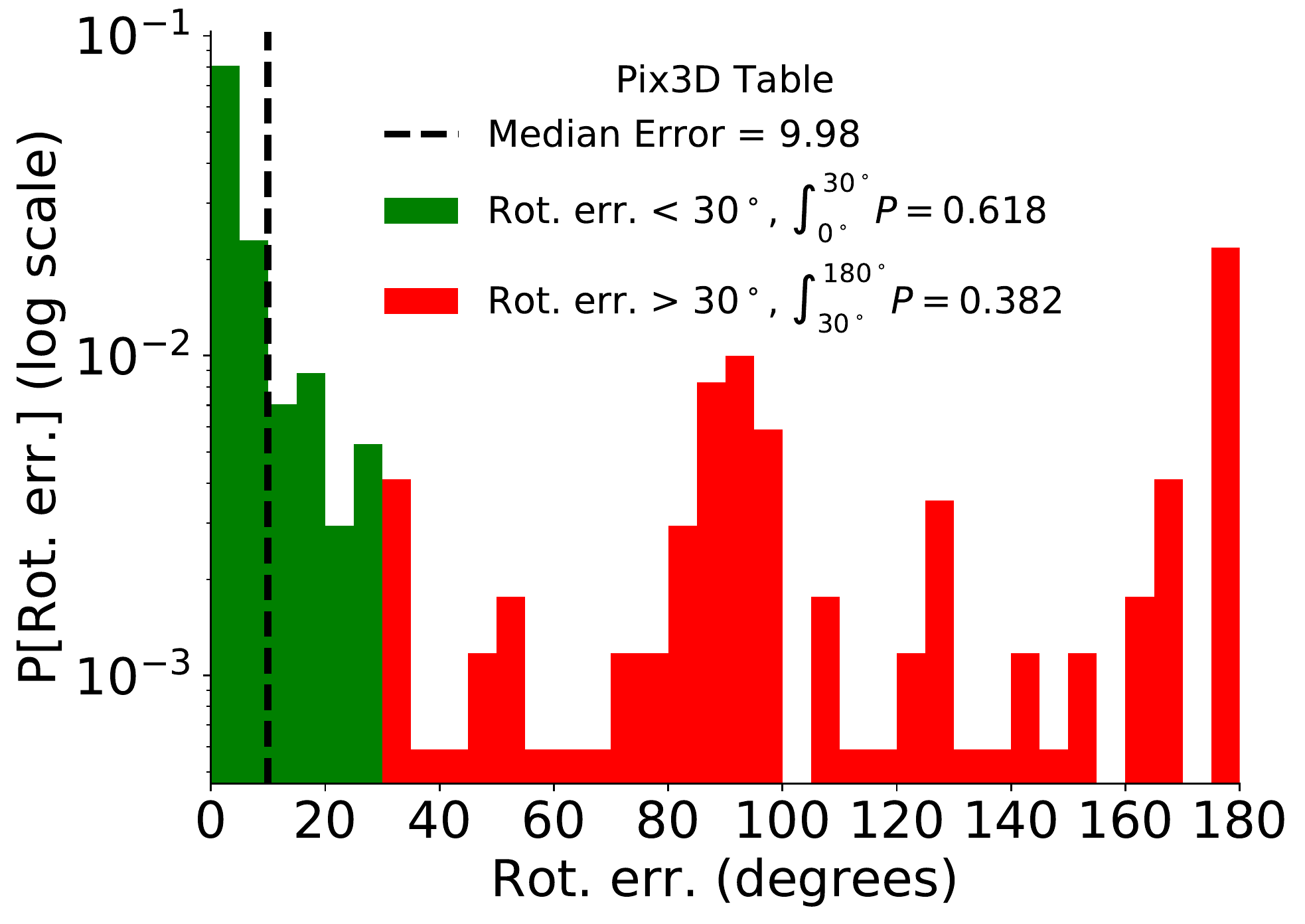}
        \end{minipage}\\[1mm]
     \small{4}
        \begin{minipage}{0.8\columnwidth}
            \centering
            \includegraphics[width=\textwidth]{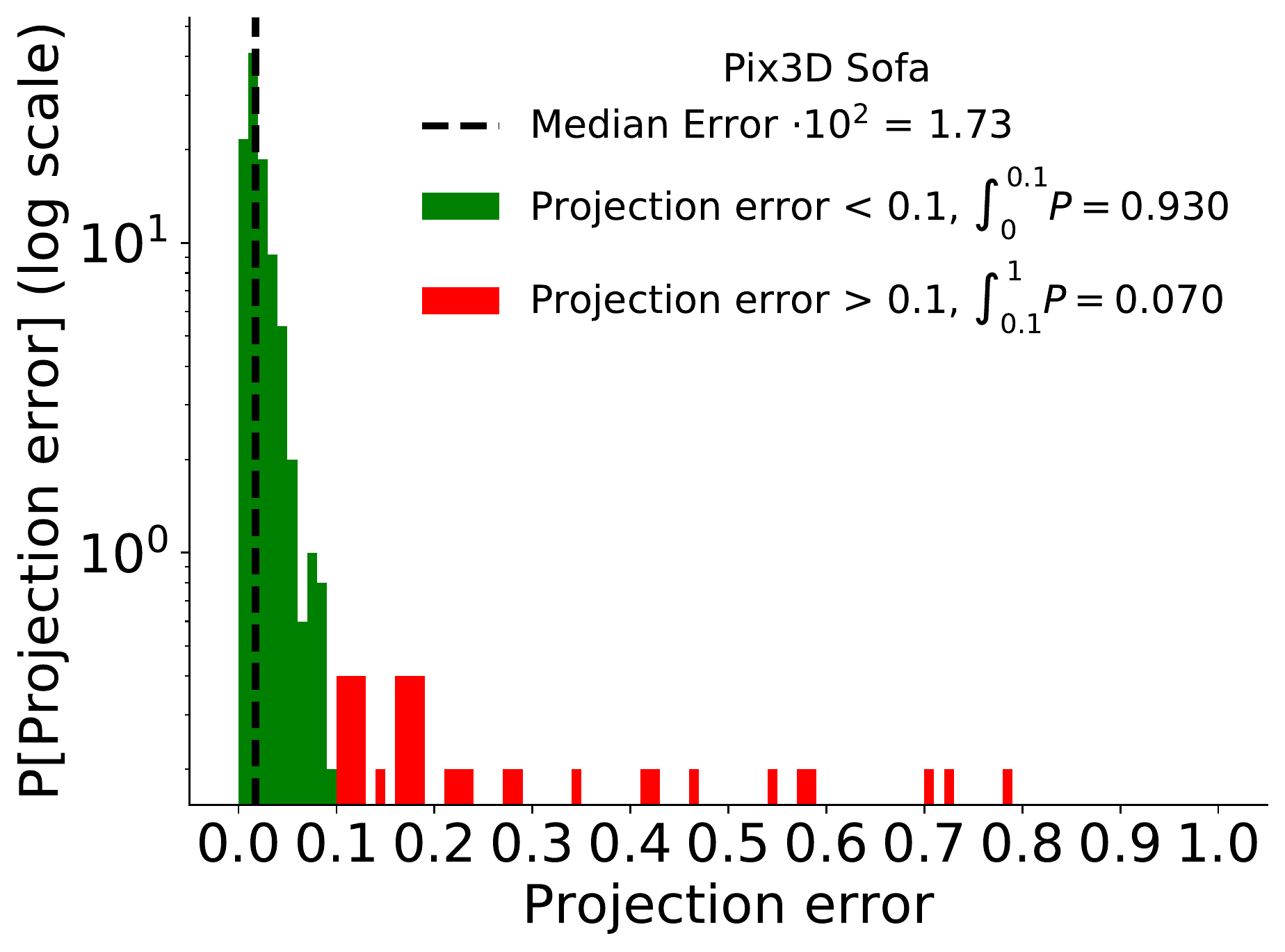}
        \end{minipage}
        \begin{minipage}{0.8\columnwidth}
            \centering
            \includegraphics[width=\textwidth]{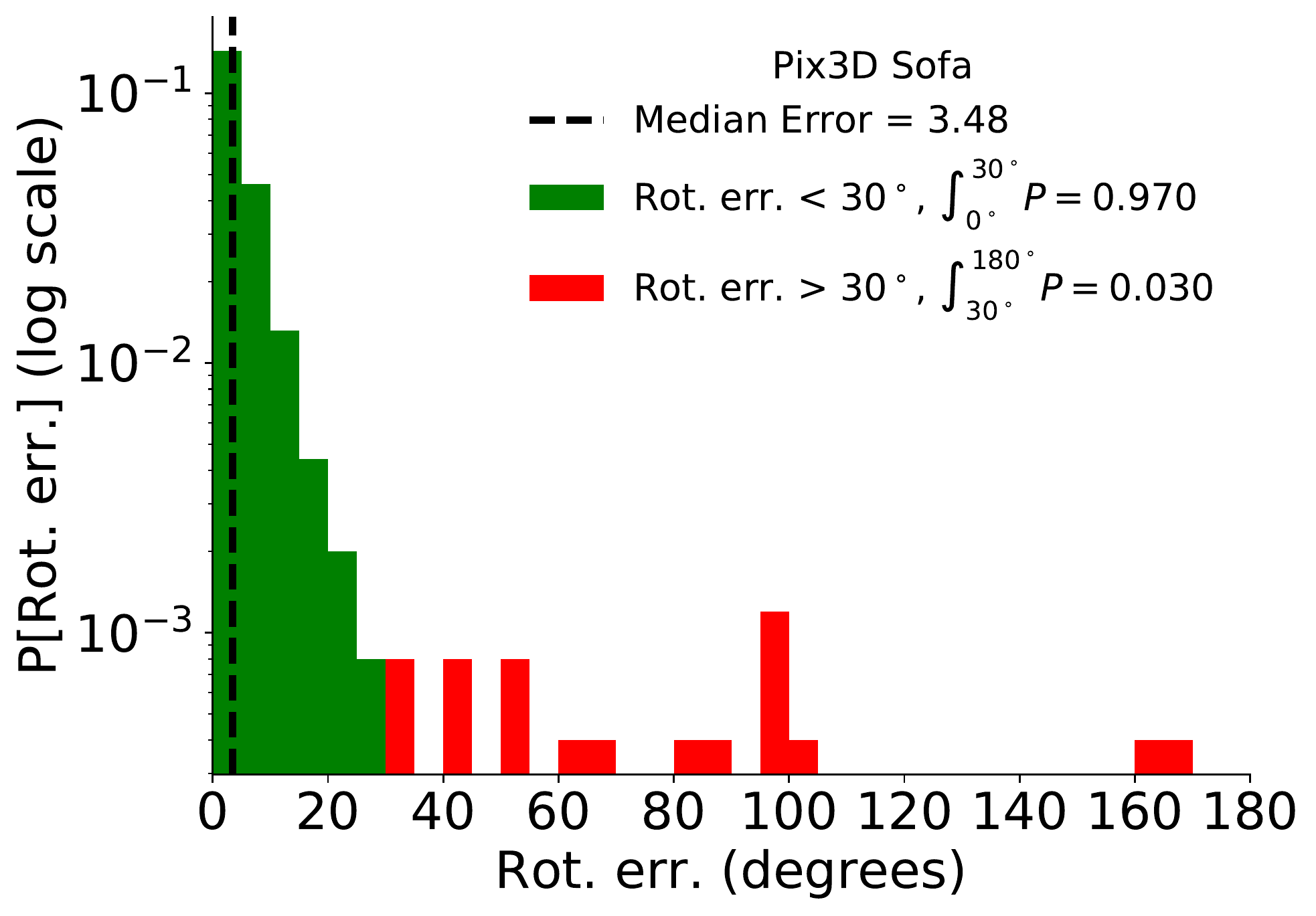}
        \end{minipage}\\[1mm]
            \vspace*{-3mm}
    \caption{\textbf{Projection error histograms (left) and rotation error histograms (right) for the Pix3D object classes.} Please note the logarithmic scale of the y-axis.} 
    \label{fig:err_hist_pix3d}
\end{figure*}

\begin{figure*}[t]
    \centering
    \small{1}
        \begin{minipage}{0.8\columnwidth}
            \textbf{Projection error histograms}
            \centering
            \includegraphics[width=\textwidth]{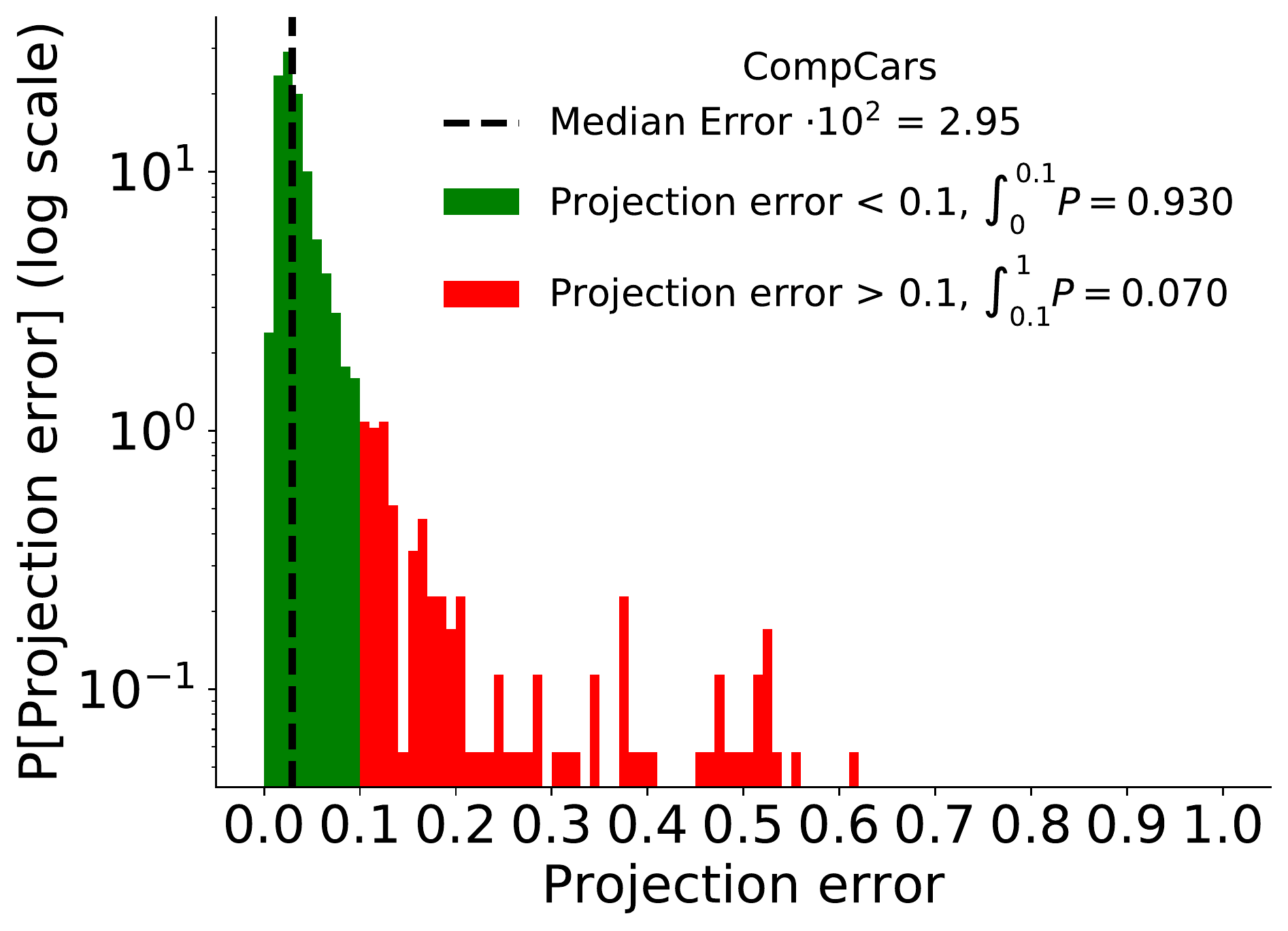}
        \end{minipage}
        \begin{minipage}{0.8\columnwidth}
            \textbf{Rotation error histograms}
            \centering
            \includegraphics[width=\textwidth]{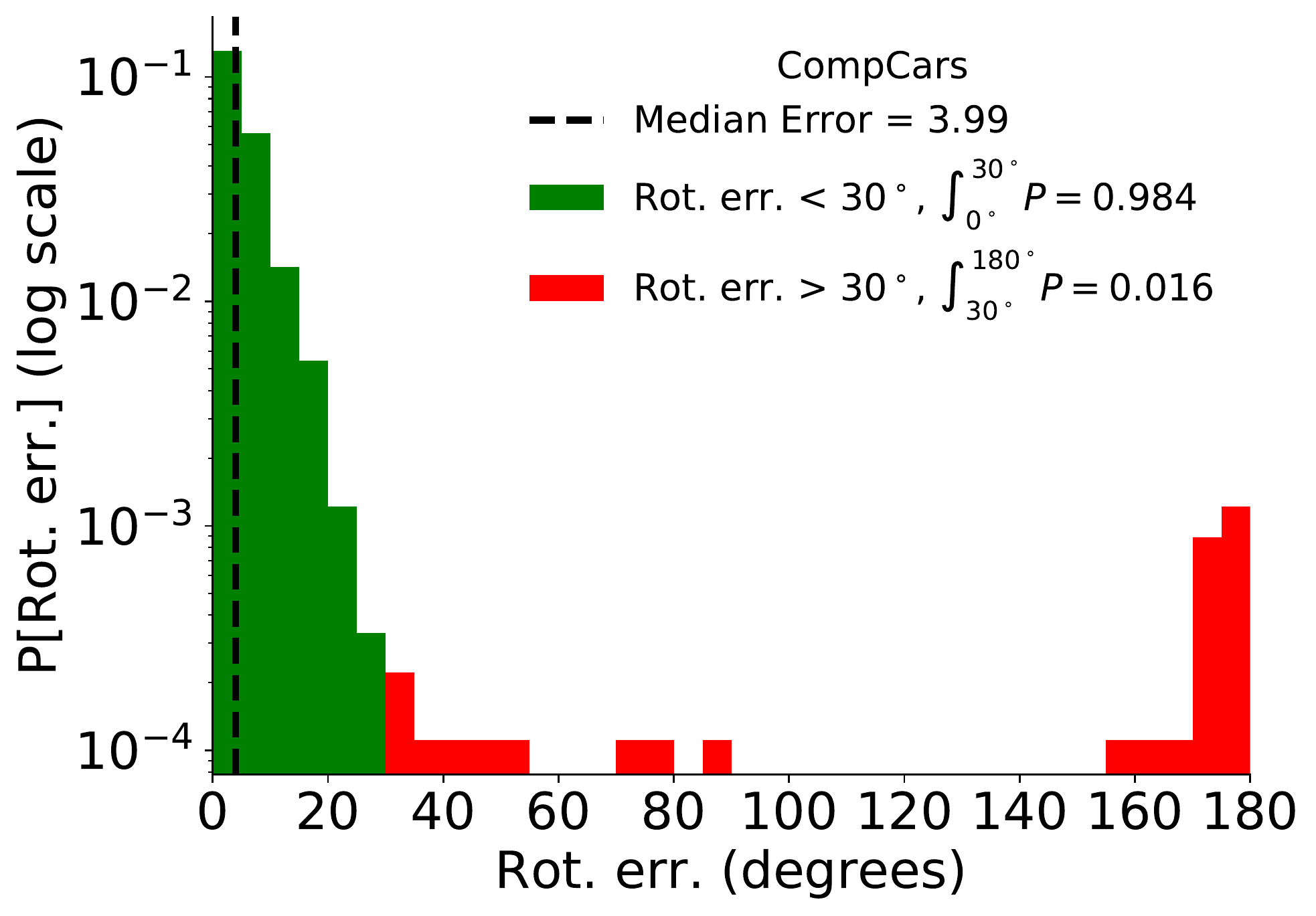}
        \end{minipage}\\[1mm]
    \small{2}
        \begin{minipage}{0.8\columnwidth}
            \centering
            \includegraphics[width=\textwidth]{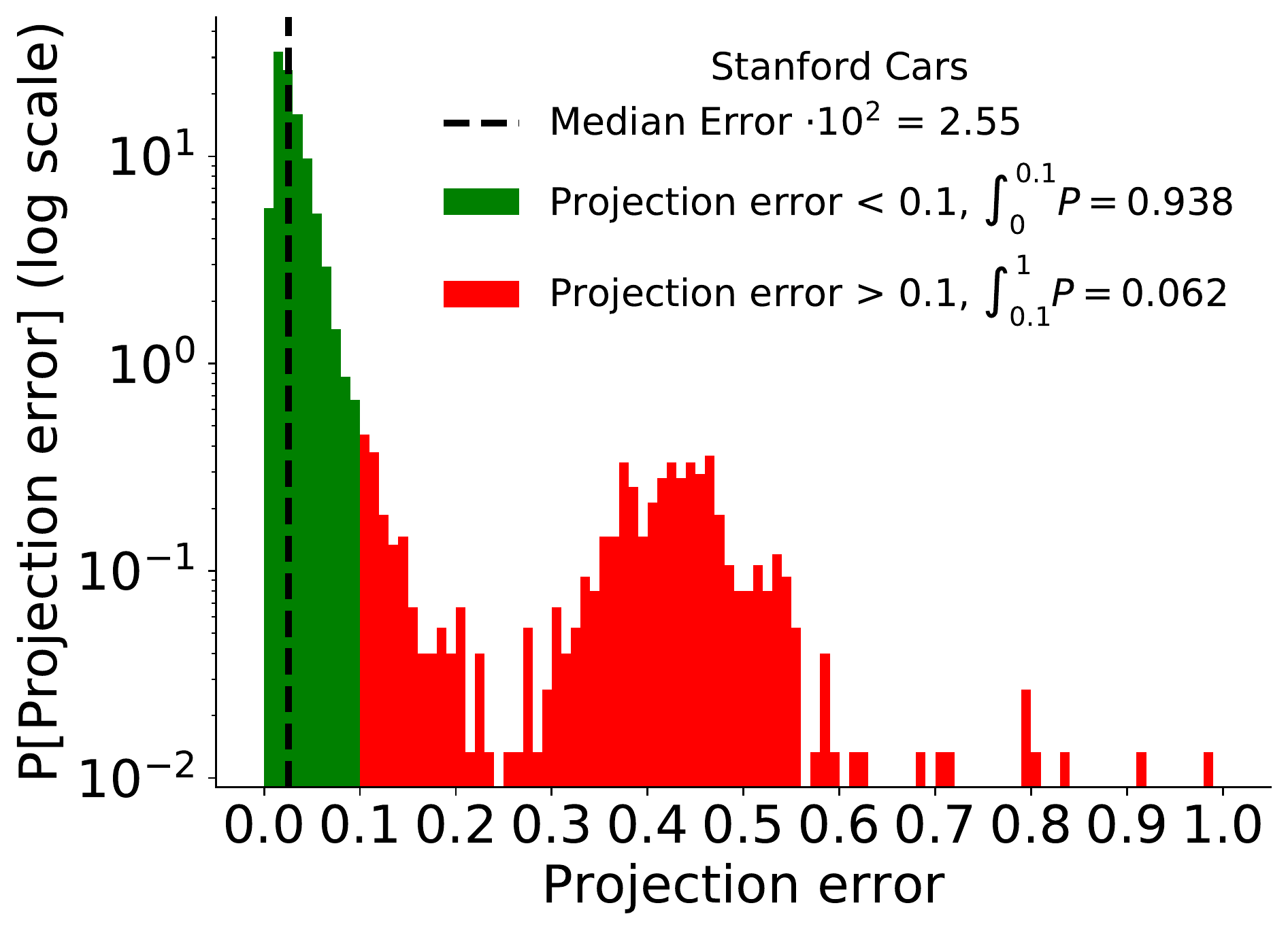}
        \end{minipage}
        \begin{minipage}{0.8\columnwidth}
            \centering
            \includegraphics[width=\textwidth]{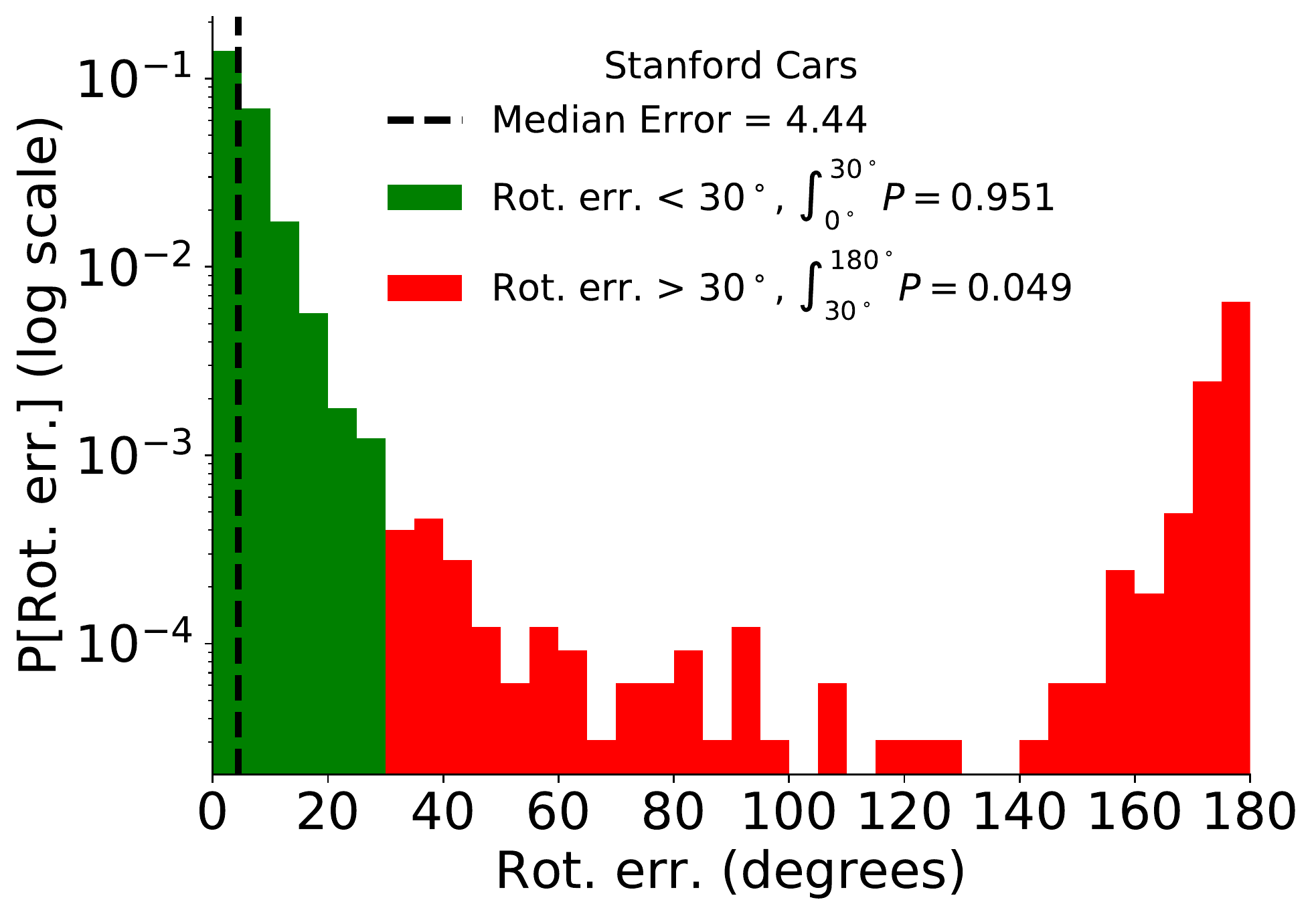}
        \end{minipage}\\[1mm]
    \vspace*{-3mm}
    \caption{\textbf{Projection error histograms (left) and rotation error histograms (right) for the CompCars (first row) and Stanford Cars (second row) datasets.} Please note the logarithmic scale of the y-axis.} 
    \label{fig:err_hist_cars}
\end{figure*}

In Figure~\ref{fig:err_hist_pix3d} we show the distributions of rotation and reprojection errors for the Pix3D dataset and in Figure~\ref{fig:err_hist_cars} for the CompCars and Stanford Cars datasets. For the Pix3D chair and table classes we observe peaks at $\sim\hspace{-0.75mm}90^\circ$ intervals, which suggests that many errors in those classes come from symmetrical objects that cause problems for our approach. For the car datasets we observe a large peak at $\sim\hspace{-0.75mm}180^\circ$, which also shows that some of the car models are fitted to incorrect orientations due to (almost) symmetrical models.

\begin{figure*}[t]
    \centering
    \small{1}
        \begin{minipage}{0.8\columnwidth}
            \textbf{Projection accuracy at different thresholds}
            \centering
            \includegraphics[width=\textwidth]{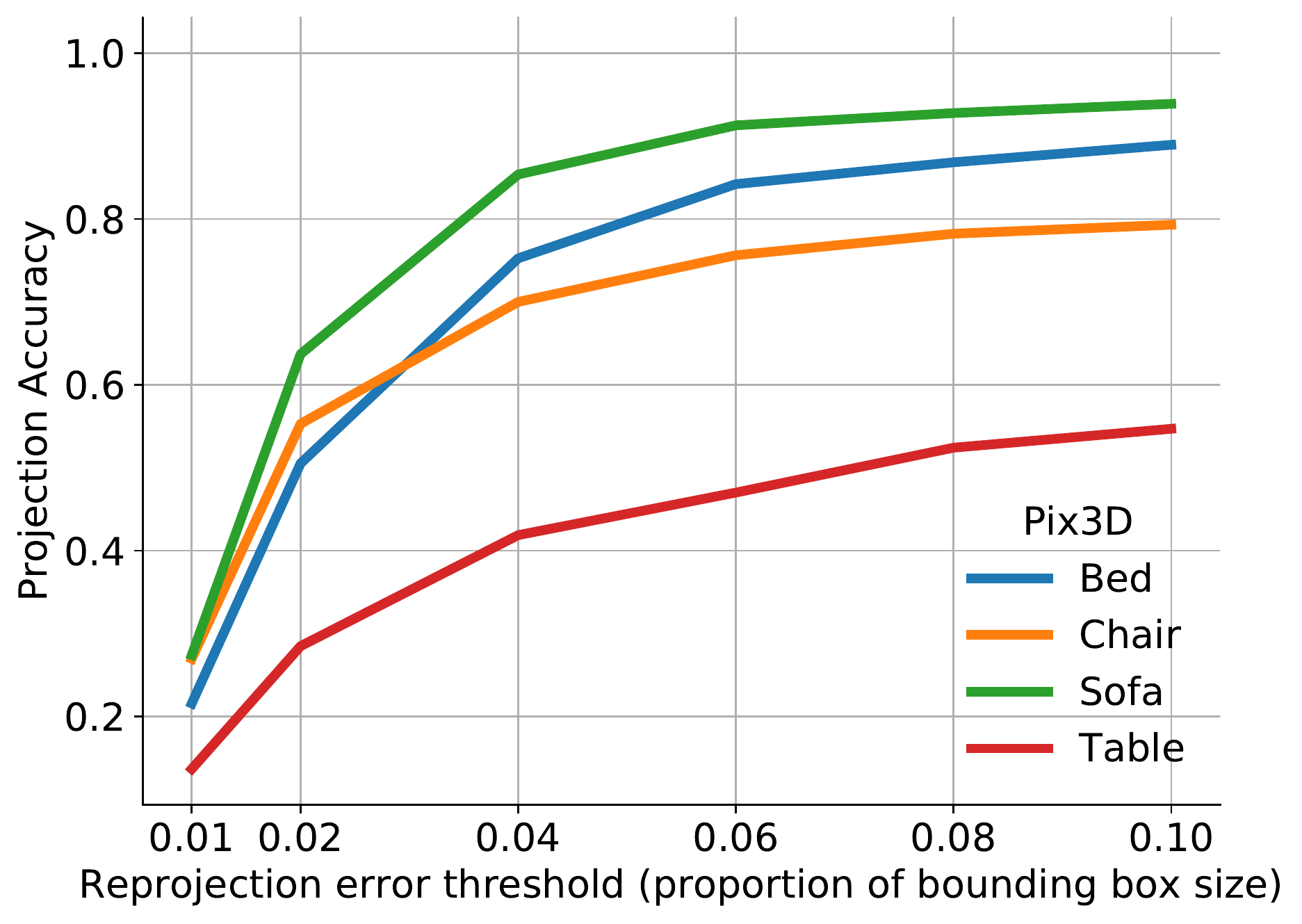}
        \end{minipage}
        \begin{minipage}{0.8\columnwidth}
            \textbf{Rotation accuracy at different thresholds}
            \centering
            \includegraphics[width=\textwidth]{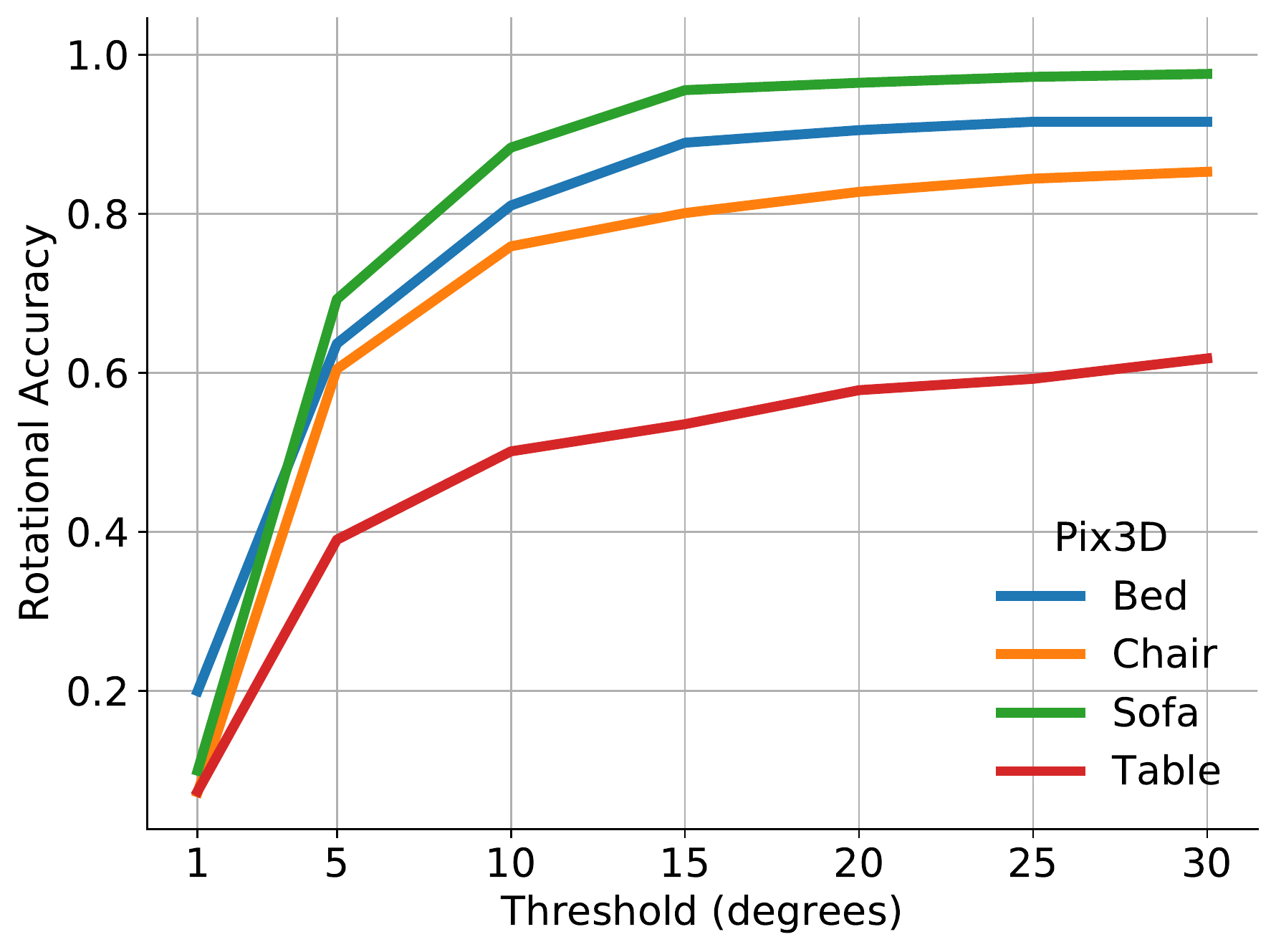}
        \end{minipage}\\[1mm]
    \small{2}
        \begin{minipage}{0.8\columnwidth}
            \centering
            \includegraphics[width=\textwidth]{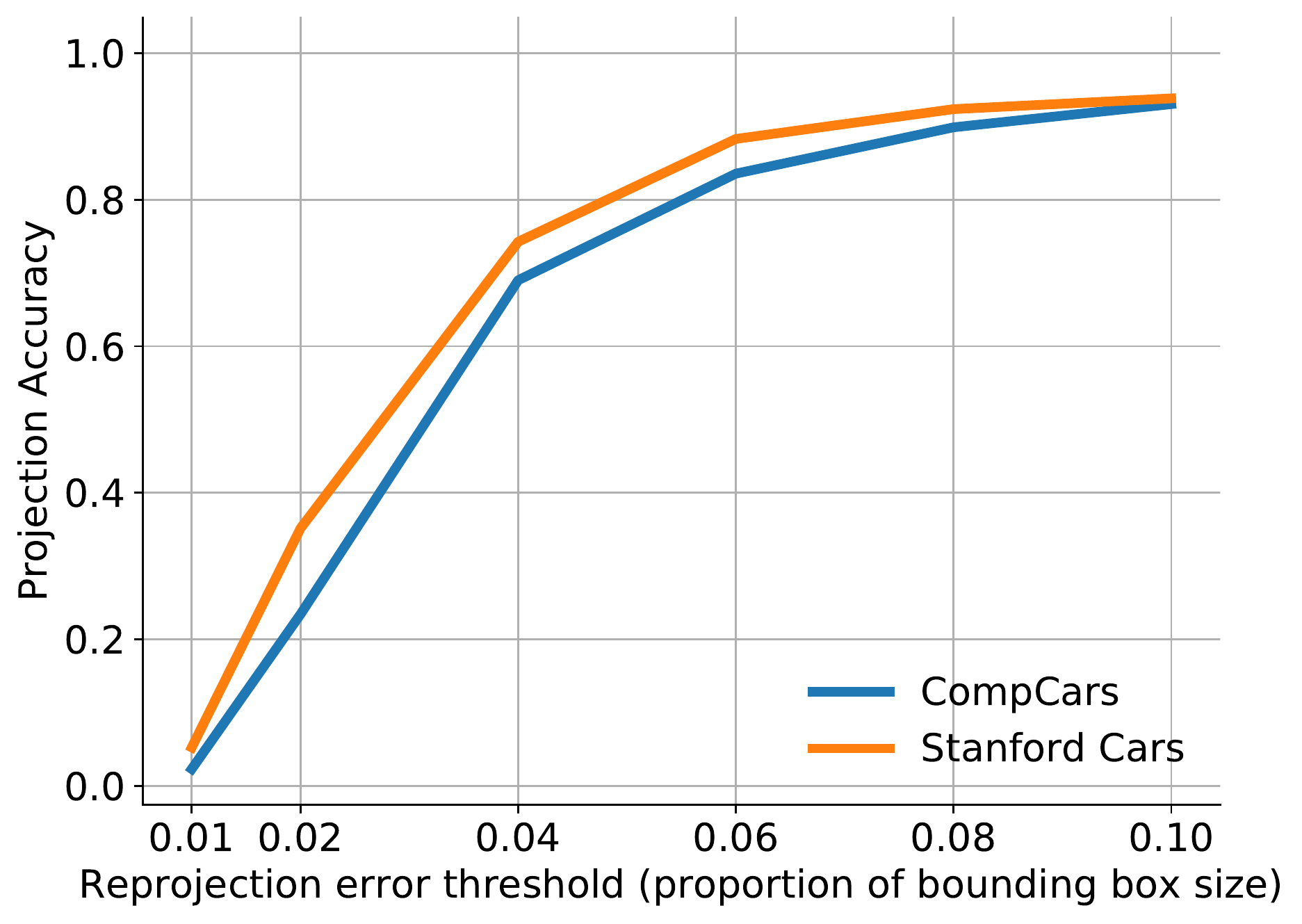}
        \end{minipage}
        \begin{minipage}{0.8\columnwidth}
            \centering
            \includegraphics[width=\textwidth]{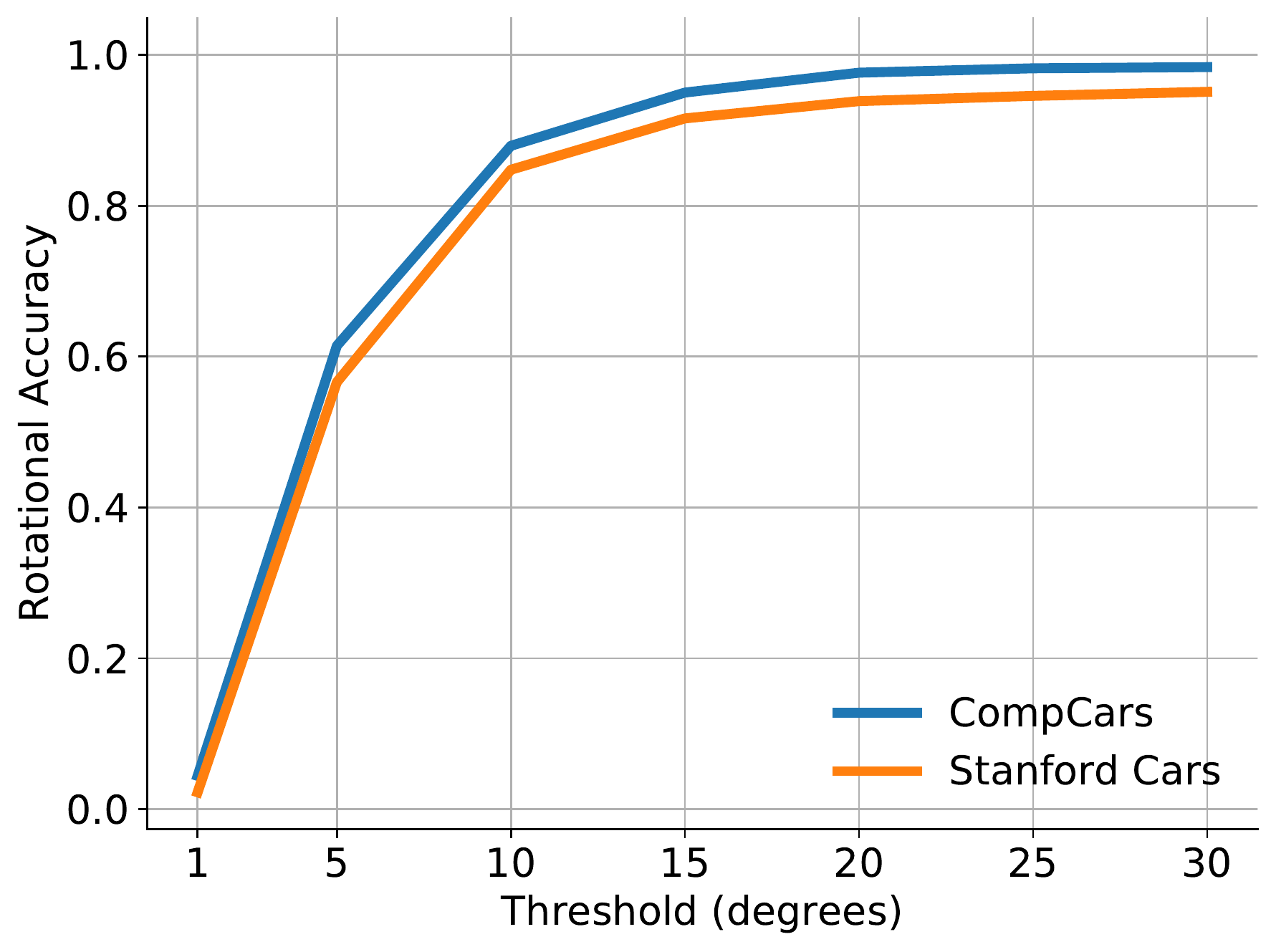}
        \end{minipage}\\[1mm]
        \vspace*{-3mm}
    \caption{{Projection and rotation accuracies at different error thresholds.}} 
    \label{fig:threshold_errs}
\end{figure*}

Figure~\ref{fig:threshold_errs} shows rotation and projection accuracies at different projection and rotation error thresholds. The standard thresholds used in previous work are quite loose and correspond to the right-most endpoints of the reported graphs, \ie, reprojection error of 0.1 (10\% of the object bounding box size) and rotation error of 30 degrees. We observe that the accuracy of our approach drops only slightly over a range of tighter thresholds, up to 0.05 relative reprojection error and up to about 15$^\circ$ rotation error.  For stricter thresholds (below around 0.05 and 15$^\circ$), the accuracy of our model starts dropping  significantly, which shows that there is still space for improvement in future work. 

\begin{figure}[t]
    \centering
        \begin{minipage}{0.31\columnwidth}
            {\small Input image}
            \centering
            \includegraphics[width=\textwidth]{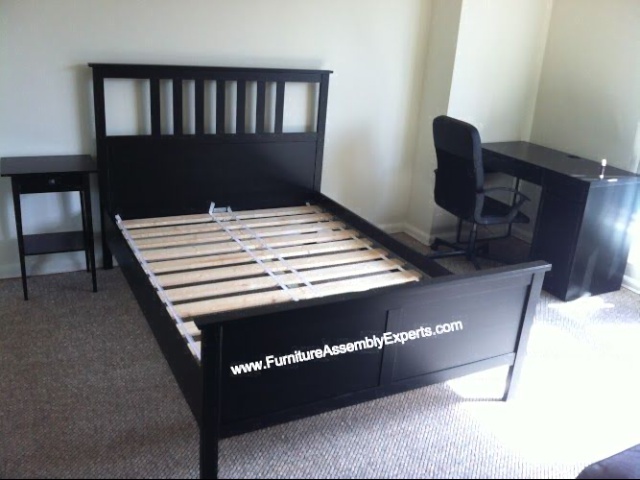}
        \end{minipage}
        \begin{minipage}{0.31\columnwidth}
            {\small Ground truth}
            \centering
            \includegraphics[width=\textwidth]{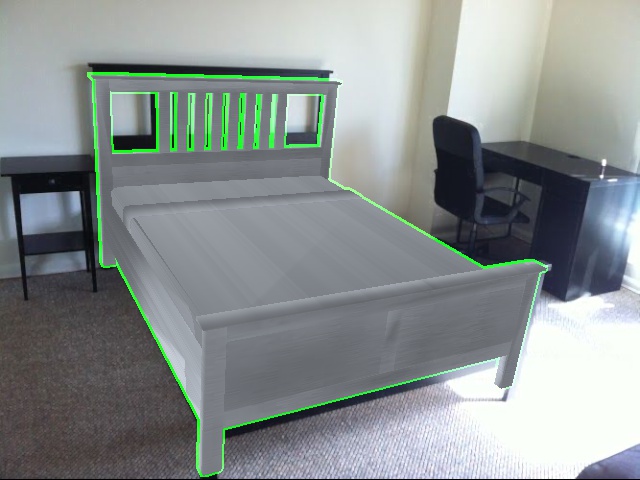}
        \end{minipage}
        \begin{minipage}{0.31\columnwidth}
            {\small Our prediction}
            \centering
            \includegraphics[width=\textwidth]{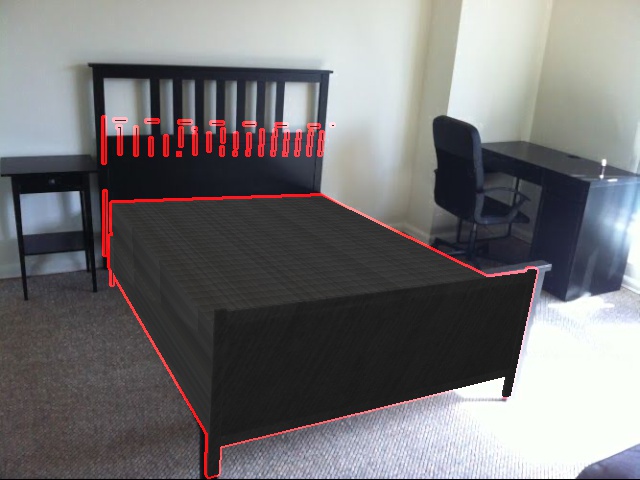}
        \end{minipage}\\[1mm]
    \caption{{\bf Inaccuracies in ground truth annotations in the Pix3D dataset.} Example of an alignment with an incorrect 3D model predicted by our approach (right) that results in a lower 3D translation and focal length errors compared to the aligned ground truth 3D model (middle). This is caused by a mismatch between the bed depicted in the input image (with no mattress) and the ground truth 3D model. }
    \label{fig:bad_model_good_error}
    \vspace*{-5mm}
\end{figure}

\begin{table*}[t]
	\centering
	\small
	\setlength{\tabcolsep}{3.2pt}
	\begin{tabular}{lcc|c|cc|c|c|c|cc}
		\toprule
		\multicolumn{3}{c}{}&\multicolumn{1}{c}{\bf Detection}&\multicolumn{2}{c}{\bf Rotation}&\multicolumn{1}{c}{\bf Translation}&\multicolumn{1}{c}{\bf Pose}&\multicolumn{1}{c}{\bf Focal}&\multicolumn{2}{c}{\bf Projection}\\

		\cmidrule(lr){4-4}\cmidrule(lr){5-6}\cmidrule(lr){7-7}\cmidrule(lr){8-8}\cmidrule(lr){9-9}\cmidrule(lr){10-11}
		\multirow{2}{*}{Method}&\multicolumn{1}{c}{\multirow{2}{*}{Dataset}}&\multicolumn{1}{c}{\multirow{2}{*}{Class}}&\multicolumn{1}{c}{\multirow{2}{*}{$Acc_{D_{0.5}}$}}&\multicolumn{1}{c}{$MedErr_R$}&\multicolumn{1}{c}{\multirow{2}{*}{$Acc_{R\frac{\pi}{6}}$}}&\multicolumn{1}{c}{$MedErr_{t}$}&\multicolumn{1}{c}{$MedErr_{R,t}$}&\multicolumn{1}{c}{$MedErr_f$}&\multicolumn{1}{c}{$MedErr_{P}$}&\multicolumn{1}{c}{\multirow{2}{*}{$Acc_{P_{0.1}}$}}\\
		&&\multicolumn{1}{c}{}&\multicolumn{1}{c}{}&\multicolumn{1}{c}{$\cdot1$}&\multicolumn{1}{c}{}&\multicolumn{1}{c}{$\cdot10^{1}$}&\multicolumn{1}{c}{$\cdot10^{1}$}&\multicolumn{1}{c}{$\cdot10^{1}$}&\multicolumn{1}{c}{$\cdot10^{2}$}\\
		\midrule
		\midrule
		\cite{wang20183d}&\multirow{5}{*}{Pix3D}&\multirow{5}{*}{bed} & 98.4\% & 5.82 & 95.3\% & 1.95 & 1.56 & 2.22 & 6.05 & 74.9\%\\
		\cite{grabner2019gp2c} LF &&& 99.0\% & 5.13 & 96.3\% & 1.41 & 1.04 & 1.43& 3.52&90.6\%\\
		\cite{grabner2019gp2c} BB &&& \bf99.5\% & 5.40 & \bf \textbf{97.9\%} & 1.66 & 1.17 & 1.59 & 3.55 &\bf{93.2\%}\\

        Ours  &&& 98.4\% & \textbf{3.16} & 91.6\% & \textbf{1.28} & \textbf{0.93} & \textbf{1.28} & \textbf{1.91} & 88.9\%\\

		\midrule
		\cite{wang20183d}&\multirow{5}{*}{Pix3D}&\multirow{5}{*}{chair}&94.9\%&7.52&88.0\%&2.69&1.58&1.98&6.04&75.3\%\\
		\cite{grabner2019gp2c}-LF&&&95.2\%&7.52&88.8\%&1.92&1.21&1.62&3.41&88.2\%\\
		\cite{grabner2019gp2c}-BB&&&\bf97.3\%&6.95&\bf{91.0\%}& 1.68&1.08&1.58&3.24&\bf{90.9\%}\\

        Ours  &&& 91.8\% & \textbf{3.56} & 85.4\% & \textbf{1.49} & \textbf{0.94} & \textbf{1.36} & \textbf{1.73} & 79.3\%\\
		
		\midrule
		\cite{wang20183d}&\multirow{5}{*}{Pix3D}&\multirow{5}{*}{sofa}&96.5\%&4.73&94.8\%&2.28&1.62&2.42&4.33&82.2\%\\
		\cite{grabner2019gp2c} LF&&&96.5\%&4.49&95.0\%&1.92&1.33&1.79&2.56&93.7\%\\
		\cite{grabner2019gp2c} BB&&&\bf98.3\%&4.40&97.0\%&1.63&1.16&1.73&2.13&\bf95.6\%\\
		
        Ours  &&& 96.9\% & \textbf{2.98} & 97.6\% & \textbf{1.29} & \textbf{0.83} & \textbf{1.36} & \textbf{1.52} & 93.9\%\\

		\midrule
		\cite{wang20183d}&\multirow{5}{*}{Pix3D}&\multirow{5}{*}{table}&94.0\%&10.94&72.9\%&3.16&2.28&3.03&8.90&53.6\%\\
		\cite{grabner2019gp2c} LF&&&94.0\%& 10.53&73.5\%& 2.16& \bf{1.62}& \textbf{2.05}&5.92&69.5\%\\
		\cite{grabner2019gp2c} BB&&& \bf95.7\%&10.80& \bf{77.2\%}&2.81&1.78&2.10&\textbf{5.74}&\bf{72.4\%}\\
		
        Ours  &&& 94.9\% & \textbf{9.98} & 61.8\% & \textbf{1.90} & 1.68 & 2.13 & 6.72 & 54.7\%\\
		 
	\bottomrule
	\end{tabular}
	\vspace*{-1mm}
	\caption{{\bf Comparison with the state of the art for 6D pose and focal length prediction} on the Pix3D dataset split by class. {\bf Bold} denotes the best result among directly comparable methods. See section~\ref{per-class-pix3d} in this supplementary for a more detailed analysis of the results.}

	\label{table:pix3d-perclass}
\end{table*}

\subsection{Additional qualitative results}
\begin{figure*}[t]
    \centering
            \begin{minipage}{0.225\textwidth}
            {\small Input image}
            \centering
            \includegraphics[width=\textwidth]{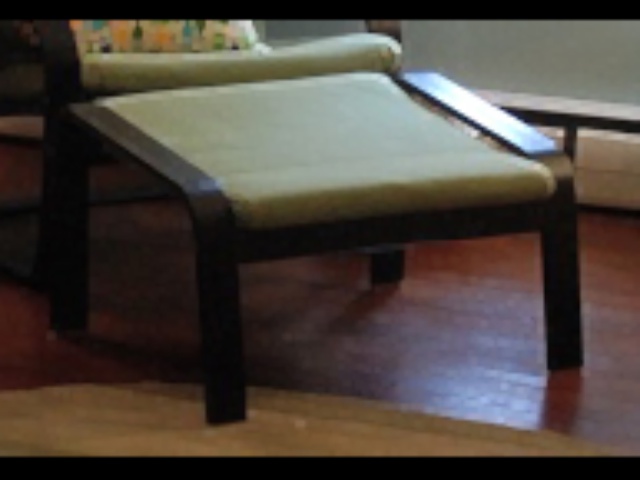}
        \end{minipage}
        \begin{minipage}{0.225\textwidth}
        {\small Ground truth}
            \centering
            \includegraphics[width=\textwidth]{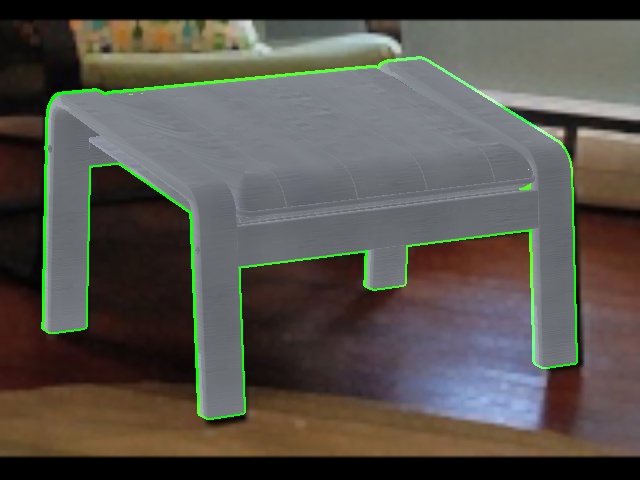}
        \end{minipage}
        \begin{minipage}{0.225\textwidth}
        {\small Our prediction}
            \centering
            \includegraphics[width=\textwidth]{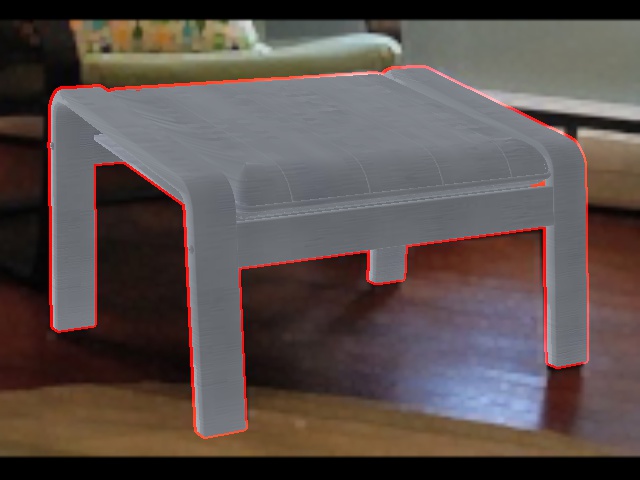}
        \end{minipage}\\[1mm]
        \begin{minipage}{0.225\textwidth}
            \centering
            \includegraphics[width=\textwidth]{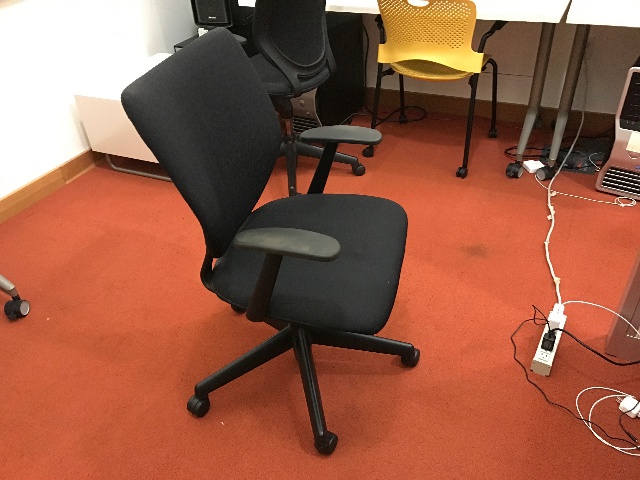}
        \end{minipage}
        \begin{minipage}{0.225\textwidth}
            \centering
            \includegraphics[width=\textwidth]{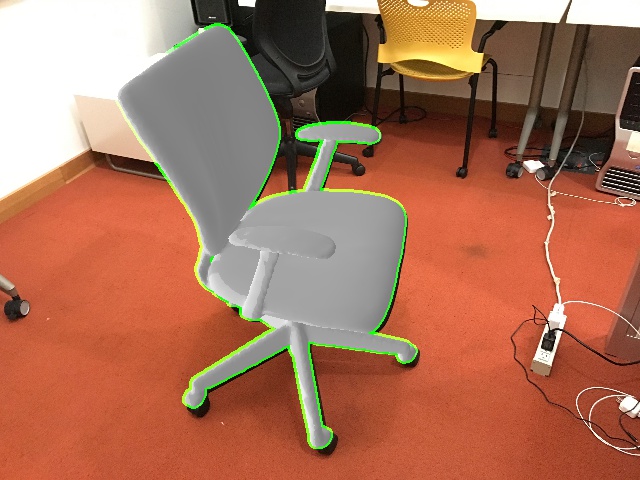}
        \end{minipage}
        \begin{minipage}{0.225\textwidth}
            \centering
            \includegraphics[width=\textwidth]{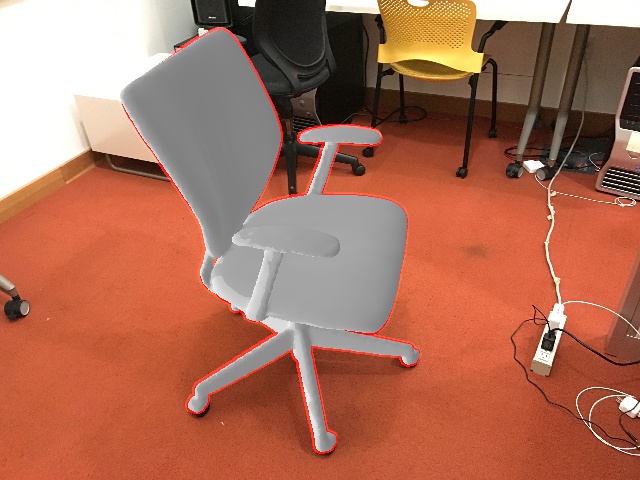}
        \end{minipage}\\[1mm]
        \begin{minipage}{0.225\textwidth}
            \centering
            \includegraphics[width=\textwidth]{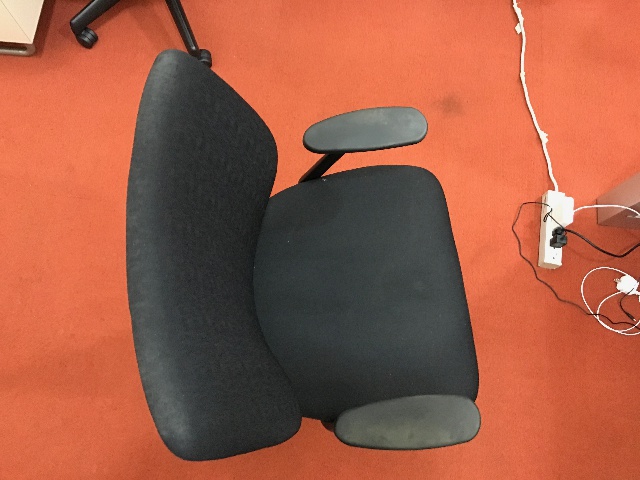}
        \end{minipage}
        \begin{minipage}{0.225\textwidth}
            \centering
            \includegraphics[width=\textwidth]{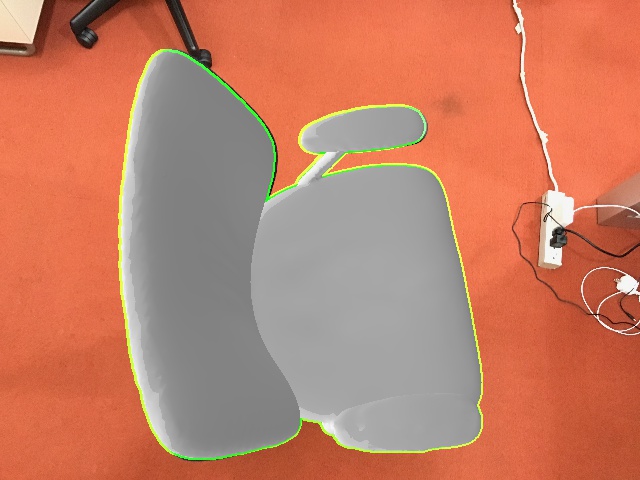}
        \end{minipage}
        \begin{minipage}{0.225\textwidth}
            \centering
            \includegraphics[width=\textwidth]{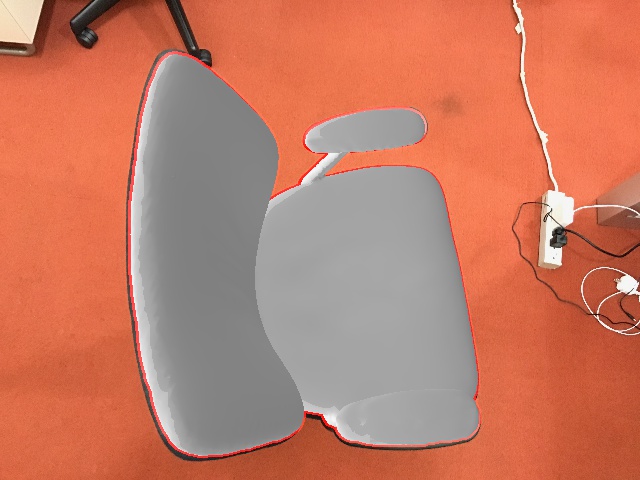}
        \end{minipage}\\[1mm]
        \begin{minipage}{0.225\textwidth}
            \centering
            \includegraphics[width=\textwidth]{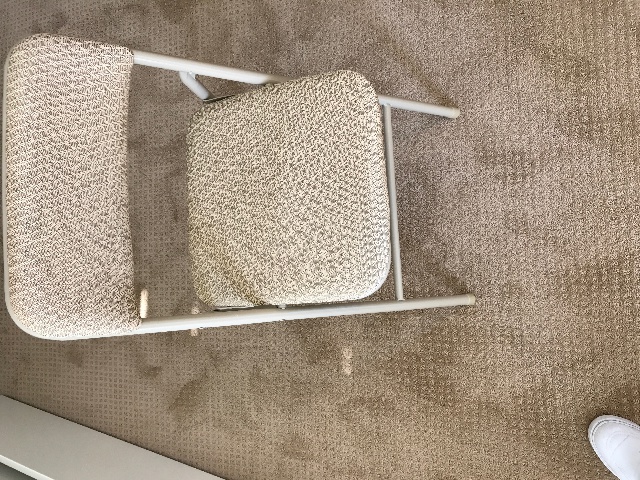}
        \end{minipage}
        \begin{minipage}{0.225\textwidth}
            \centering
            \includegraphics[width=\textwidth]{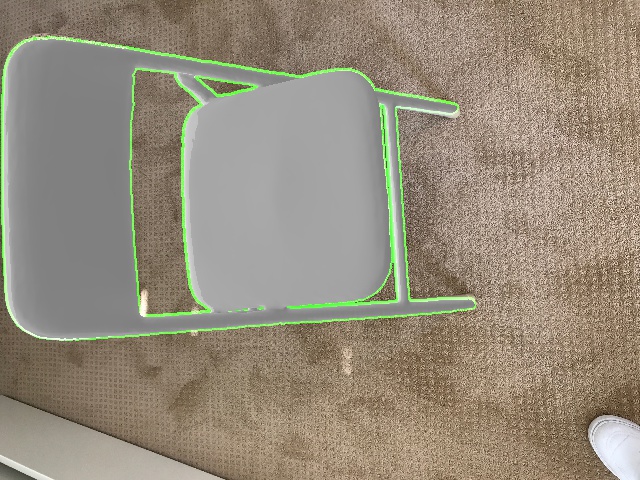}
        \end{minipage}
        \begin{minipage}{0.225\textwidth}
            \centering
            \includegraphics[width=\textwidth]{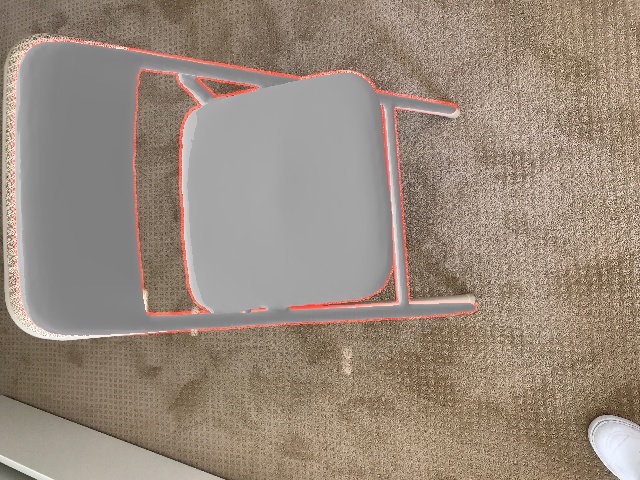}
        \end{minipage}\\[1mm]
        \begin{minipage}{0.225\textwidth}
            \centering
            \includegraphics[width=\textwidth]{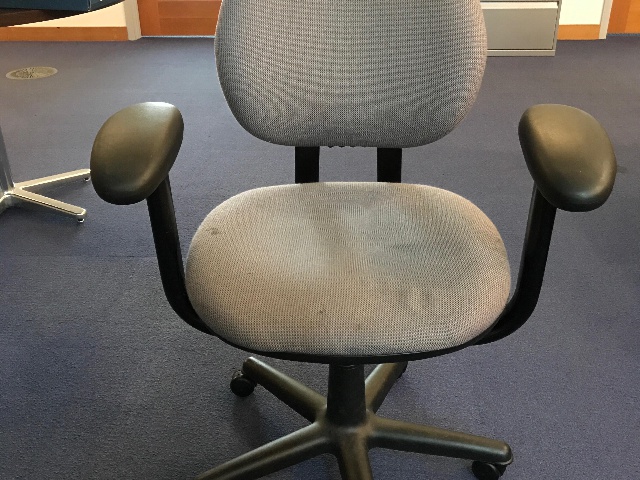}
        \end{minipage}
        \begin{minipage}{0.225\textwidth}
            \centering
            \includegraphics[width=\textwidth]{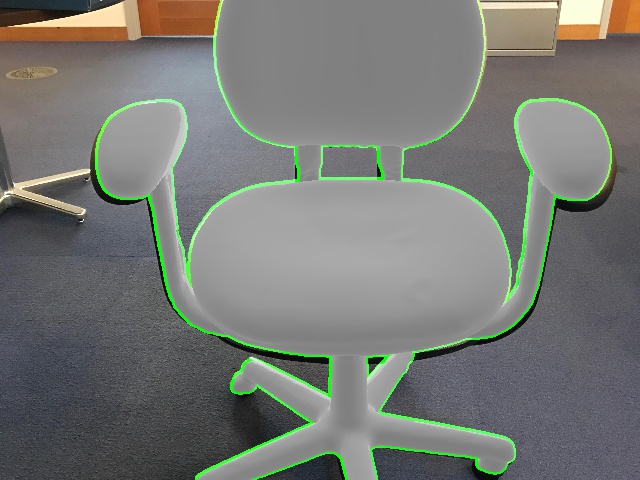}
        \end{minipage}
        \begin{minipage}{0.225\textwidth}
            \centering
            \includegraphics[width=\textwidth]{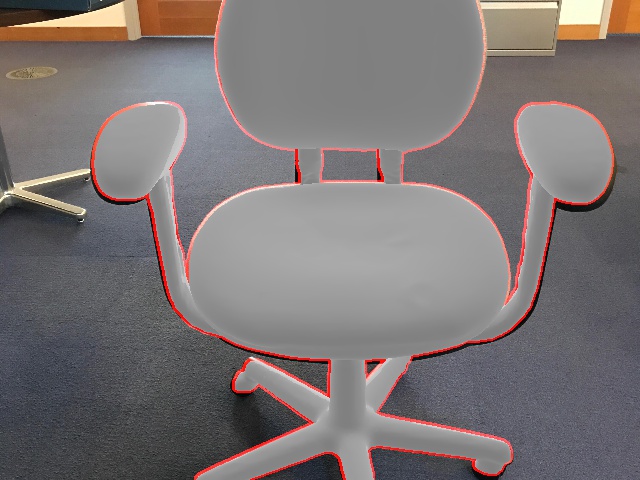}
        \end{minipage}\\[1mm]
        \begin{minipage}{0.225\textwidth}
            \centering
            \includegraphics[width=\textwidth]{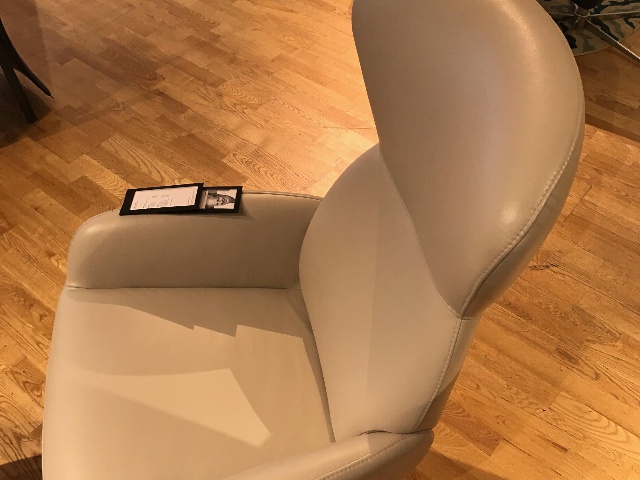}
        \end{minipage}
        \begin{minipage}{0.225\textwidth}
            \centering
            \includegraphics[width=\textwidth]{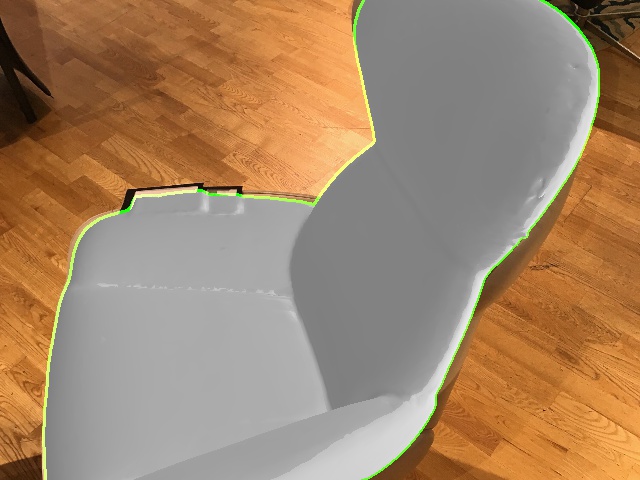}
        \end{minipage}
        \begin{minipage}{0.225\textwidth}
            \centering
            \includegraphics[width=\textwidth]{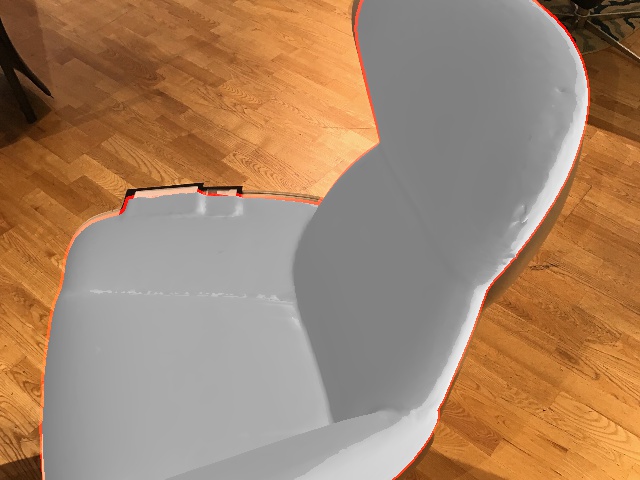}
        \end{minipage}\\[1mm]
        \begin{minipage}{0.225\textwidth}
            \centering
            \includegraphics[width=\textwidth]{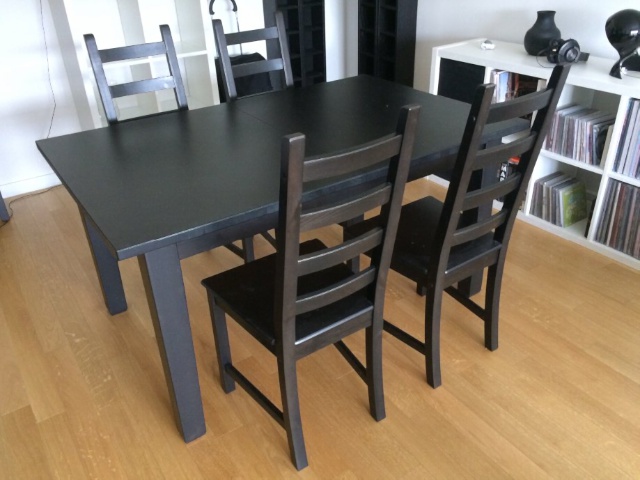}
        \end{minipage}
        \begin{minipage}{0.225\textwidth}
            \centering
            \includegraphics[width=\textwidth]{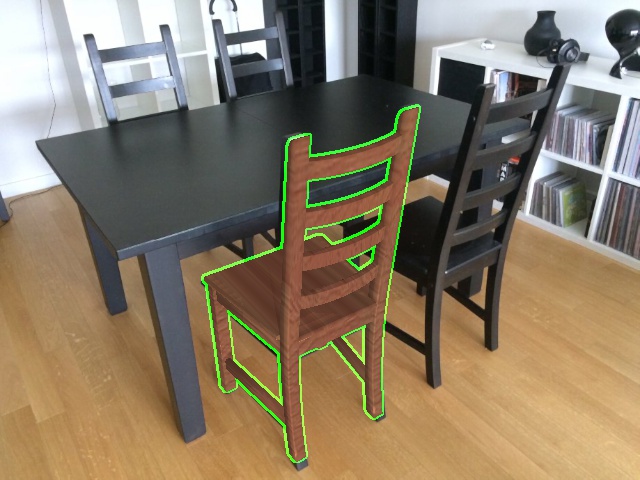}
        \end{minipage}
        \begin{minipage}{0.225\textwidth}
            \centering
            \includegraphics[width=\textwidth]{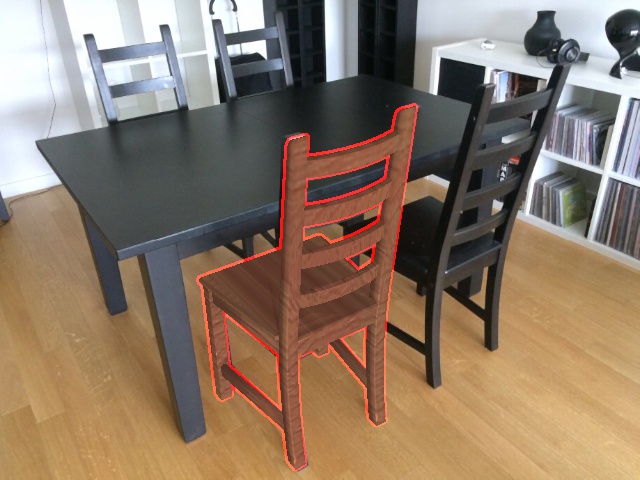}
        \end{minipage}\\[1mm]
    \caption{Qualitative results for Pix3D chairs - part 1.}
    \label{pix3d-chair-q-1}
\end{figure*}

\begin{figure*}[t]
    \centering
            \begin{minipage}{0.225\textwidth}
            {\small Input image}
            \centering
            \includegraphics[width=\textwidth]{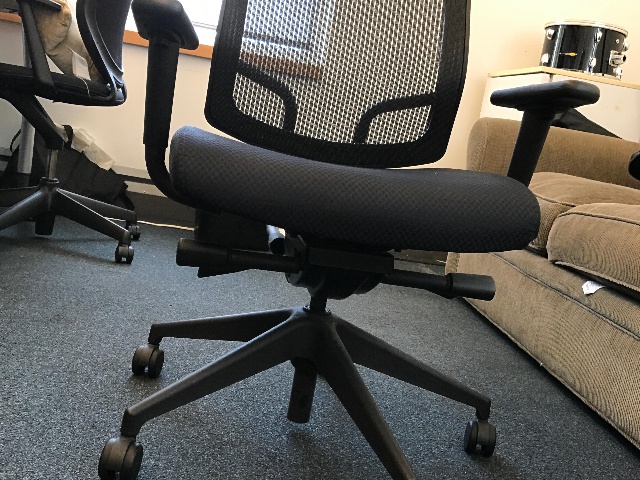}
        \end{minipage}
        \begin{minipage}{0.225\textwidth}
        {\small Ground truth}
            \centering
            \includegraphics[width=\textwidth]{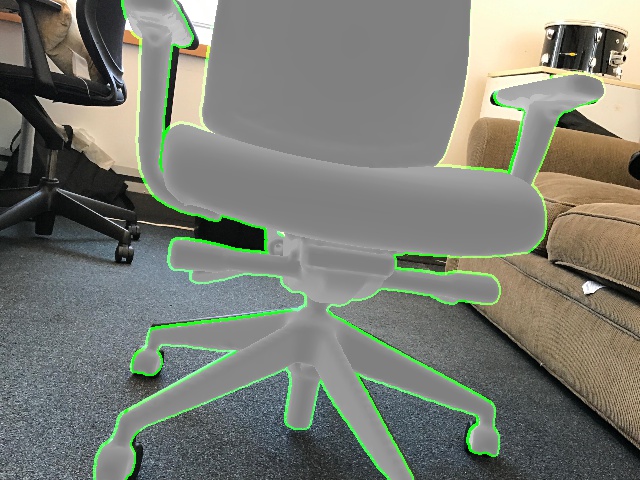}
        \end{minipage}
        \begin{minipage}{0.225\textwidth}
        {\small Our prediction}
            \centering
            \includegraphics[width=\textwidth]{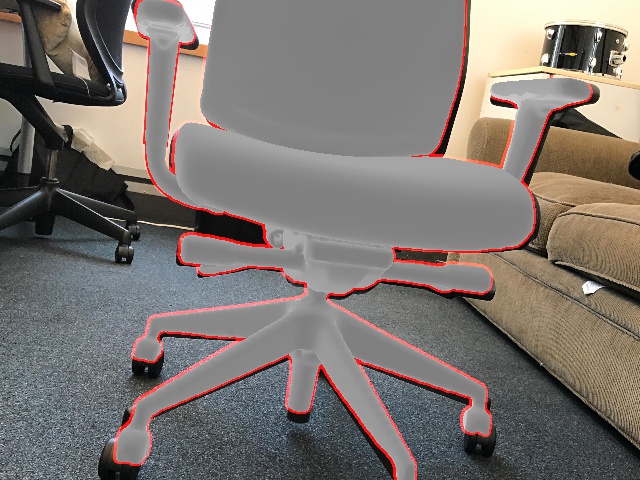}
        \end{minipage}\\[1mm]
        \begin{minipage}{0.225\textwidth}
            \centering
            \includegraphics[width=\textwidth]{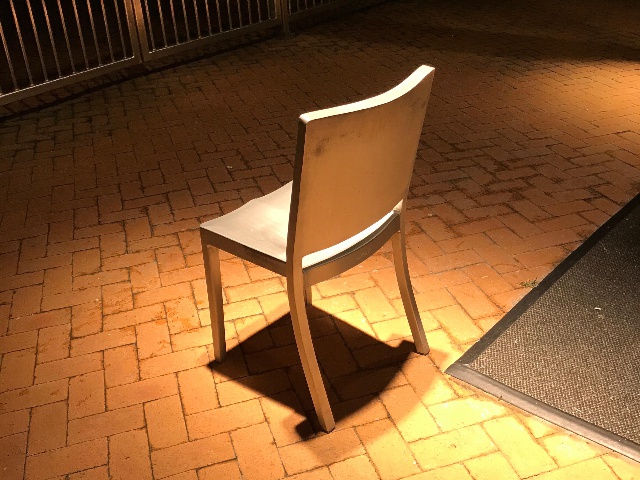}
        \end{minipage}
        \begin{minipage}{0.225\textwidth}
            \centering
            \includegraphics[width=\textwidth]{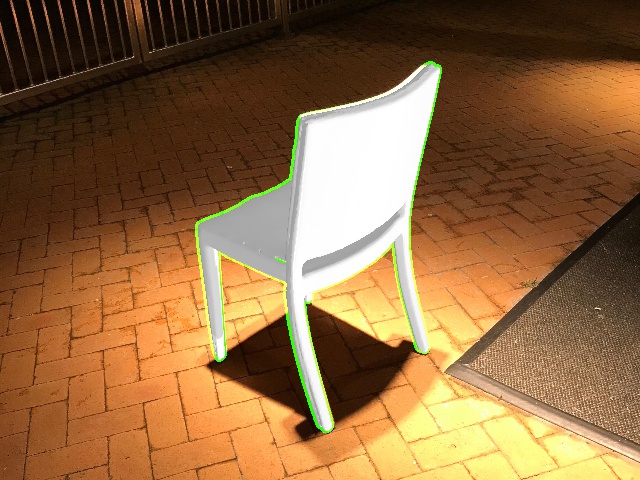}
        \end{minipage}
        \begin{minipage}{0.225\textwidth}
            \centering
            \includegraphics[width=\textwidth]{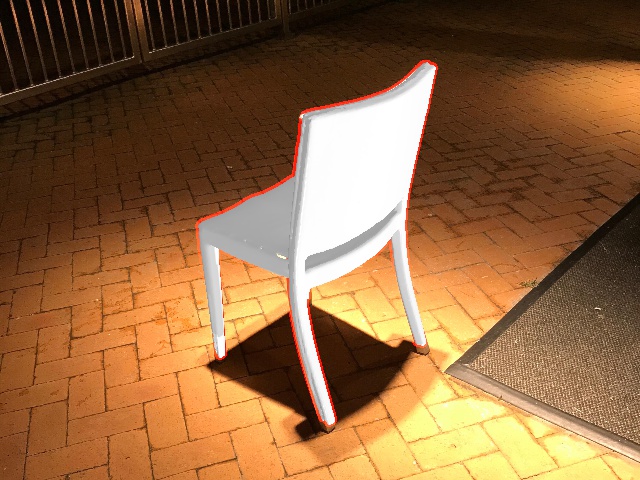}
        \end{minipage}\\[1mm]
        \begin{minipage}{0.225\textwidth}
            \centering
            \includegraphics[width=\textwidth]{figures/chairs/entry_132_input.jpeg}
        \end{minipage}
        \begin{minipage}{0.225\textwidth}
            \centering
            \includegraphics[width=\textwidth]{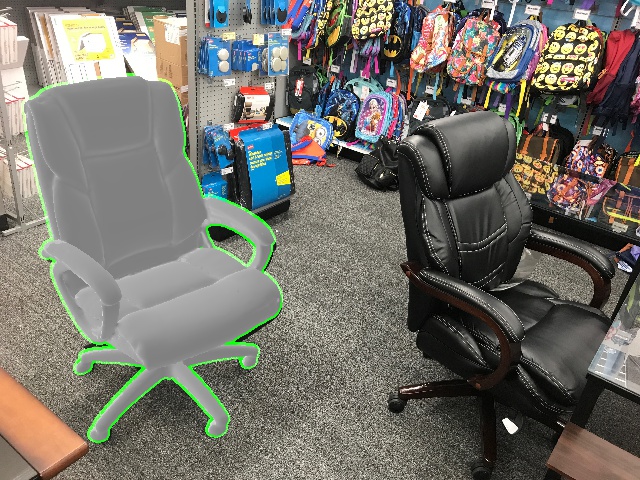}
        \end{minipage}
        \begin{minipage}{0.225\textwidth}
            \centering
            \includegraphics[width=\textwidth]{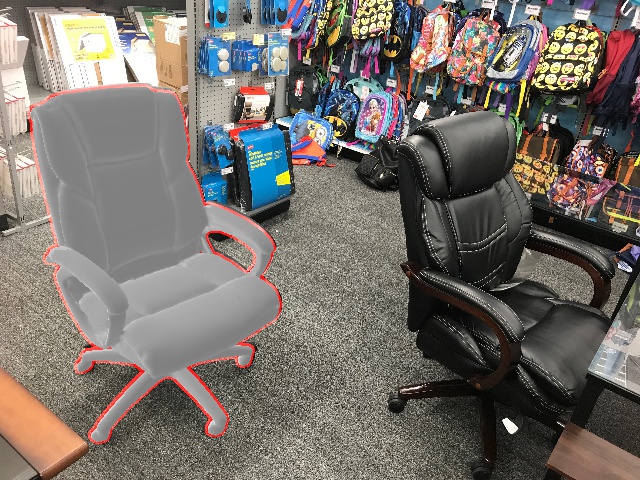}
        \end{minipage}\\[1mm]
        \begin{minipage}{0.225\textwidth}
            \centering
            \includegraphics[width=\textwidth]{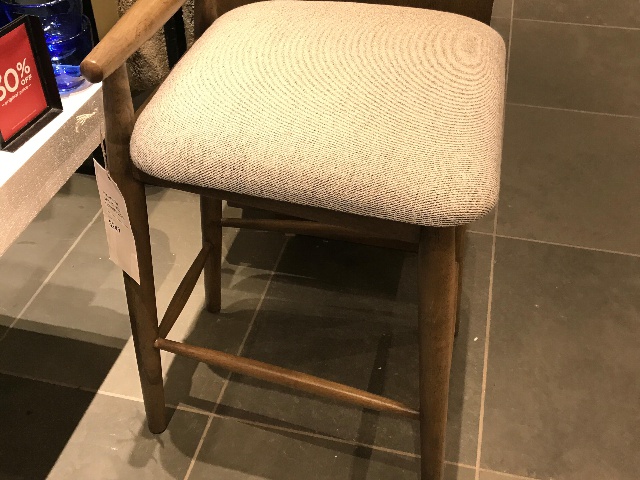}
        \end{minipage}
        \begin{minipage}{0.225\textwidth}
            \centering
            \includegraphics[width=\textwidth]{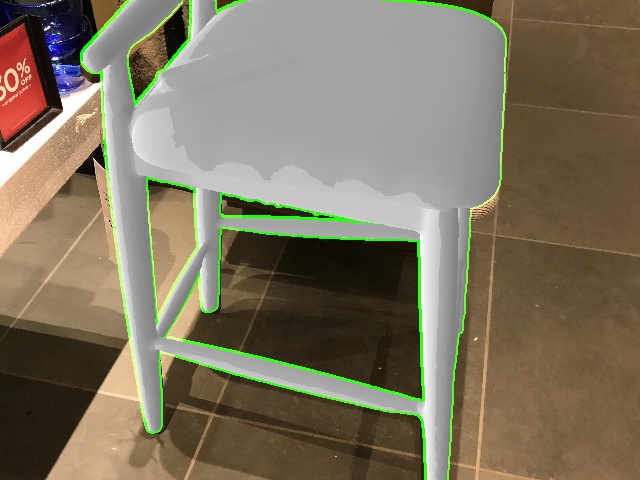}
        \end{minipage}
        \begin{minipage}{0.225\textwidth}
            \centering
            \includegraphics[width=\textwidth]{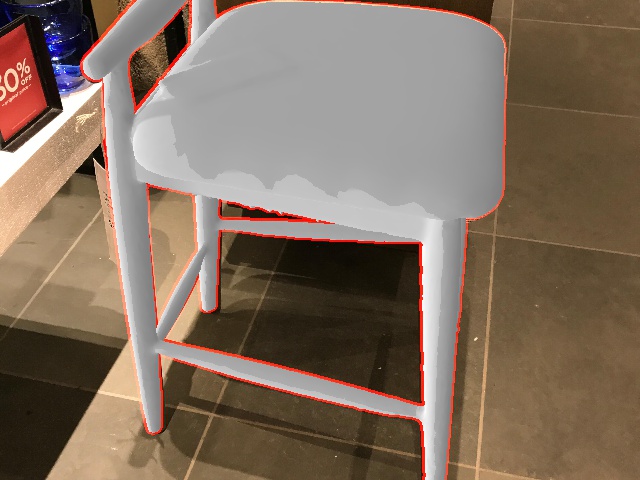}
        \end{minipage}\\[1mm]
        \begin{minipage}{0.225\textwidth}
            \centering
            \includegraphics[width=\textwidth]{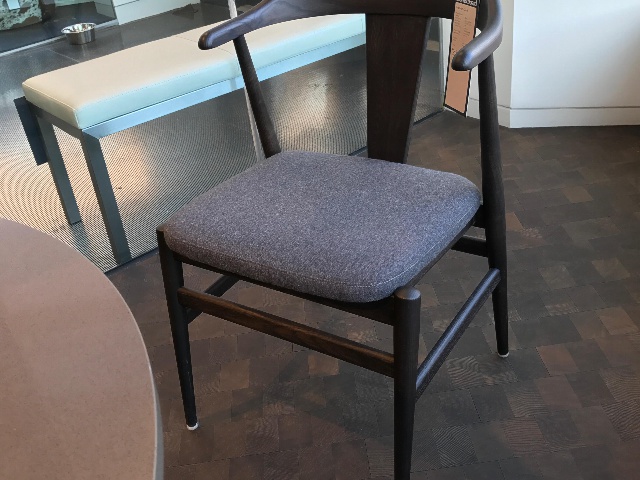}
        \end{minipage}
        \begin{minipage}{0.225\textwidth}
            \centering
            \includegraphics[width=\textwidth]{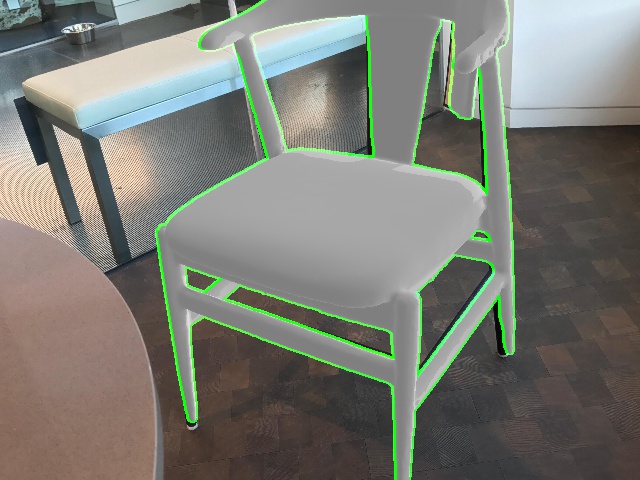}
        \end{minipage}
        \begin{minipage}{0.225\textwidth}
            \centering
            \includegraphics[width=\textwidth]{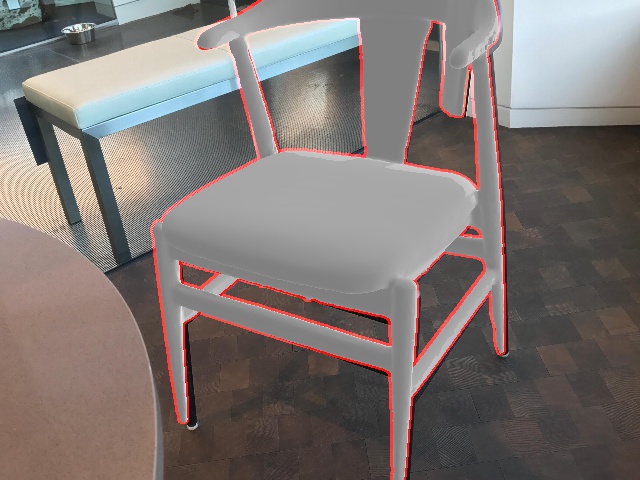}
        \end{minipage}\\[1mm]
                \begin{minipage}{0.225\textwidth}
            \centering
            \includegraphics[width=\textwidth]{figures/chairs/entry_461_input.jpeg}
        \end{minipage}
        \begin{minipage}{0.225\textwidth}
            \centering
            \includegraphics[width=\textwidth]{figures/chairs/entry_461_gt.jpeg}
        \end{minipage}
        \begin{minipage}{0.225\textwidth}
            \centering
            \includegraphics[width=\textwidth]{figures/chairs/entry_461_pred.jpeg}
        \end{minipage}\\[1mm]
                        \begin{minipage}{0.225\textwidth}
            \centering
            \includegraphics[width=\textwidth]{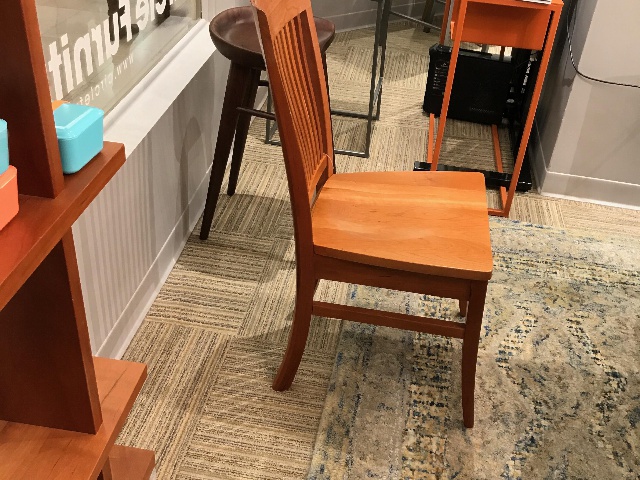}
        \end{minipage}
        \begin{minipage}{0.225\textwidth}
            \centering
            \includegraphics[width=\textwidth]{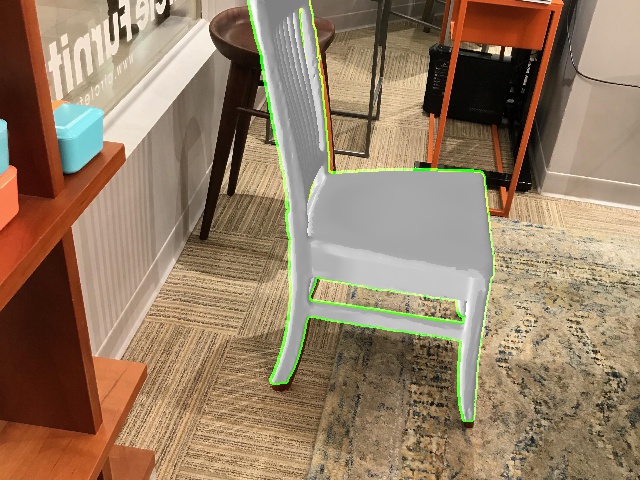}
        \end{minipage}
        \begin{minipage}{0.225\textwidth}
            \centering
            \includegraphics[width=\textwidth]{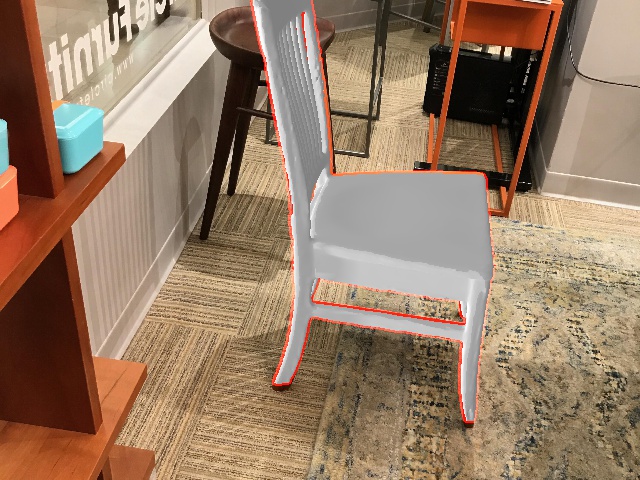}
        \end{minipage}\\[1mm]

    \caption{Qualitative results for Pix3D chairs - part 2.}
    \label{pix3d-chair-q-2}
\end{figure*}

\begin{figure*}[t]
    \centering
        \begin{minipage}{0.225\textwidth}
        {\small Input image}
            \centering
            \includegraphics[width=\textwidth]{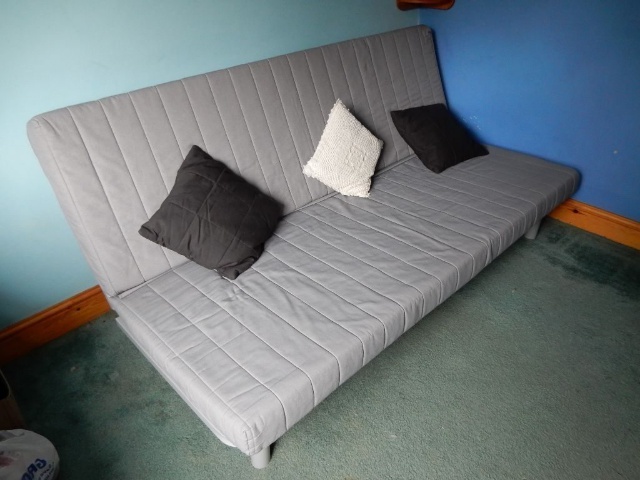}
        \end{minipage}
        \begin{minipage}{0.225\textwidth}
        {\small Ground truth}
            \centering
            \includegraphics[width=\textwidth]{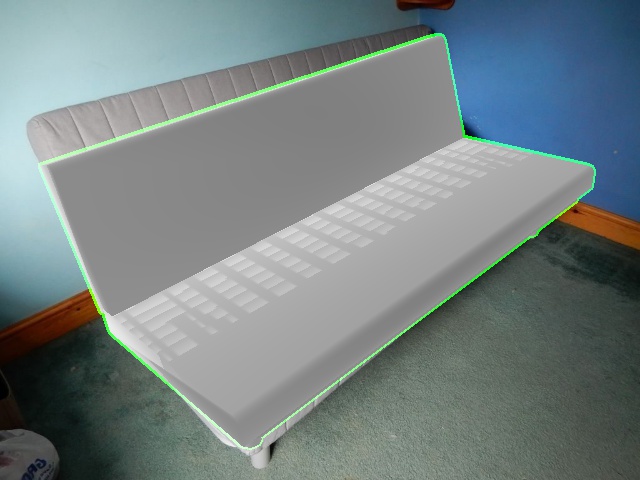}
        \end{minipage}
        \begin{minipage}{0.225\textwidth}
        {\small Our prediction}
            \centering
            \includegraphics[width=\textwidth]{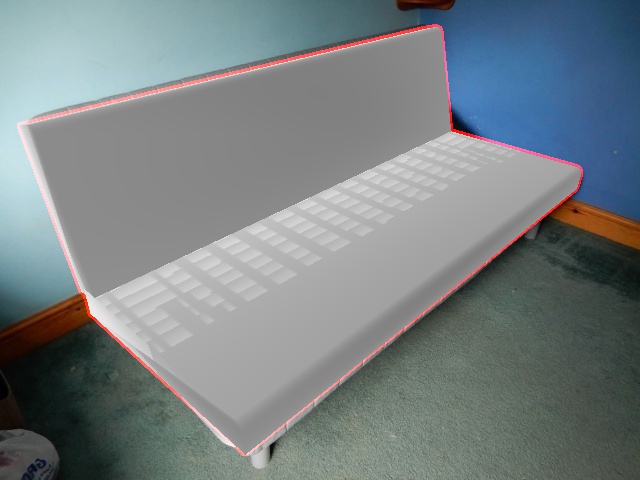}
        \end{minipage}\\[1mm]
        \begin{minipage}{0.225\textwidth}
            \centering
            \includegraphics[width=\textwidth]{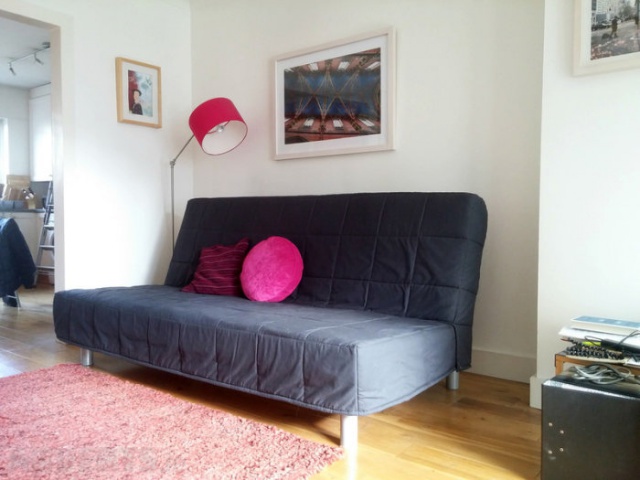}
        \end{minipage}
        \begin{minipage}{0.225\textwidth}
            \centering
            \includegraphics[width=\textwidth]{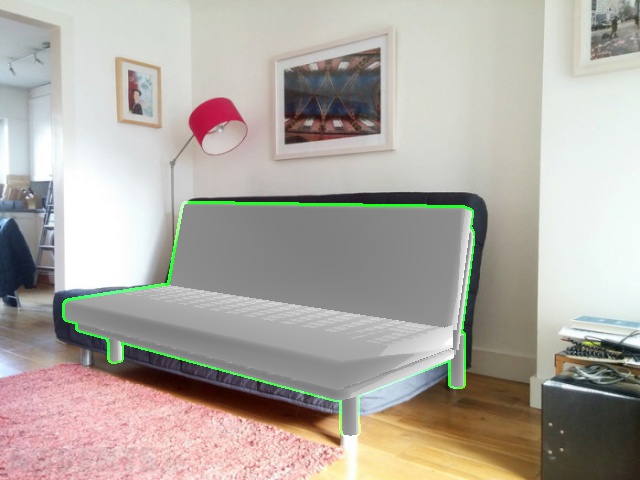}
        \end{minipage}
        \begin{minipage}{0.225\textwidth}
            \centering
            \includegraphics[width=\textwidth]{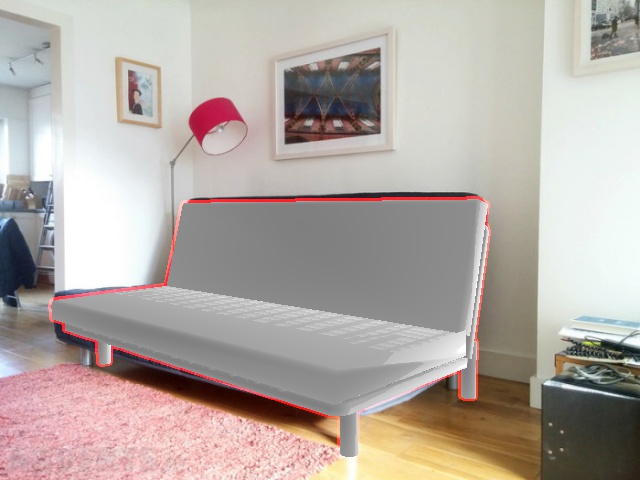}
        \end{minipage}\\[1mm]
        \begin{minipage}{0.225\textwidth}
            \centering
            \includegraphics[width=\textwidth]{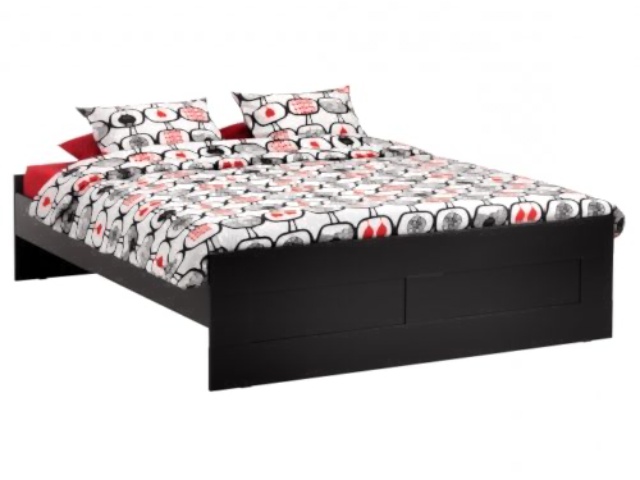}
        \end{minipage}
        \begin{minipage}{0.225\textwidth}
            \centering
            \includegraphics[width=\textwidth]{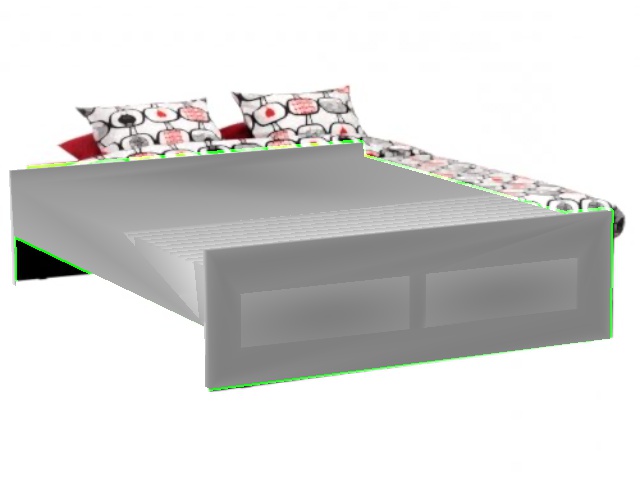}
        \end{minipage}
        \begin{minipage}{0.225\textwidth}
            \centering
            \includegraphics[width=\textwidth]{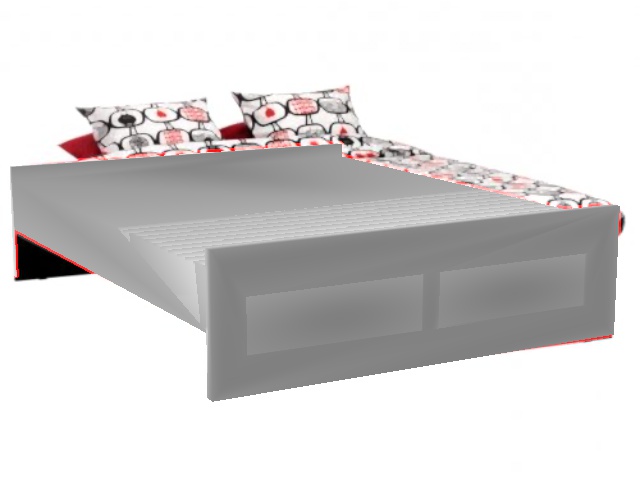}
        \end{minipage}\\[1mm]
            \begin{minipage}{0.225\textwidth}
            \centering
            \includegraphics[width=\textwidth]{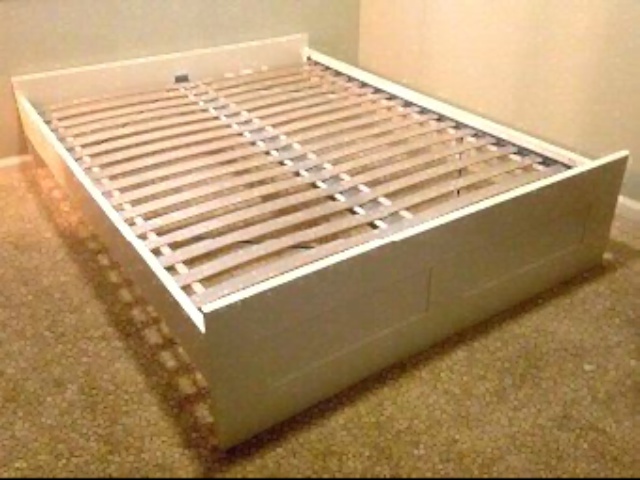}
        \end{minipage}
        \begin{minipage}{0.225\textwidth}
            \centering
            \includegraphics[width=\textwidth]{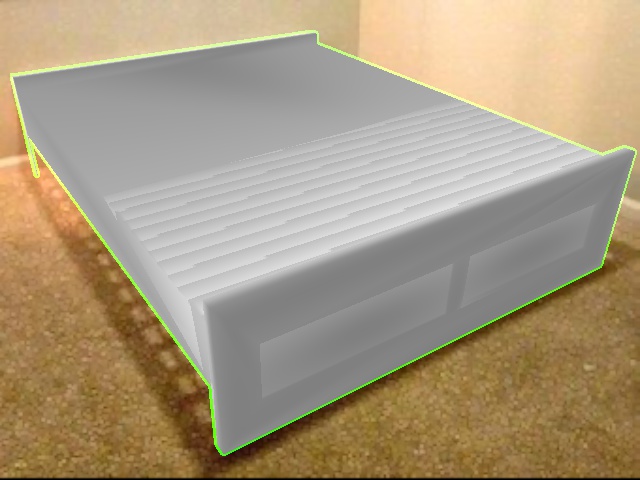}
        \end{minipage}
        \begin{minipage}{0.225\textwidth}
            \centering
            \includegraphics[width=\textwidth]{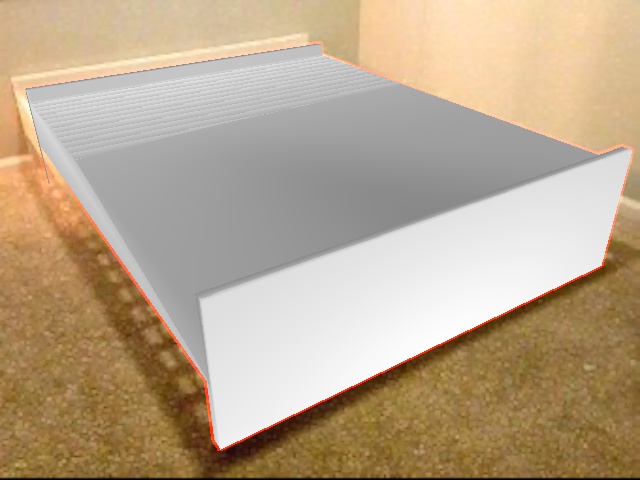}
        \end{minipage}\\[1mm]
                    \begin{minipage}{0.225\textwidth}
            \centering
            \includegraphics[width=\textwidth]{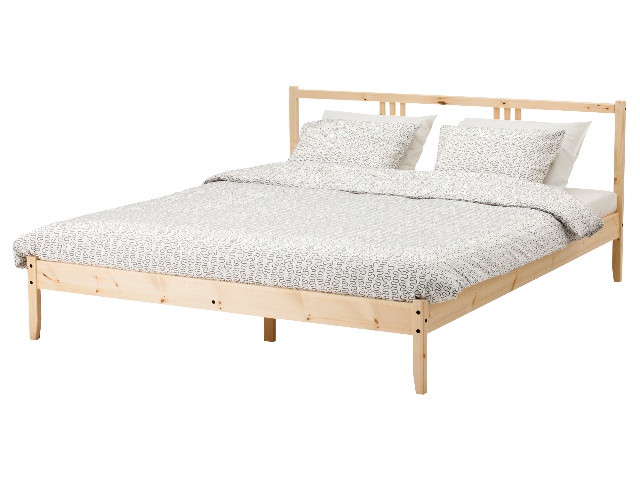}
        \end{minipage}
        \begin{minipage}{0.225\textwidth}
            \centering
            \includegraphics[width=\textwidth]{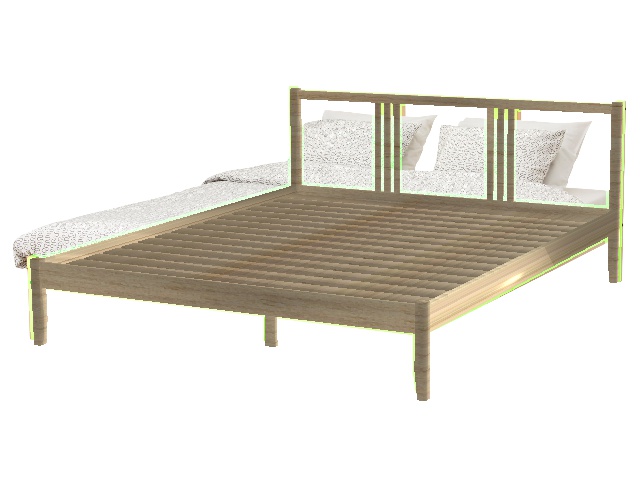}
        \end{minipage}
        \begin{minipage}{0.225\textwidth}
            \centering
            \includegraphics[width=\textwidth]{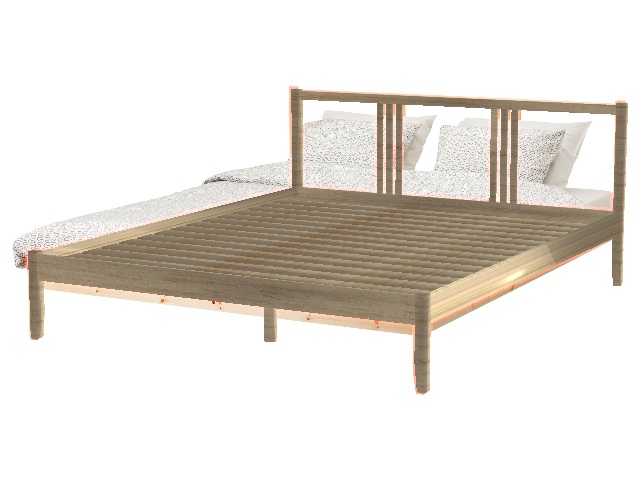}
        \end{minipage}\\[1mm]
        \begin{minipage}{0.225\textwidth}
            \centering
            \includegraphics[width=\textwidth]{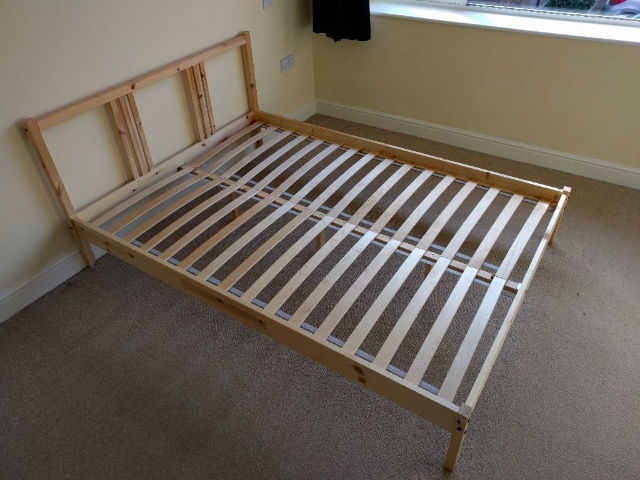}
        \end{minipage}
        \begin{minipage}{0.225\textwidth}
            \centering
            \includegraphics[width=\textwidth]{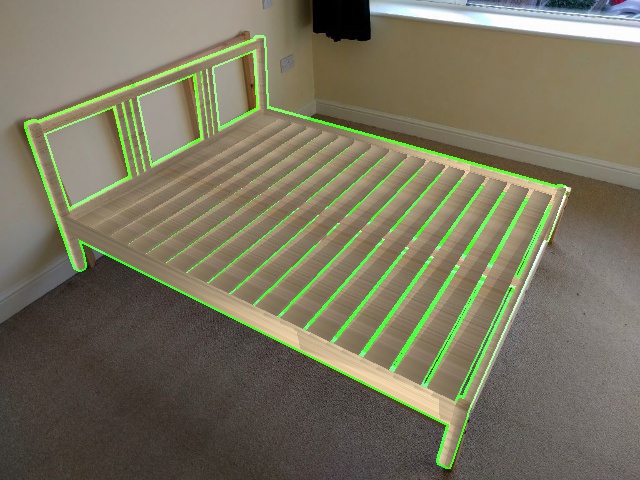}
        \end{minipage}
        \begin{minipage}{0.225\textwidth}
            \centering
            \includegraphics[width=\textwidth]{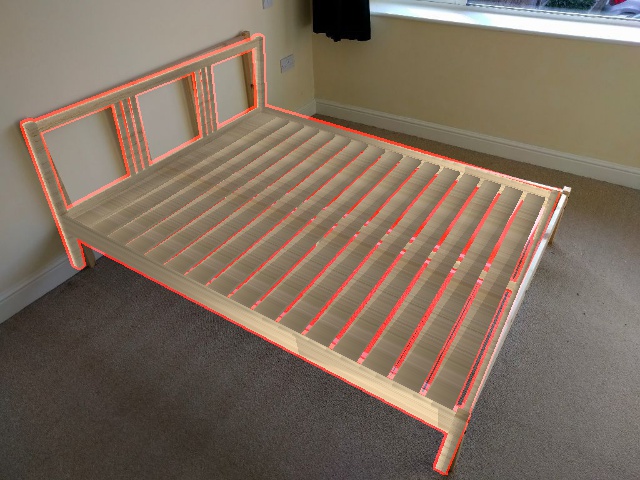}
        \end{minipage}\\[1mm]
        \begin{minipage}{0.225\textwidth}
            \centering
            \includegraphics[width=\textwidth]{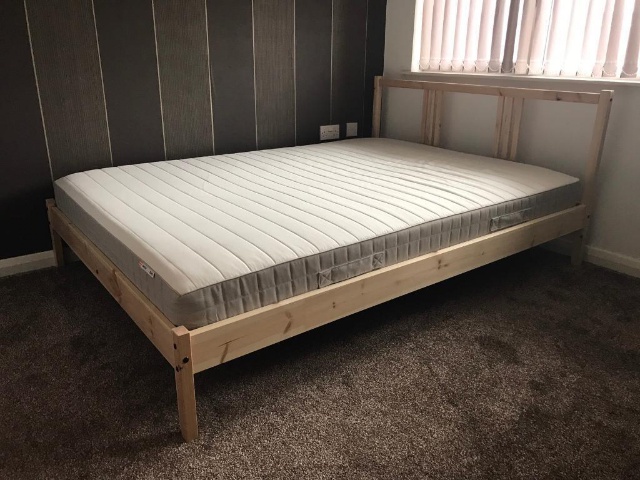}
        \end{minipage}
        \begin{minipage}{0.225\textwidth}
            \centering
            \includegraphics[width=\textwidth]{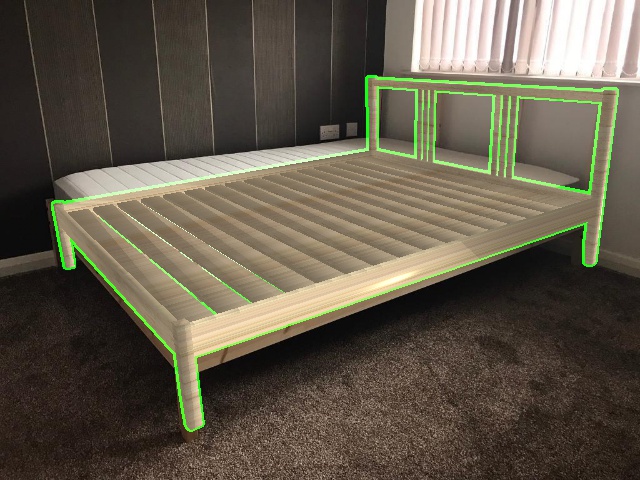}
        \end{minipage}
        \begin{minipage}{0.225\textwidth}
            \centering
            \includegraphics[width=\textwidth]{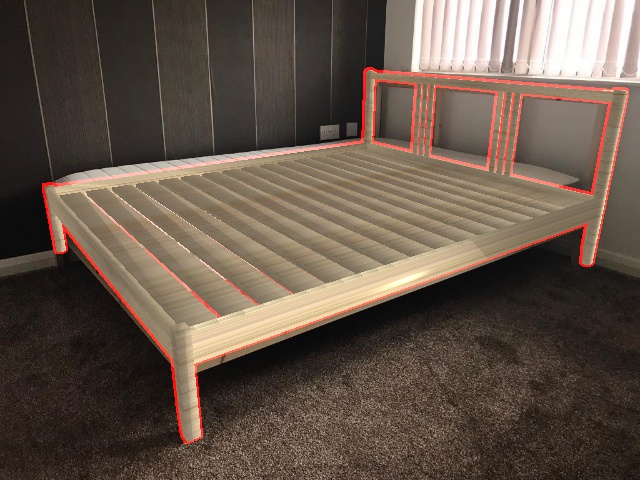}
        \end{minipage}\\[1mm]
    \caption{Qualitative results for Pix3D beds - part 1.}
    \label{pix3d-beds-q-1}
\end{figure*}

\begin{figure*}[t]
    \centering
        \begin{minipage}{0.225\textwidth}
        {\small Input image}
            \centering
            \includegraphics[width=\textwidth]{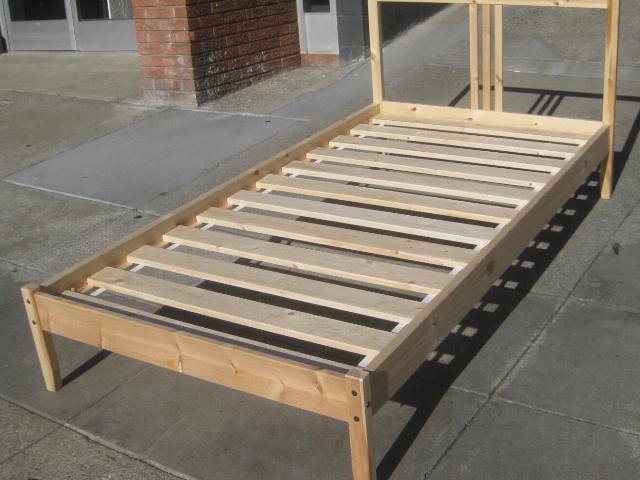}
        \end{minipage}
        \begin{minipage}{0.225\textwidth}
        {\small Ground truth}
            \centering
            \includegraphics[width=\textwidth]{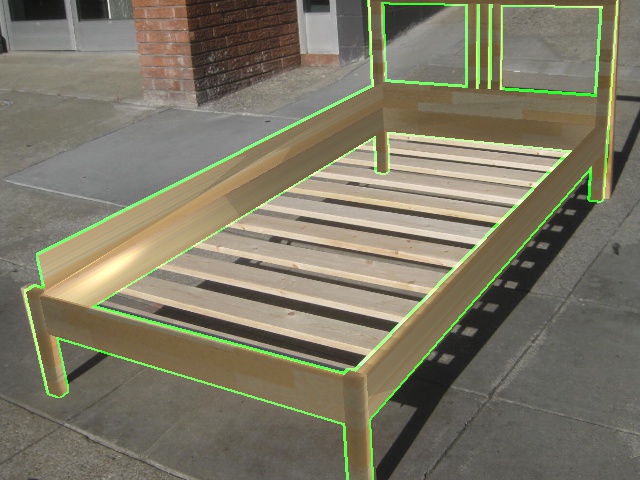}
        \end{minipage}
        \begin{minipage}{0.225\textwidth}
        {\small Our prediction}
            \centering
            \includegraphics[width=\textwidth]{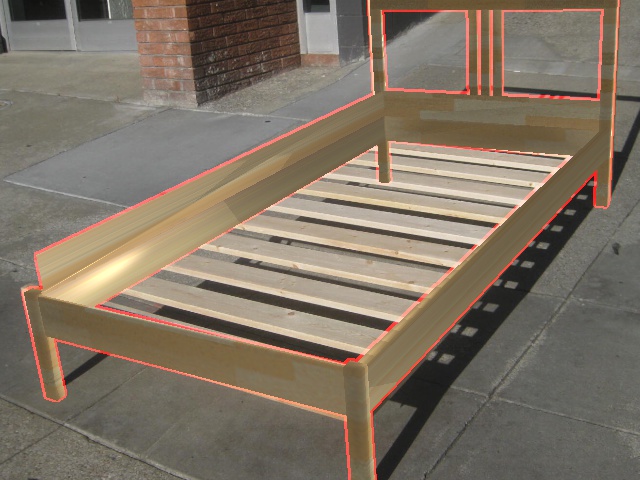}
        \end{minipage}\\[1mm]
        \begin{minipage}{0.225\textwidth}
            \centering
            \includegraphics[width=\textwidth]{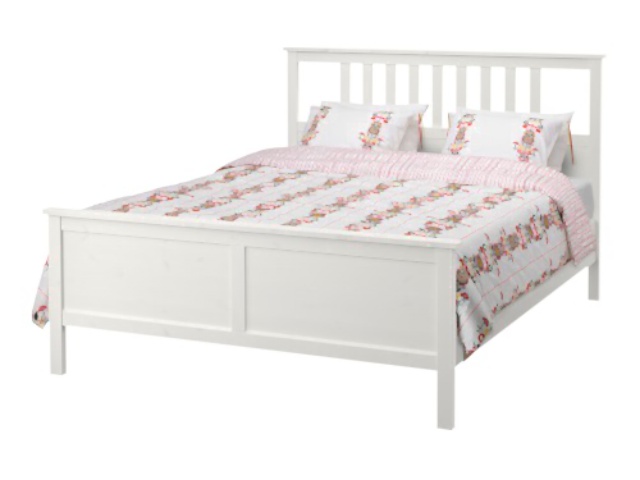}
        \end{minipage}
        \begin{minipage}{0.225\textwidth}
            \centering
            \includegraphics[width=\textwidth]{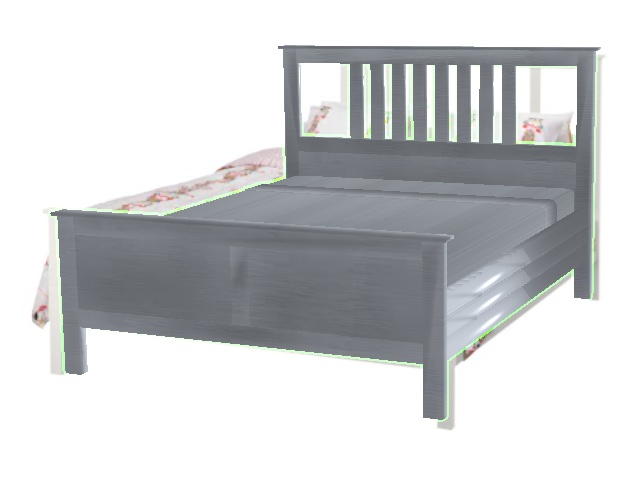}
        \end{minipage}
        \begin{minipage}{0.225\textwidth}
            \centering
            \includegraphics[width=\textwidth]{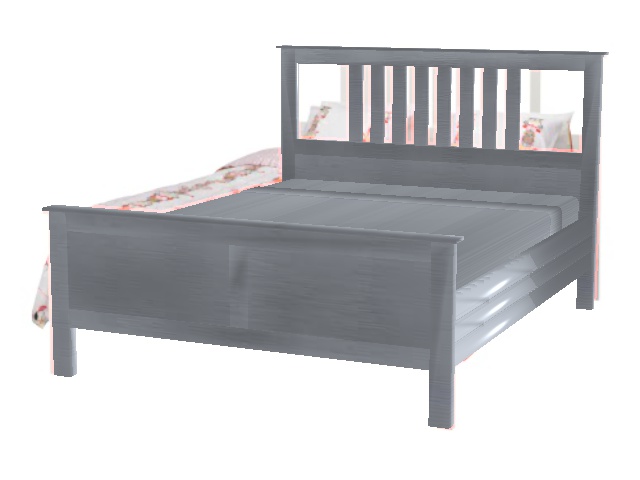}
        \end{minipage}\\[1mm]
        \begin{minipage}{0.225\textwidth}
            \centering
            \includegraphics[width=\textwidth]{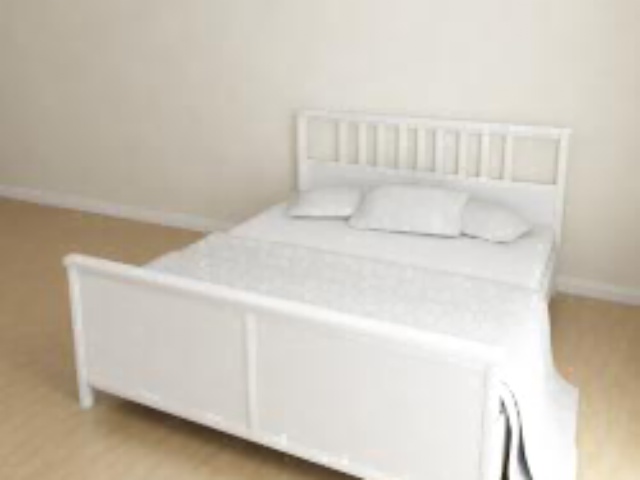}
        \end{minipage}
        \begin{minipage}{0.225\textwidth}
            \centering
            \includegraphics[width=\textwidth]{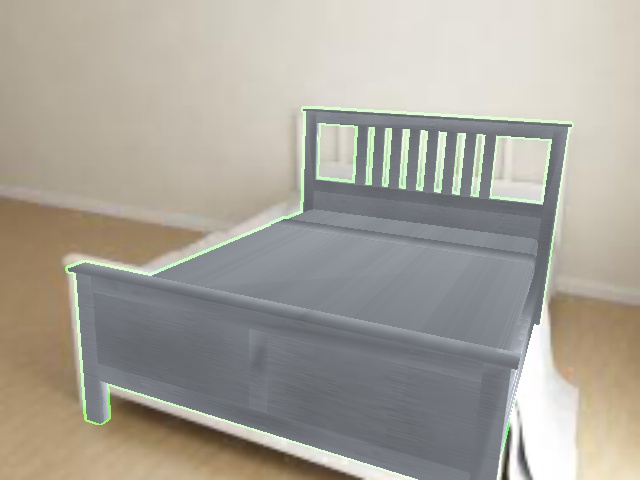}
        \end{minipage}
        \begin{minipage}{0.225\textwidth}
            \centering
            \includegraphics[width=\textwidth]{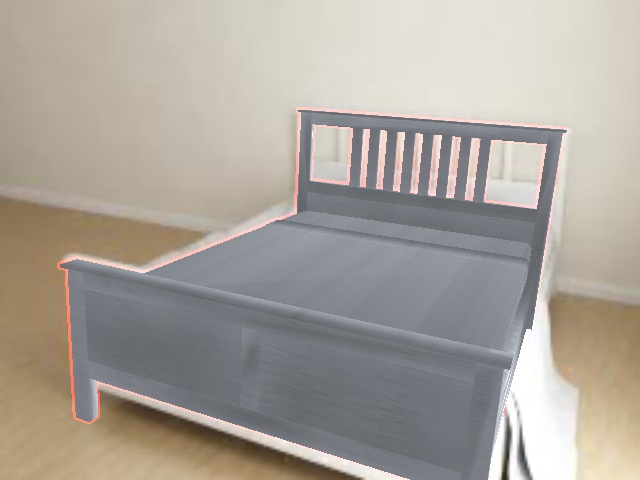}
        \end{minipage}\\[1mm]
            \begin{minipage}{0.225\textwidth}
            \centering
            \includegraphics[width=\textwidth]{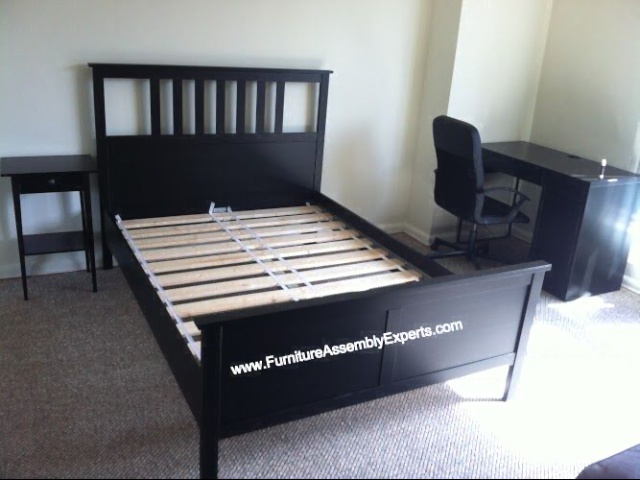}
        \end{minipage}
        \begin{minipage}{0.225\textwidth}
            \centering
            \includegraphics[width=\textwidth]{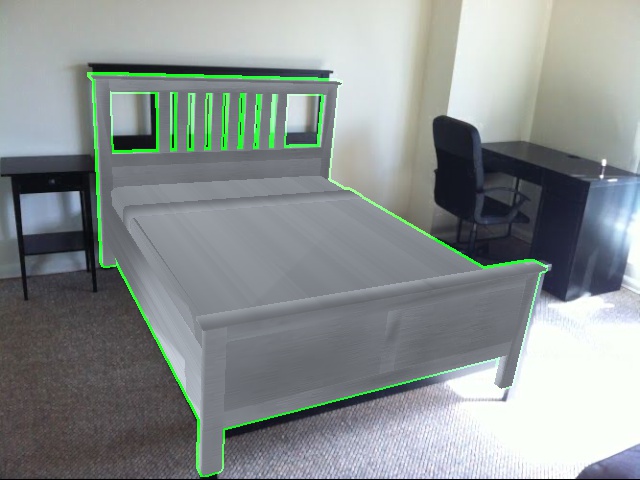}
        \end{minipage}
        \begin{minipage}{0.225\textwidth}
            \centering
            \includegraphics[width=\textwidth]{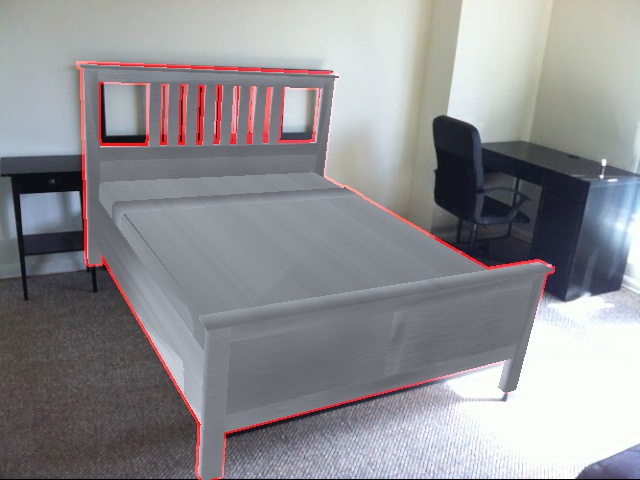}
        \end{minipage}\\[1mm]
                    \begin{minipage}{0.225\textwidth}
            \centering
            \includegraphics[width=\textwidth]{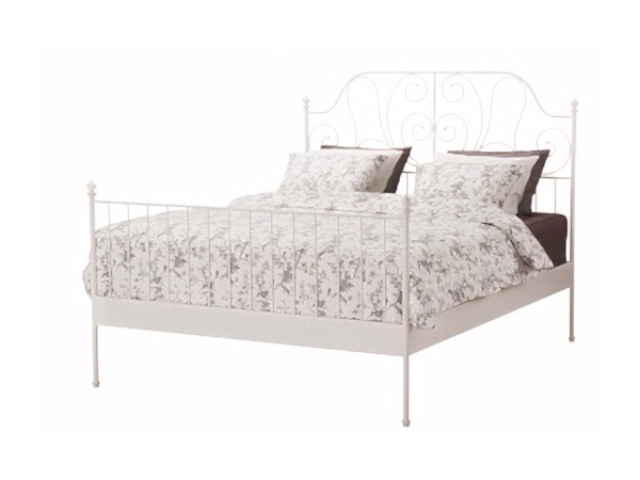}
        \end{minipage}
        \begin{minipage}{0.225\textwidth}
            \centering
            \includegraphics[width=\textwidth]{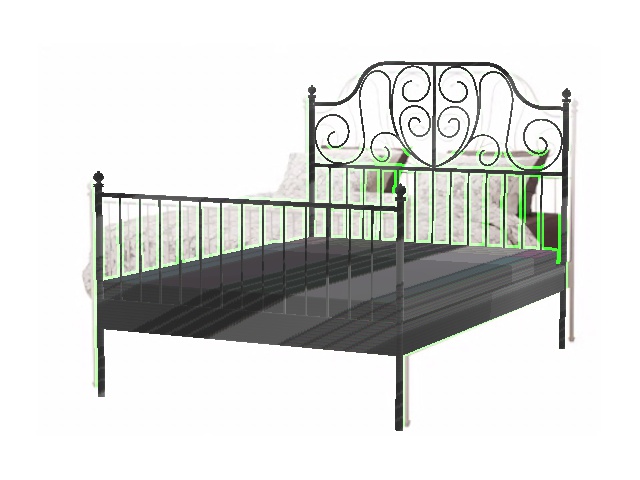}
        \end{minipage}
        \begin{minipage}{0.225\textwidth}
            \centering
            \includegraphics[width=\textwidth]{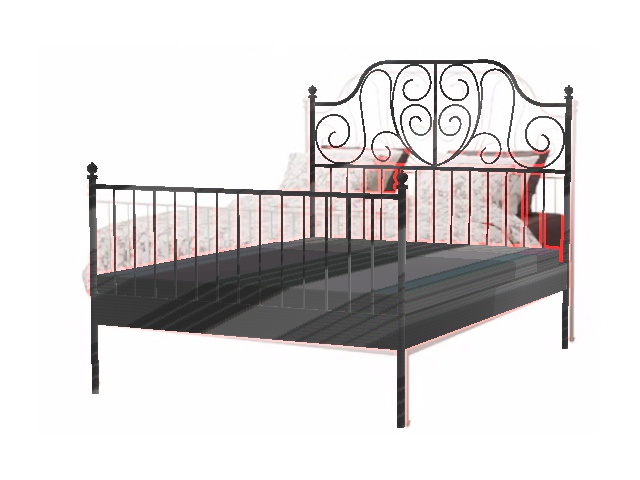}
        \end{minipage}\\[1mm]
        \begin{minipage}{0.225\textwidth}
            \centering
            \includegraphics[width=\textwidth]{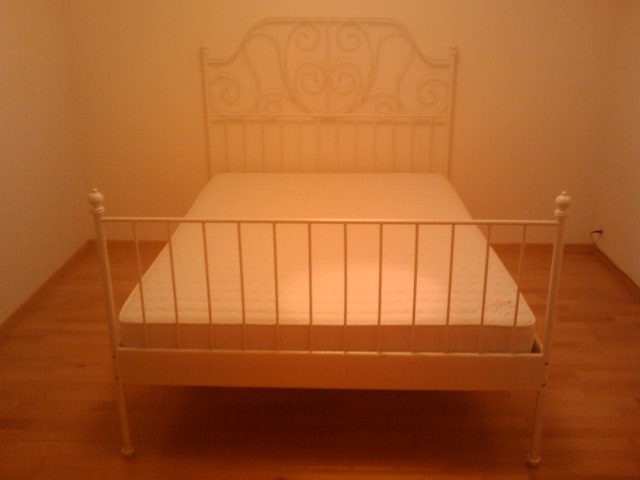}
        \end{minipage}
        \begin{minipage}{0.225\textwidth}
            \centering
            \includegraphics[width=\textwidth]{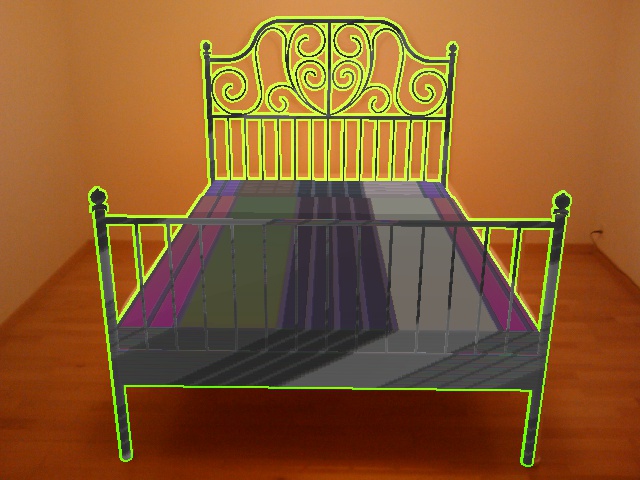}
        \end{minipage}
        \begin{minipage}{0.225\textwidth}
            \centering
            \includegraphics[width=\textwidth]{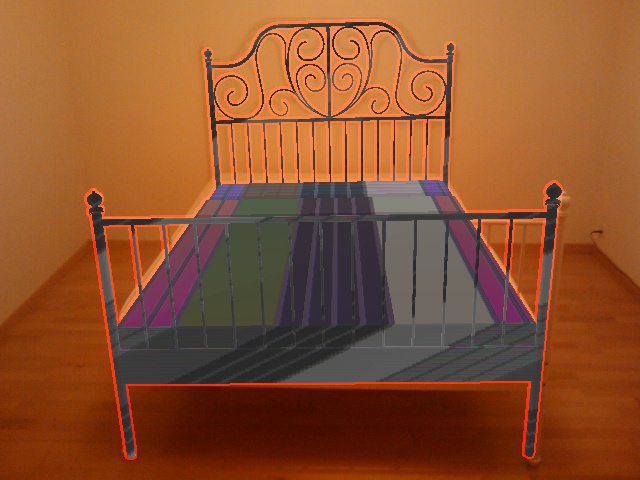}
        \end{minipage}\\[1mm]
        \begin{minipage}{0.225\textwidth}
            \centering
            \includegraphics[width=\textwidth]{figures/beds/entry_179_input.jpeg}
        \end{minipage}
        \begin{minipage}{0.225\textwidth}
            \centering
            \includegraphics[width=\textwidth]{figures/beds/entry_179_gt.jpeg}
        \end{minipage}
        \begin{minipage}{0.225\textwidth}
            \centering
            \includegraphics[width=\textwidth]{figures/beds/entry_179_pred.jpeg}
        \end{minipage}\\[1mm]
    \caption{Qualitative results for Pix3D beds - part 2.}
    \label{pix3d-beds-q-2}
\end{figure*}

\begin{figure*}[t]
    \centering
        \begin{minipage}{0.225\textwidth}
        {\small Input image}
            \centering
            \includegraphics[width=\textwidth]{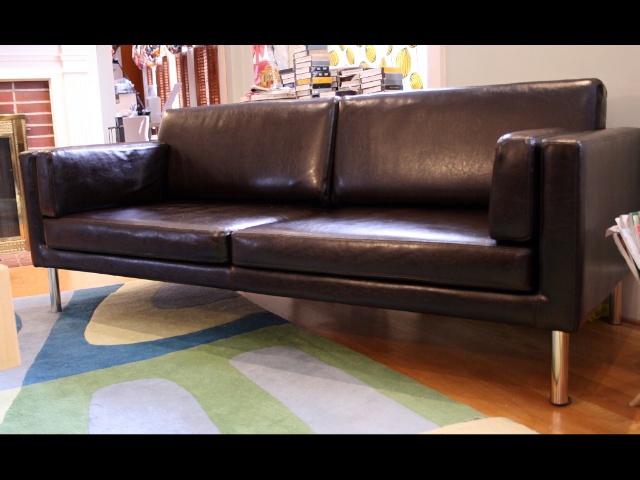}
        \end{minipage}
        \begin{minipage}{0.225\textwidth}
        {\small Ground truth}
            \centering
            \includegraphics[width=\textwidth]{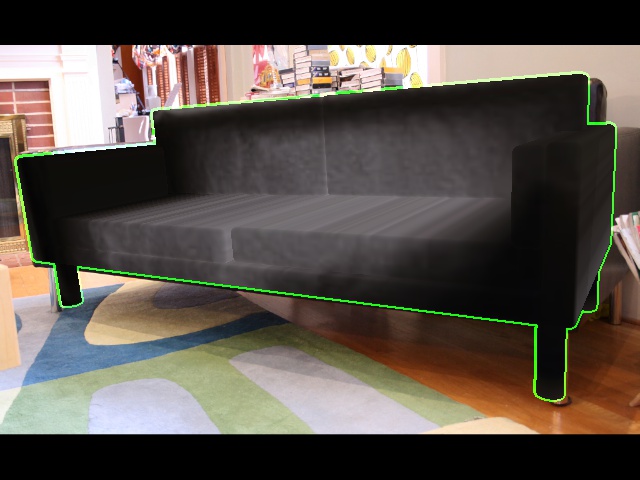}
        \end{minipage}
        \begin{minipage}{0.225\textwidth}
        {\small Our prediction}
            \centering
            \includegraphics[width=\textwidth]{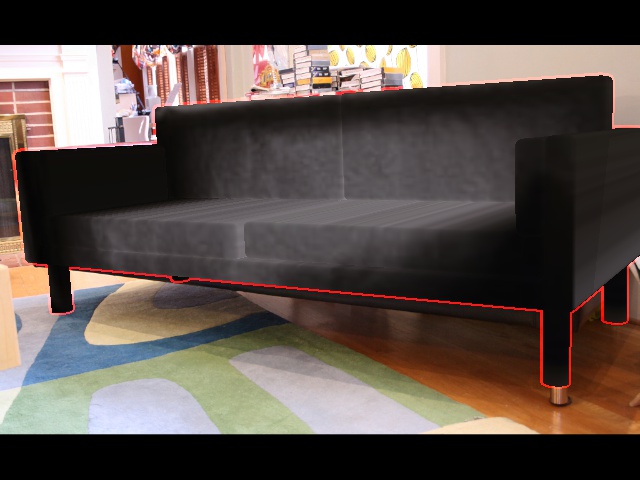}
        \end{minipage}\\[1mm]
        \begin{minipage}{0.225\textwidth}
            \centering
            \includegraphics[width=\textwidth]{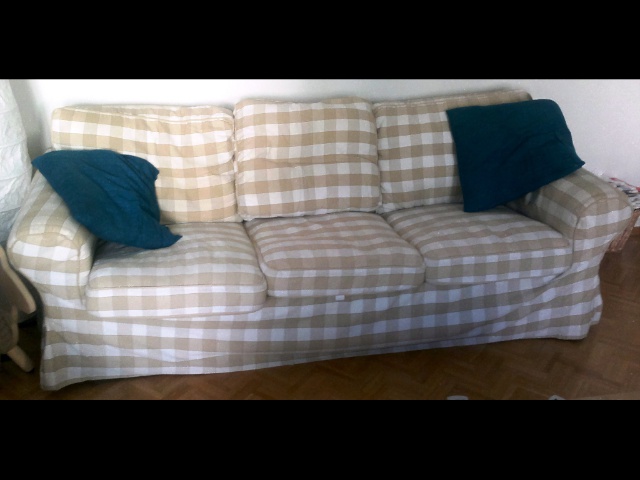}
        \end{minipage}
        \begin{minipage}{0.225\textwidth}
            \centering
            \includegraphics[width=\textwidth]{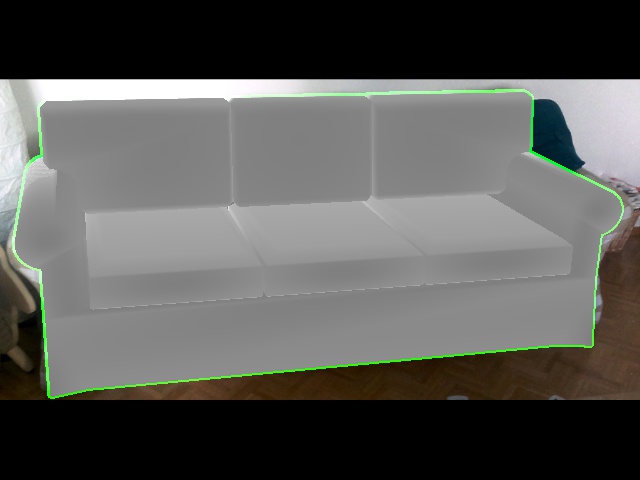}
        \end{minipage}
        \begin{minipage}{0.225\textwidth}
            \centering
            \includegraphics[width=\textwidth]{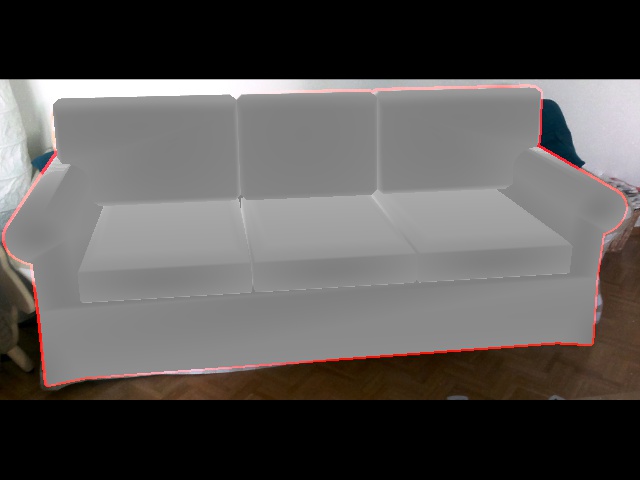}
        \end{minipage}\\[1mm]
        \begin{minipage}{0.225\textwidth}
            \centering
            \includegraphics[width=\textwidth]{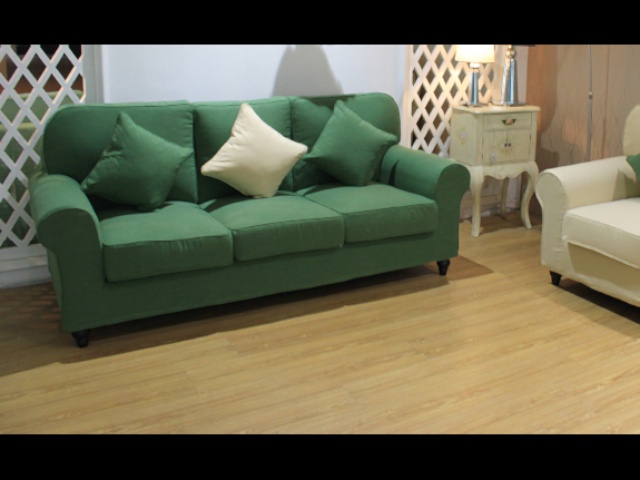}
        \end{minipage}
        \begin{minipage}{0.225\textwidth}
            \centering
            \includegraphics[width=\textwidth]{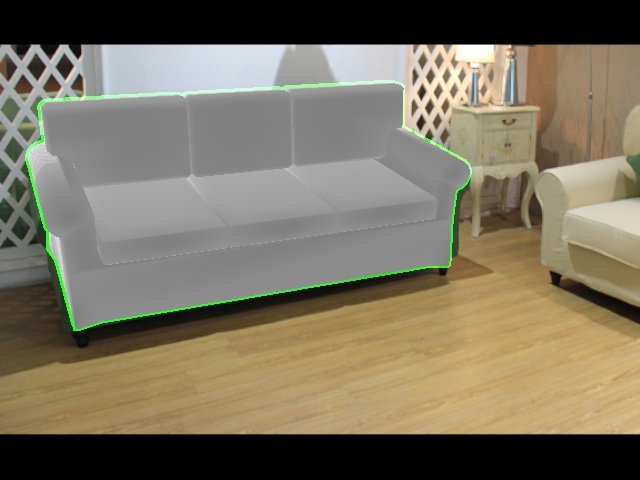}
        \end{minipage}
        \begin{minipage}{0.225\textwidth}
            \centering
            \includegraphics[width=\textwidth]{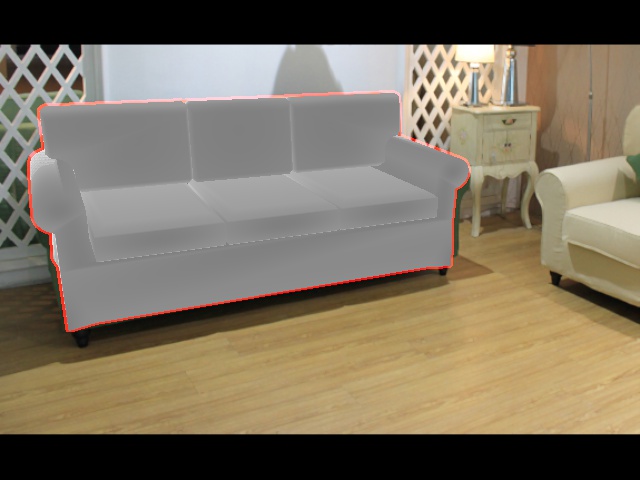}
        \end{minipage}\\[1mm]
        \begin{minipage}{0.225\textwidth}
            \centering
            \includegraphics[width=\textwidth]{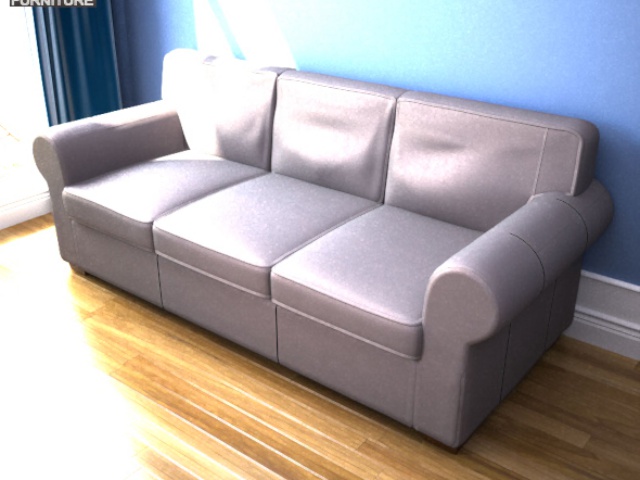}
        \end{minipage}
        \begin{minipage}{0.225\textwidth}
            \centering
            \includegraphics[width=\textwidth]{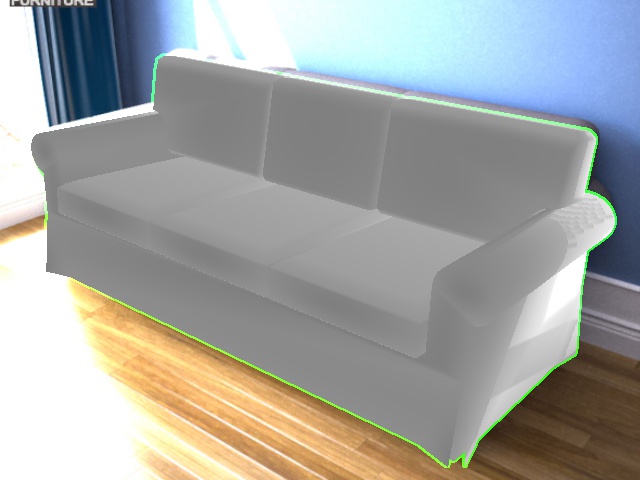}
        \end{minipage}
        \begin{minipage}{0.225\textwidth}
            \centering
            \includegraphics[width=\textwidth]{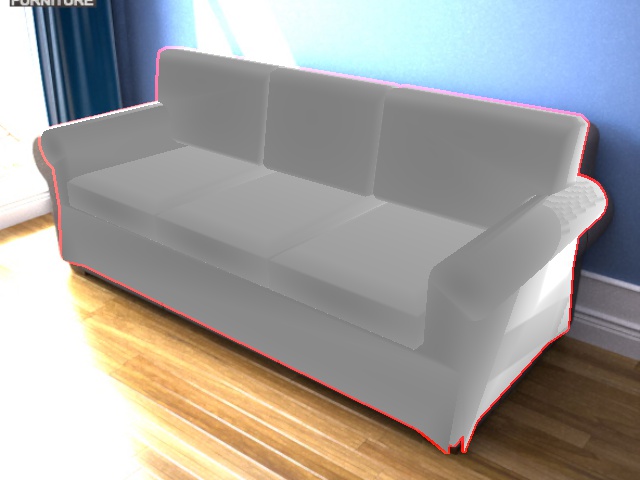}
        \end{minipage}\\[1mm]
        \begin{minipage}{0.225\textwidth}
            \centering
            \includegraphics[width=\textwidth]{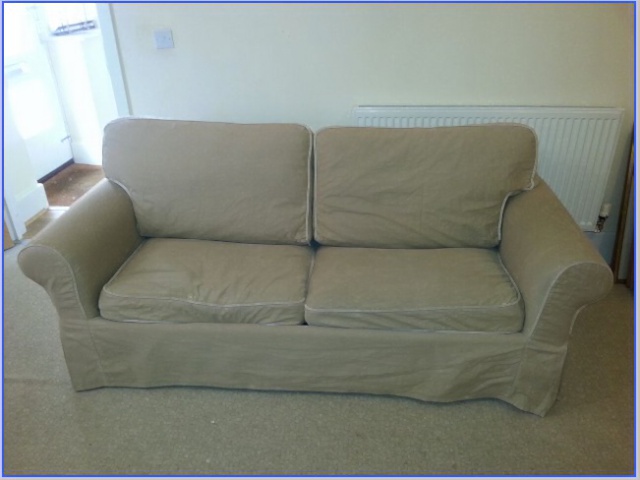}
        \end{minipage}
        \begin{minipage}{0.225\textwidth}
            \centering
            \includegraphics[width=\textwidth]{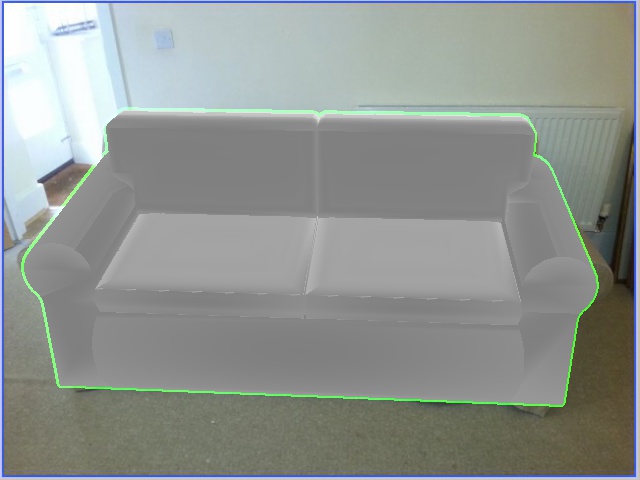}
        \end{minipage}
        \begin{minipage}{0.225\textwidth}
            \centering
            \includegraphics[width=\textwidth]{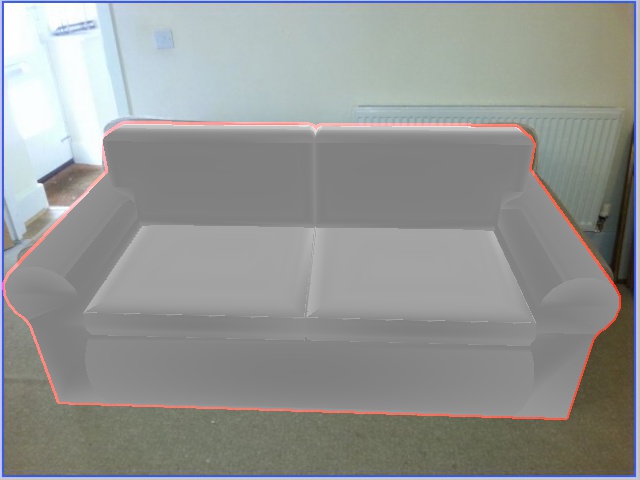}
        \end{minipage}\\[1mm]
        \begin{minipage}{0.225\textwidth}
            \centering
            \includegraphics[width=\textwidth]{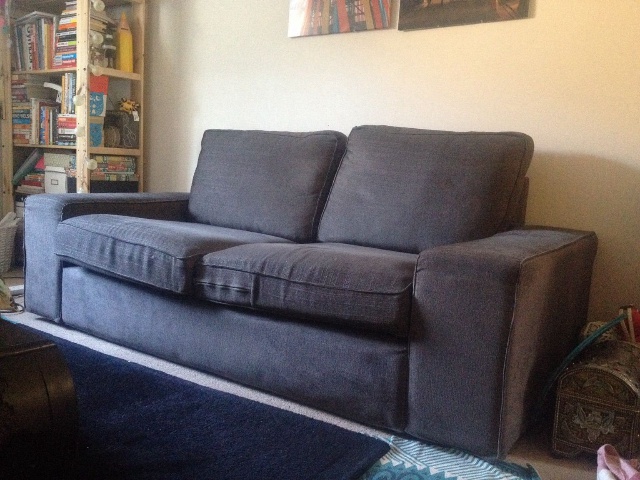}
        \end{minipage}
        \begin{minipage}{0.225\textwidth}
            \centering
            \includegraphics[width=\textwidth]{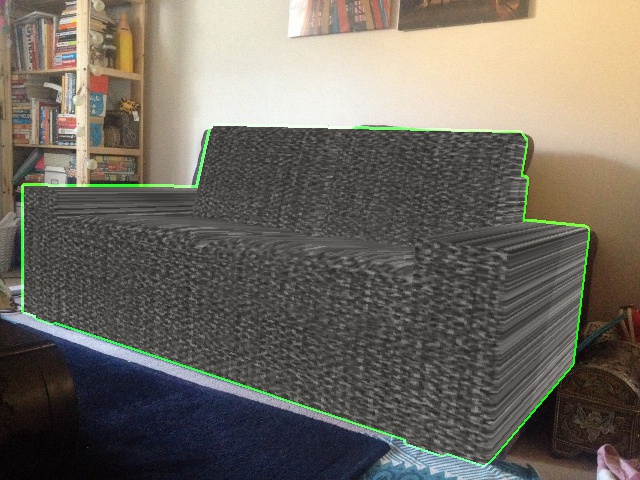}
        \end{minipage}
        \begin{minipage}{0.225\textwidth}
            \centering
            \includegraphics[width=\textwidth]{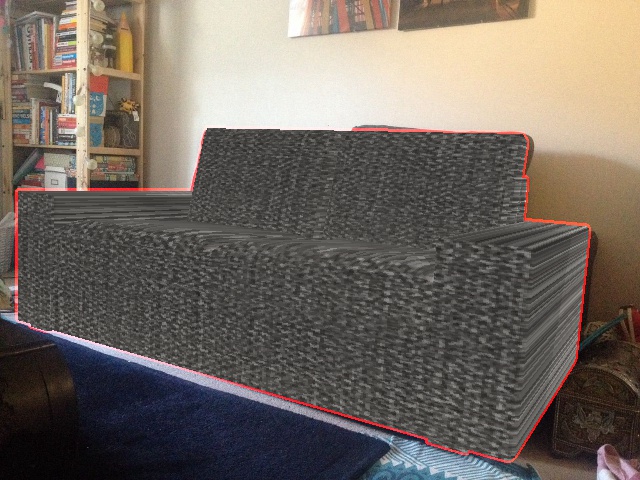}
        \end{minipage}\\[1mm]
        \begin{minipage}{0.225\textwidth}
            \centering
            \includegraphics[width=\textwidth]{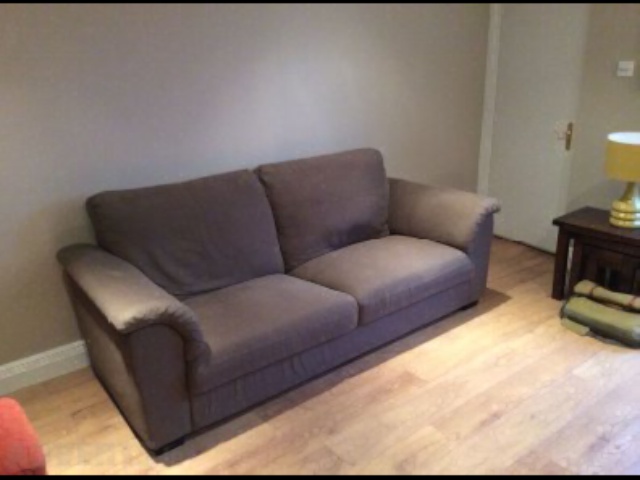}
        \end{minipage}
        \begin{minipage}{0.225\textwidth}
            \centering
            \includegraphics[width=\textwidth]{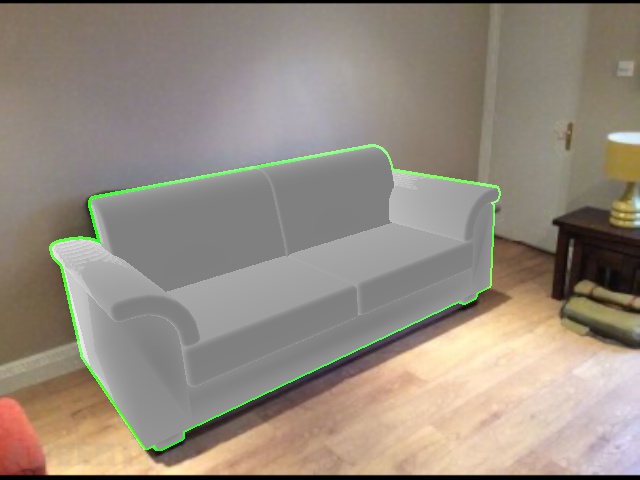}
        \end{minipage}
        \begin{minipage}{0.225\textwidth}
            \centering
            \includegraphics[width=\textwidth]{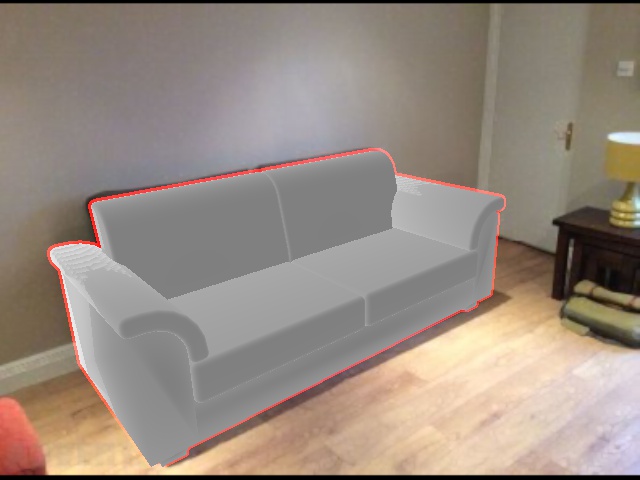}
        \end{minipage}\\[1mm]

    \caption{Qualitative results for Pix3D sofas - part 1.}
    \label{pix3d-sofas-q-1}
\end{figure*}

\begin{figure*}[t]
    \centering
        \begin{minipage}{0.225\textwidth}
        {\small Input image}
            \centering
            \includegraphics[width=\textwidth]{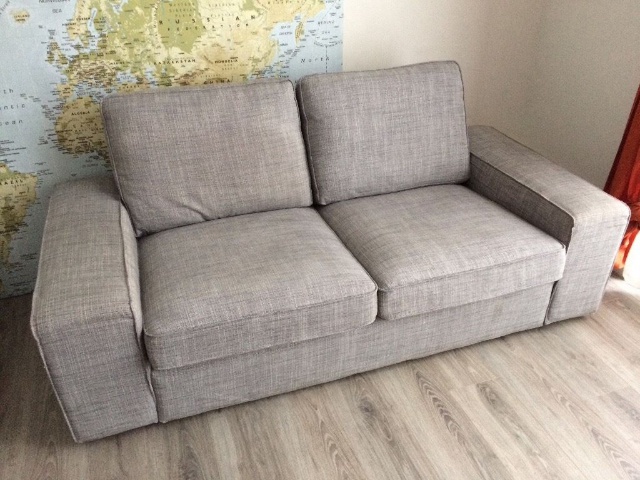}
        \end{minipage}
        \begin{minipage}{0.225\textwidth}
        {\small Ground truth}
            \centering
            \includegraphics[width=\textwidth]{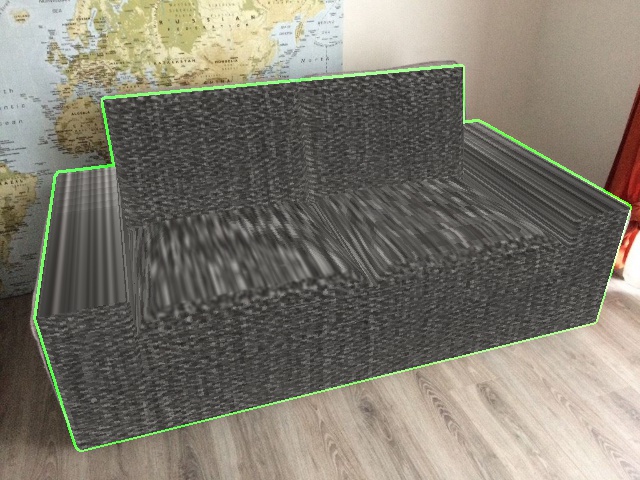}
        \end{minipage}
        \begin{minipage}{0.225\textwidth}
        {\small Our prediction}
            \centering
            \includegraphics[width=\textwidth]{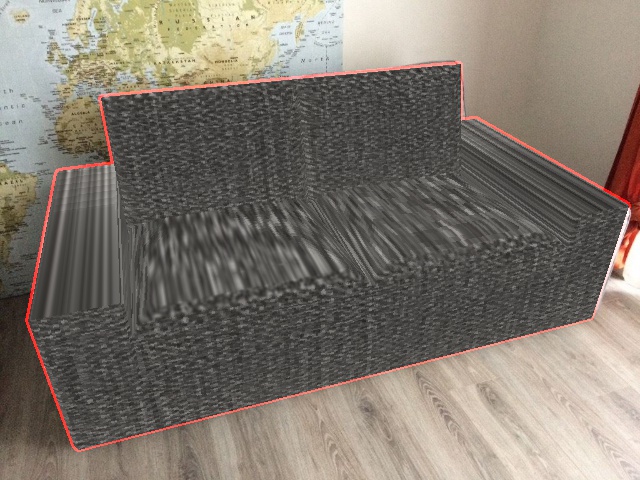}
        \end{minipage}\\[1mm]
        \begin{minipage}{0.225\textwidth}
            \centering
            \includegraphics[width=\textwidth]{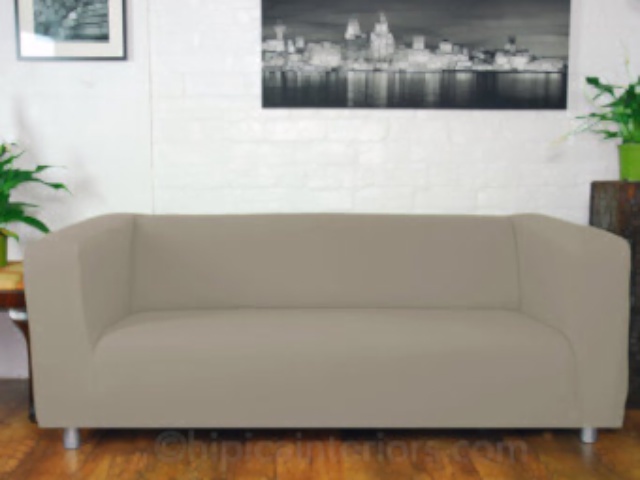}
        \end{minipage}
        \begin{minipage}{0.225\textwidth}
            \centering
            \includegraphics[width=\textwidth]{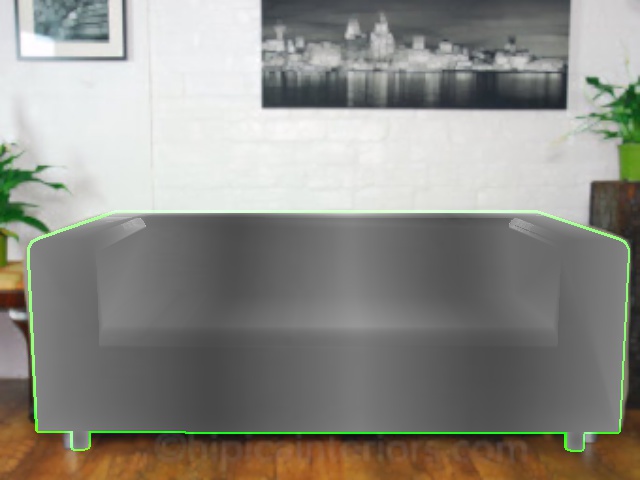}
        \end{minipage}
        \begin{minipage}{0.225\textwidth}
            \centering
            \includegraphics[width=\textwidth]{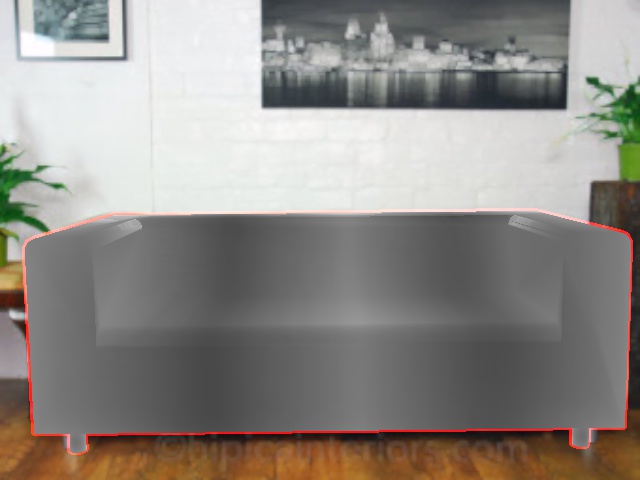}
        \end{minipage}\\[1mm]
        \begin{minipage}{0.225\textwidth}
            \centering
            \includegraphics[width=\textwidth]{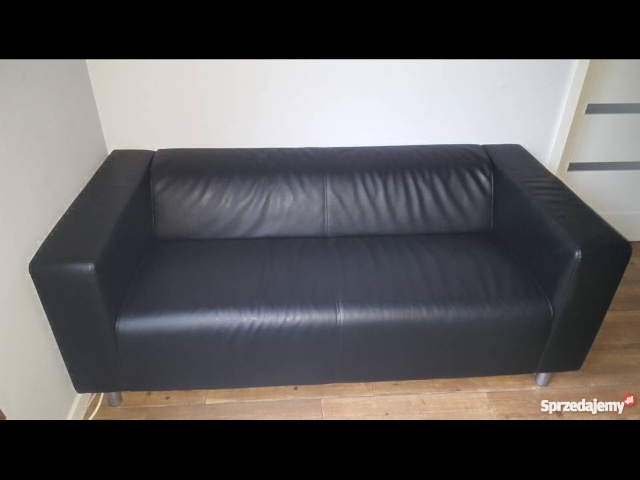}
        \end{minipage}
        \begin{minipage}{0.225\textwidth}
            \centering
            \includegraphics[width=\textwidth]{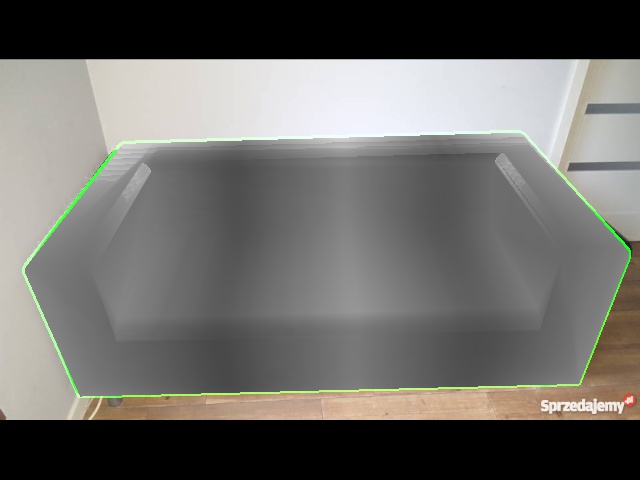}
        \end{minipage}
        \begin{minipage}{0.225\textwidth}
            \centering
            \includegraphics[width=\textwidth]{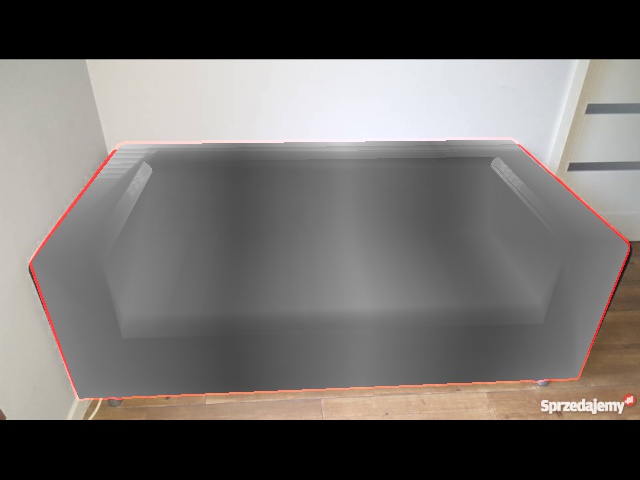}
        \end{minipage}\\[1mm]
        \begin{minipage}{0.225\textwidth}
            \centering
            \includegraphics[width=\textwidth]{figures/sofas/entry_343_input.jpeg}
        \end{minipage}
        \begin{minipage}{0.225\textwidth}
            \centering
            \includegraphics[width=\textwidth]{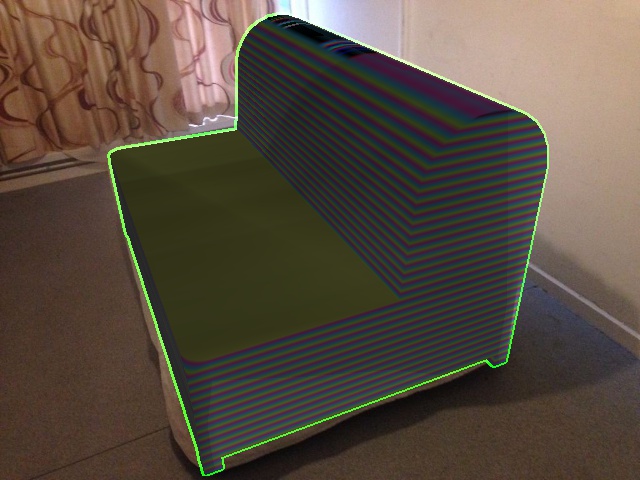}
        \end{minipage}
        \begin{minipage}{0.225\textwidth}
            \centering
            \includegraphics[width=\textwidth]{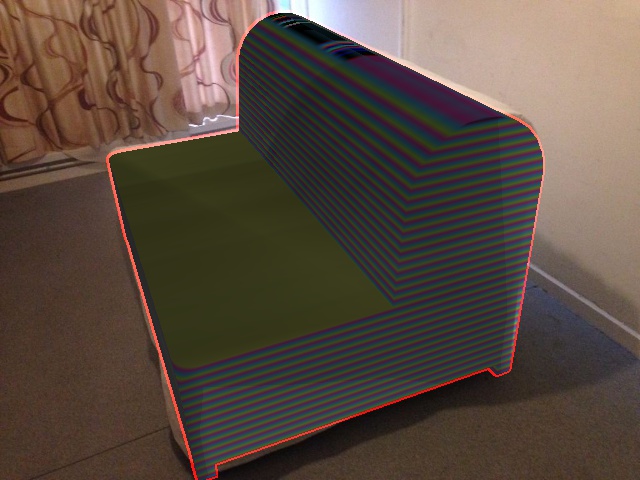}
        \end{minipage}\\[1mm]
        \begin{minipage}{0.225\textwidth}
            \centering
            \includegraphics[width=\textwidth]{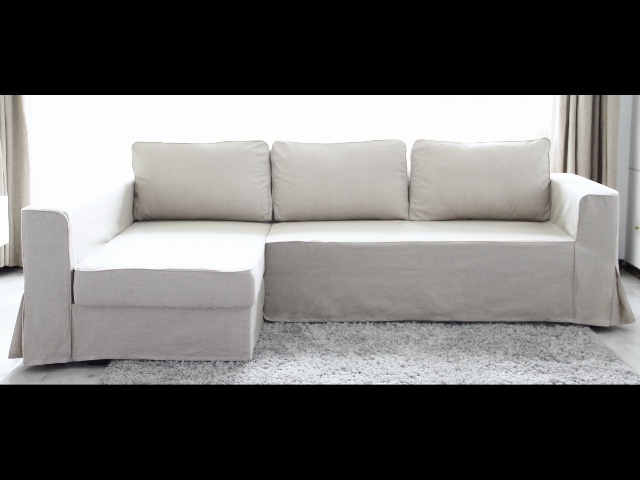}
        \end{minipage}
        \begin{minipage}{0.225\textwidth}
            \centering
            \includegraphics[width=\textwidth]{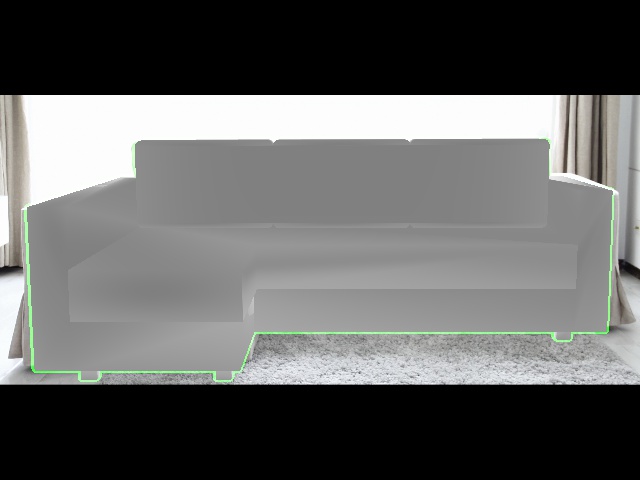}
        \end{minipage}
        \begin{minipage}{0.225\textwidth}
            \centering
            \includegraphics[width=\textwidth]{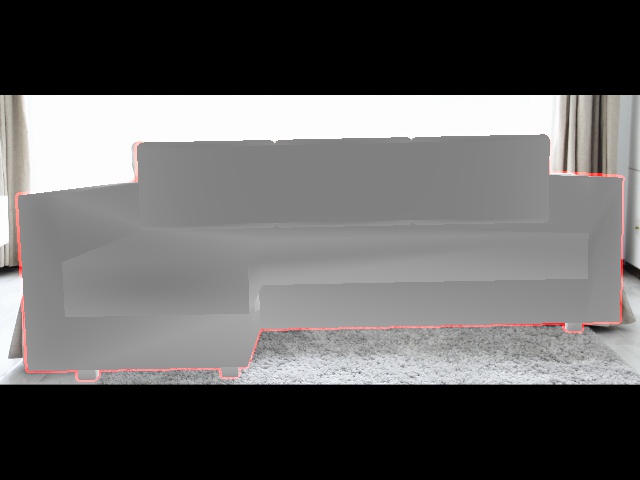}
        \end{minipage}\\[1mm]
        \begin{minipage}{0.225\textwidth}
            \centering
            \includegraphics[width=\textwidth]{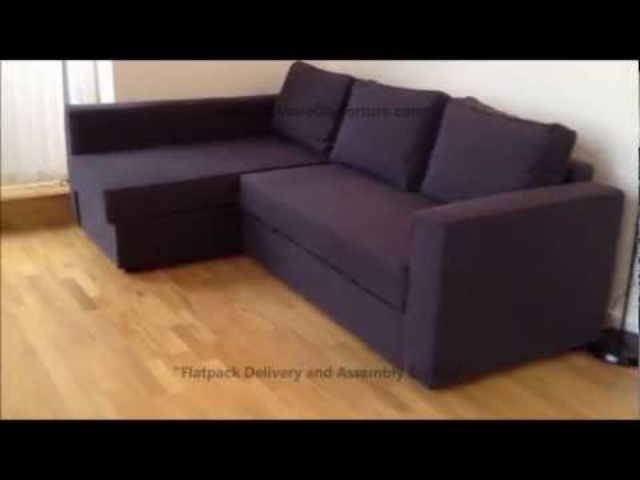}
        \end{minipage}
        \begin{minipage}{0.225\textwidth}
            \centering
            \includegraphics[width=\textwidth]{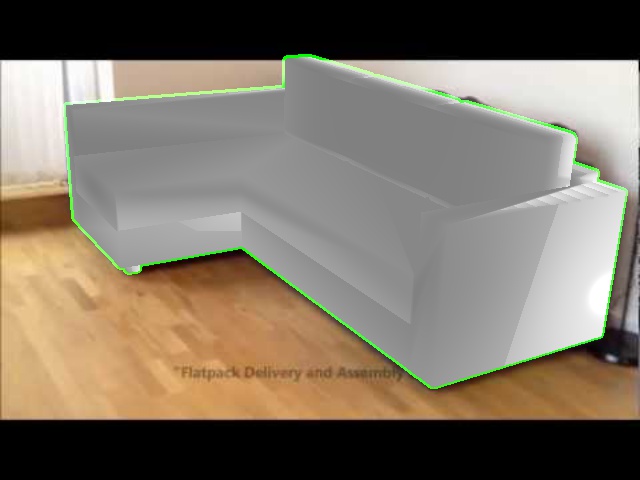}
        \end{minipage}
        \begin{minipage}{0.225\textwidth}
            \centering
            \includegraphics[width=\textwidth]{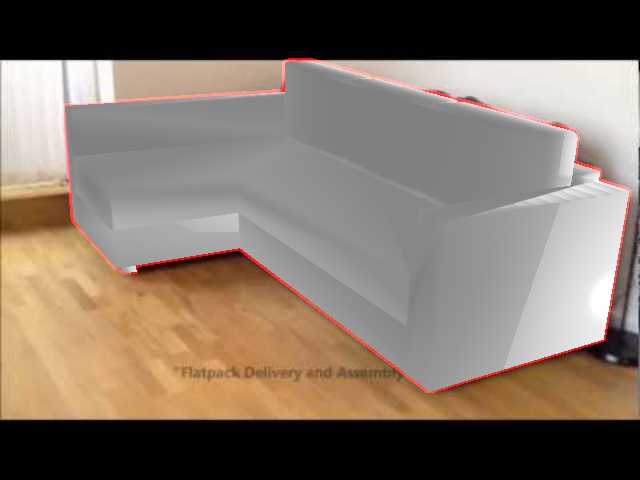}
        \end{minipage}\\[1mm]
        \begin{minipage}{0.225\textwidth}
            \centering
            \includegraphics[width=\textwidth]{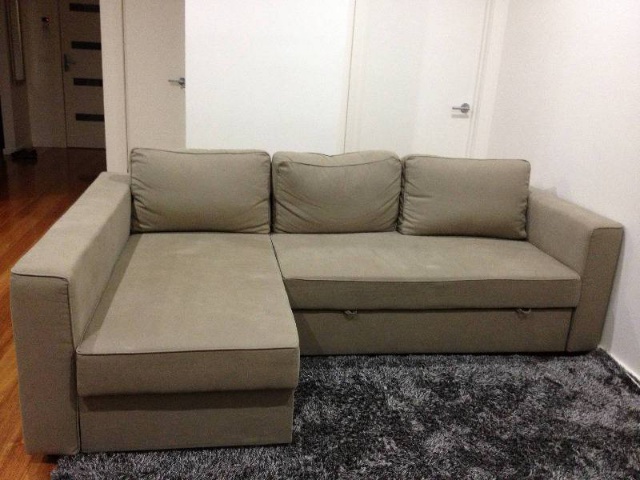}
        \end{minipage}
        \begin{minipage}{0.225\textwidth}
            \centering
            \includegraphics[width=\textwidth]{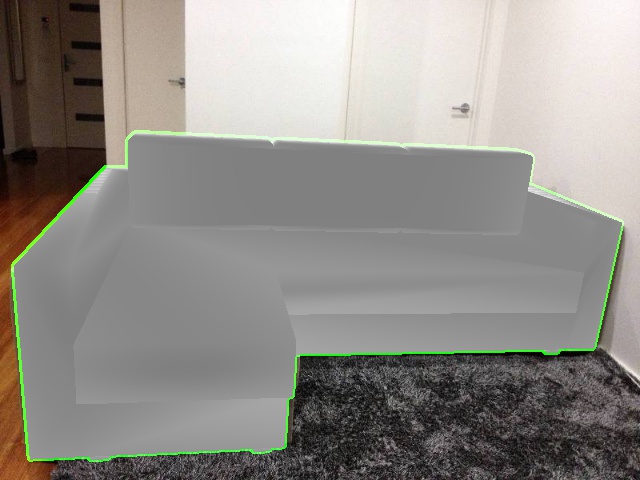}
        \end{minipage}
        \begin{minipage}{0.225\textwidth}
            \centering
            \includegraphics[width=\textwidth]{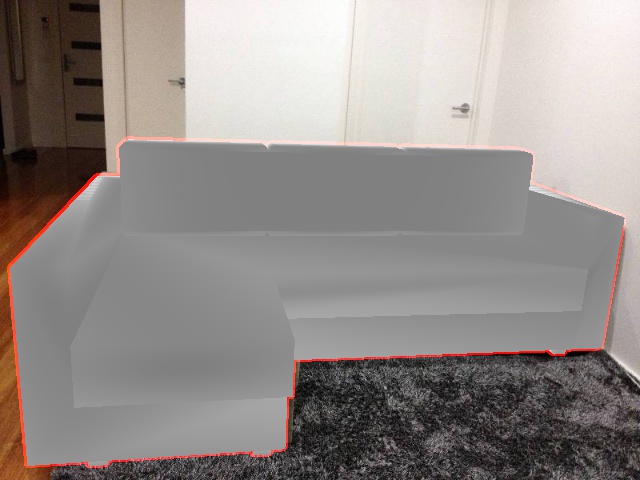}
        \end{minipage}\\[1mm]

    \caption{Qualitative results for Pix3D sofas - part 2.}
    \label{pix3d-sofas-q-2}
\end{figure*}

\begin{figure*}[t]
    \centering
        \begin{minipage}{0.225\textwidth}
        {\small Input image}
            \centering
            \includegraphics[width=\textwidth]{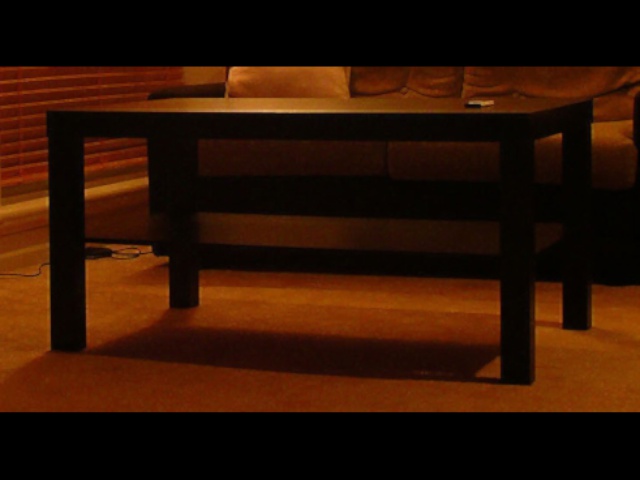}
        \end{minipage}
        \begin{minipage}{0.225\textwidth}
        {\small Ground truth}
            \centering
            \includegraphics[width=\textwidth]{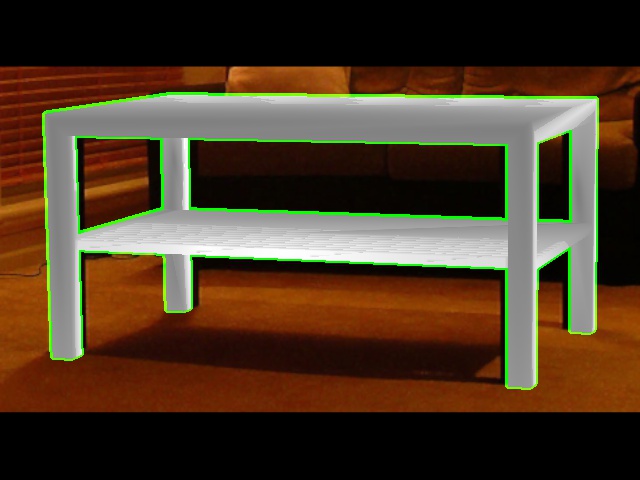}
        \end{minipage}
        \begin{minipage}{0.225\textwidth}
        {\small Our prediction}
            \centering
            \includegraphics[width=\textwidth]{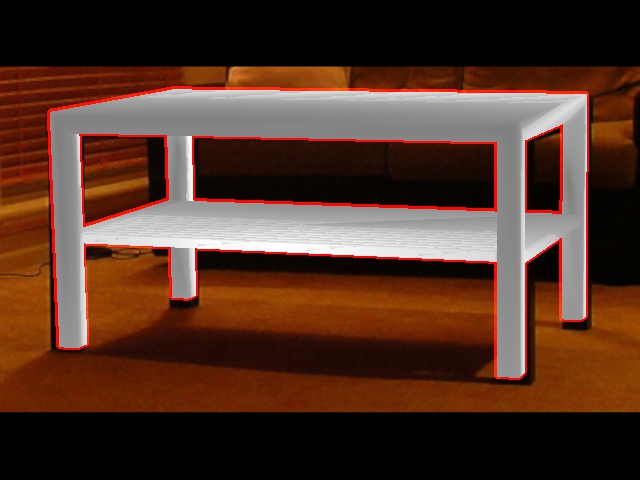}
        \end{minipage}\\[1mm]
        \begin{minipage}{0.225\textwidth}
            \centering
            \includegraphics[width=\textwidth]{figures/tables/entry_5_input.jpeg}
        \end{minipage}
        \begin{minipage}{0.225\textwidth}
            \centering
            \includegraphics[width=\textwidth]{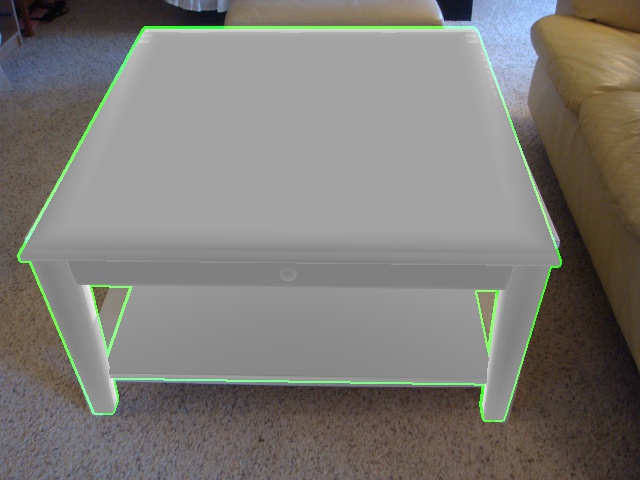}
        \end{minipage}
        \begin{minipage}{0.225\textwidth}
            \centering
            \includegraphics[width=\textwidth]{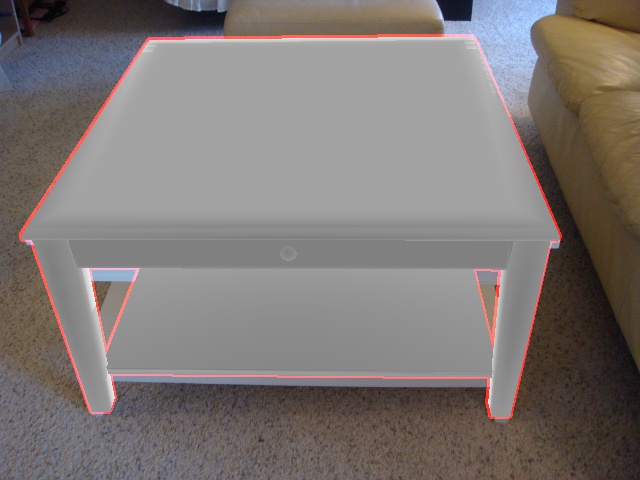}
        \end{minipage}\\[1mm]
        \begin{minipage}{0.225\textwidth}
            \centering
            \includegraphics[width=\textwidth]{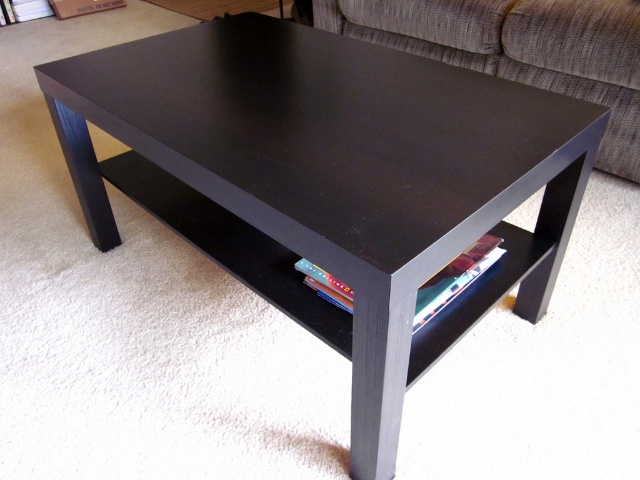}
        \end{minipage}
        \begin{minipage}{0.225\textwidth}
            \centering
            \includegraphics[width=\textwidth]{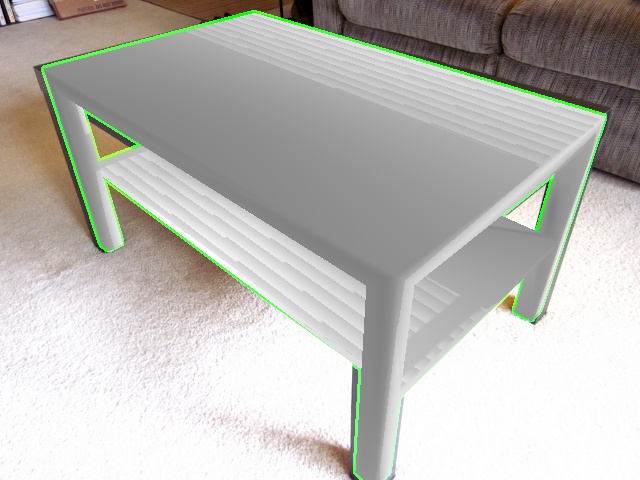}
        \end{minipage}
        \begin{minipage}{0.225\textwidth}
            \centering
            \includegraphics[width=\textwidth]{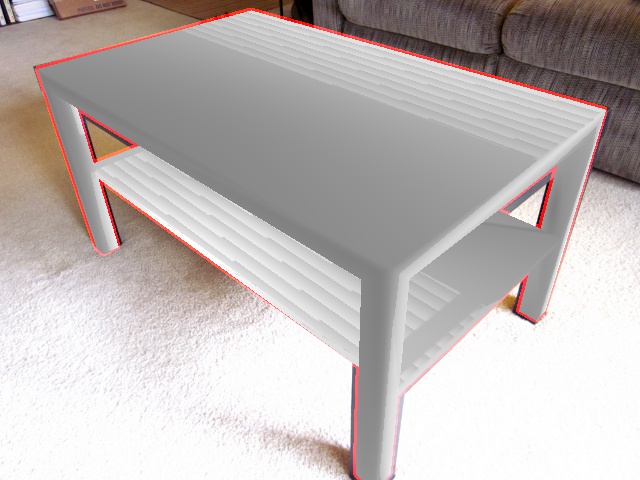}
        \end{minipage}\\[1mm]
        \begin{minipage}{0.225\textwidth}
            \centering
            \includegraphics[width=\textwidth]{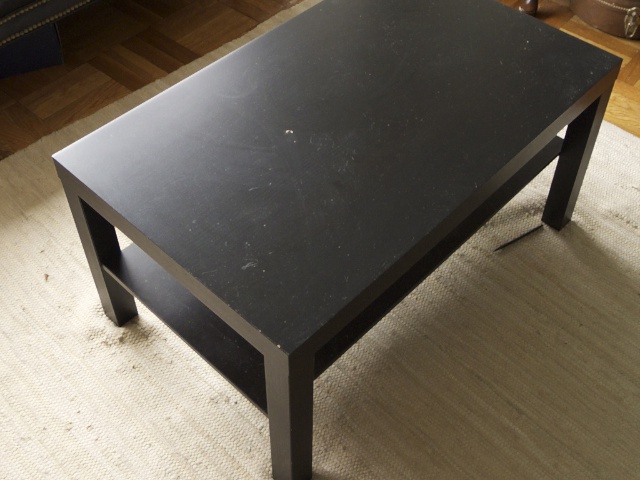}
        \end{minipage}
        \begin{minipage}{0.225\textwidth}
            \centering
            \includegraphics[width=\textwidth]{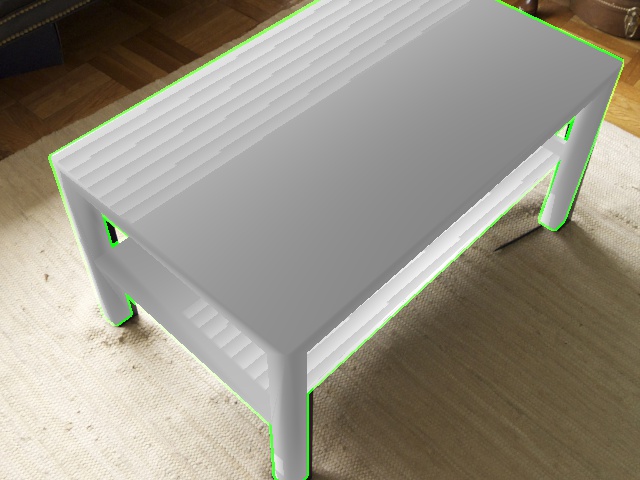}
        \end{minipage}
        \begin{minipage}{0.225\textwidth}
            \centering
            \includegraphics[width=\textwidth]{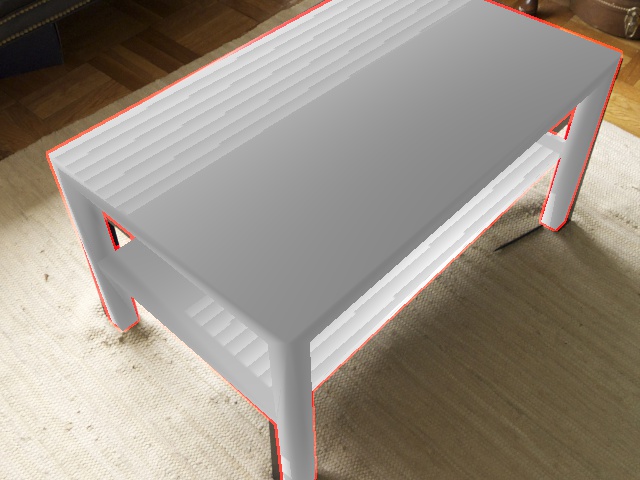}
        \end{minipage}\\[1mm]
        \begin{minipage}{0.225\textwidth}
            \centering
            \includegraphics[width=\textwidth]{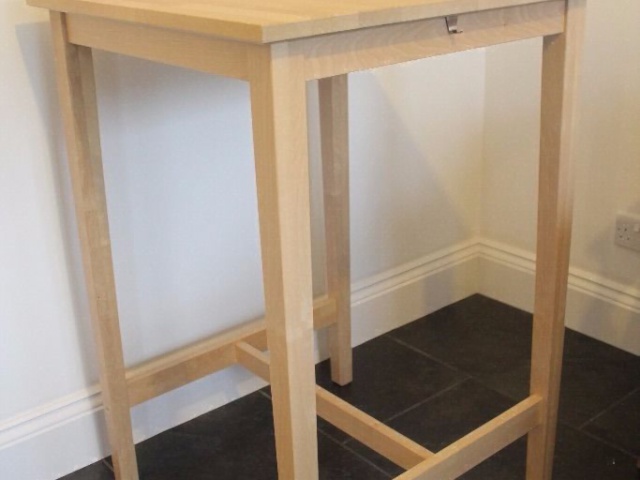}
        \end{minipage}
        \begin{minipage}{0.225\textwidth}
            \centering
            \includegraphics[width=\textwidth]{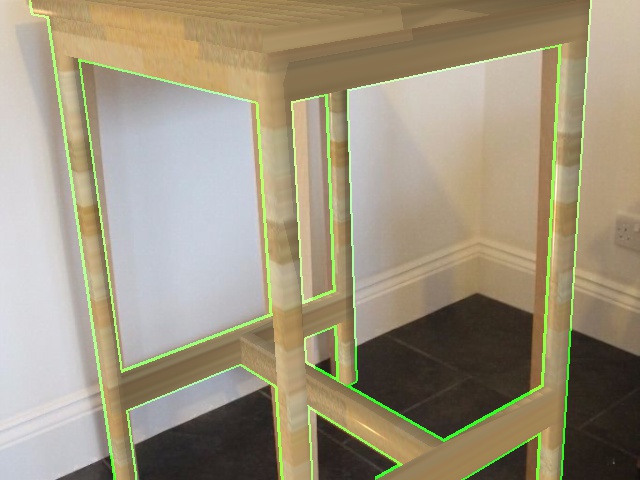}
        \end{minipage}
        \begin{minipage}{0.225\textwidth}
            \centering
            \includegraphics[width=\textwidth]{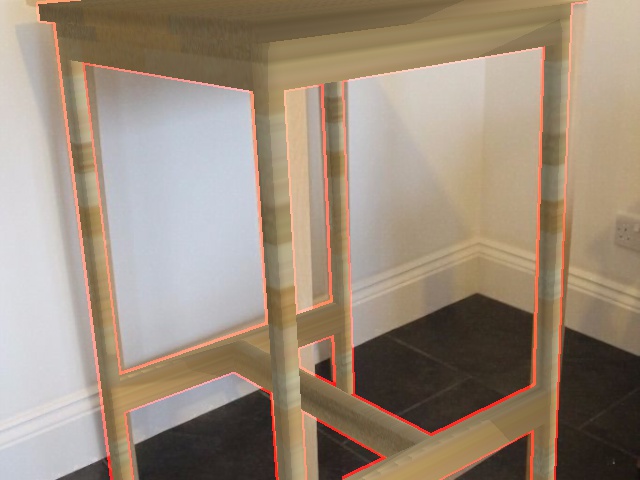}
        \end{minipage}\\[1mm]
                \begin{minipage}{0.225\textwidth}
            \centering
            \includegraphics[width=\textwidth]{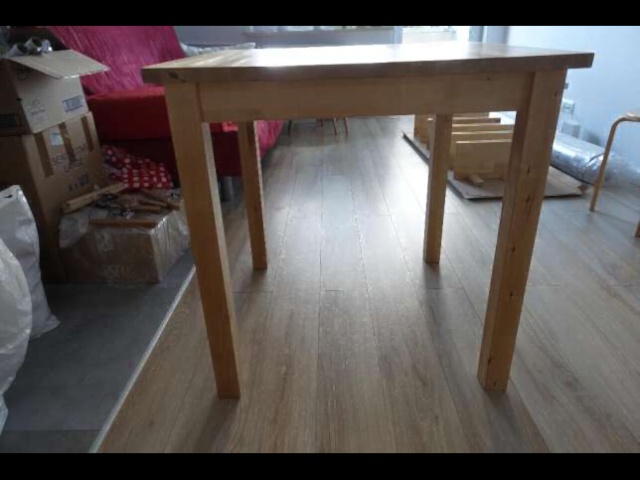}
        \end{minipage}
        \begin{minipage}{0.225\textwidth}
            \centering
            \includegraphics[width=\textwidth]{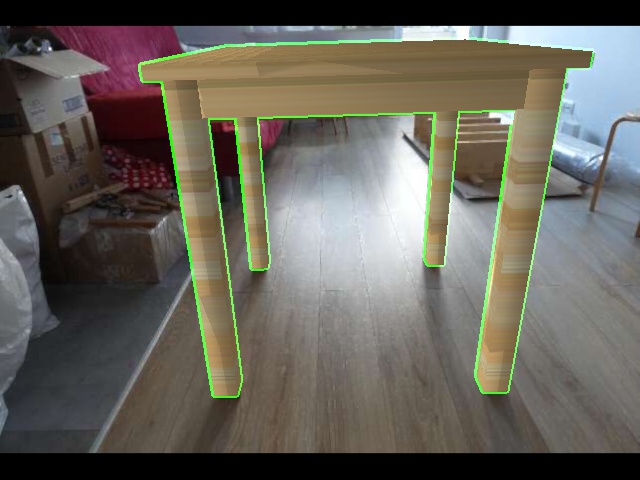}
        \end{minipage}
        \begin{minipage}{0.225\textwidth}
            \centering
            \includegraphics[width=\textwidth]{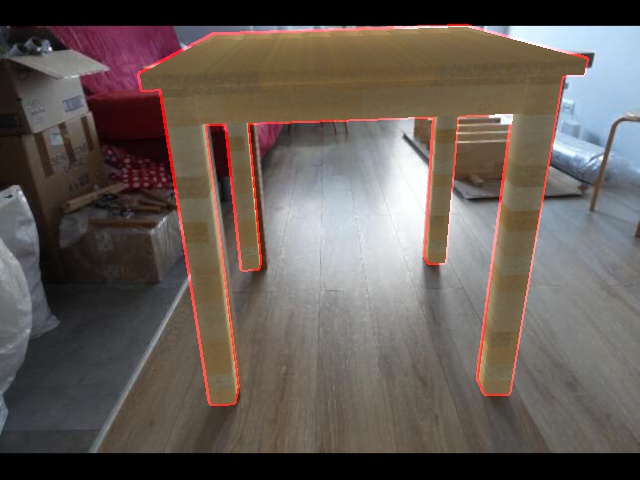}
        \end{minipage}\\[1mm]
        \begin{minipage}{0.225\textwidth}
            \centering
            \includegraphics[width=\textwidth]{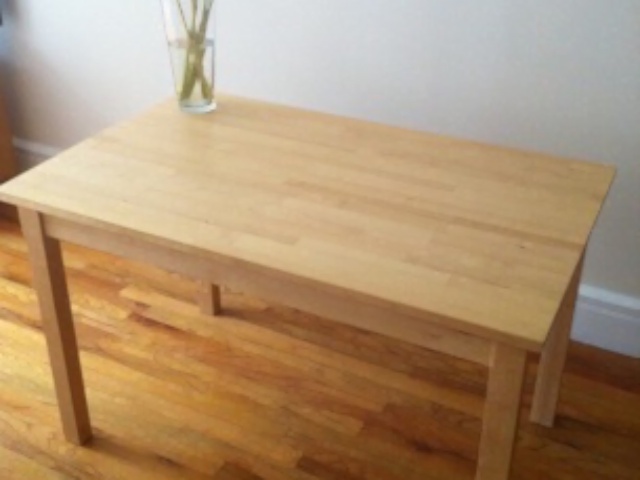}
        \end{minipage}
        \begin{minipage}{0.225\textwidth}
            \centering
            \includegraphics[width=\textwidth]{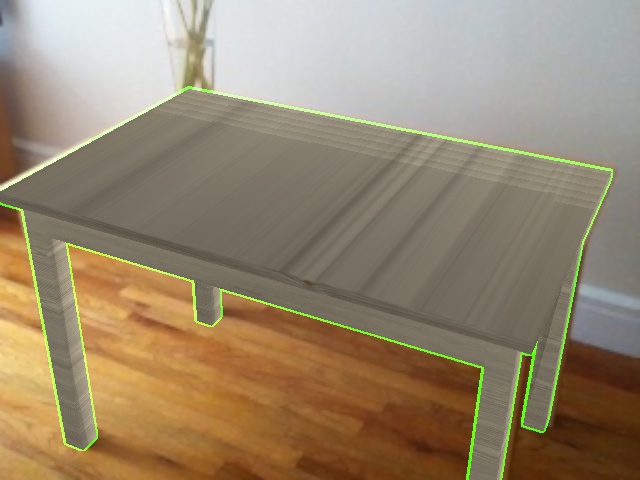}
        \end{minipage}
        \begin{minipage}{0.225\textwidth}
            \centering
            \includegraphics[width=\textwidth]{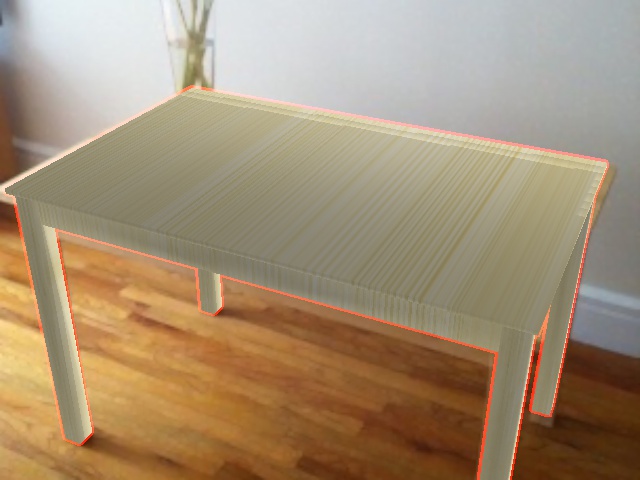}
        \end{minipage}\\[1mm]

    \caption{Qualitative results for Pix3D tables - part 1.}
    \label{pix3d-tables-q-1}
\end{figure*}

\begin{figure*}[t]
    \centering
        \begin{minipage}{0.225\textwidth}
        {\small Input image}
            \centering
            \includegraphics[width=\textwidth]{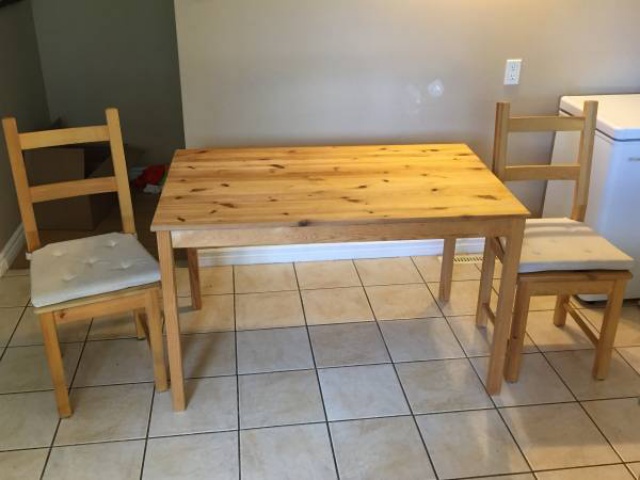}
        \end{minipage}
        \begin{minipage}{0.225\textwidth}
        {\small Ground truth}
            \centering
            \includegraphics[width=\textwidth]{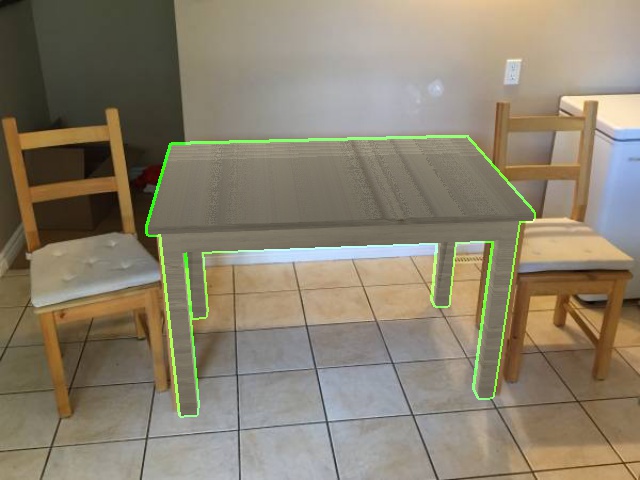}
        \end{minipage}
        \begin{minipage}{0.225\textwidth}
        {\small Our prediction}
            \centering
            \includegraphics[width=\textwidth]{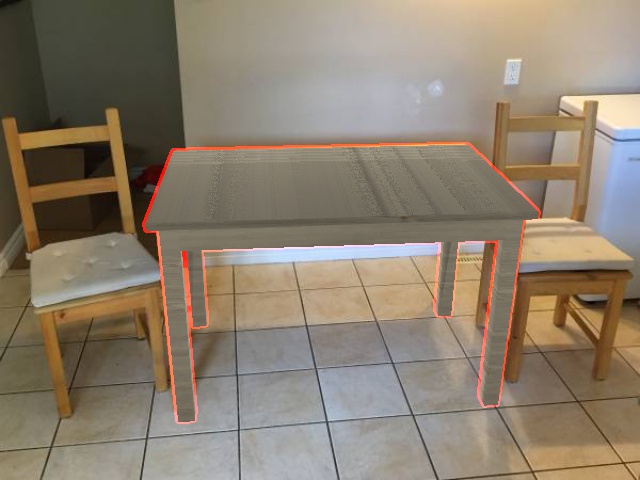}
        \end{minipage}\\[1mm]
        \begin{minipage}{0.225\textwidth}
            \centering
            \includegraphics[width=\textwidth]{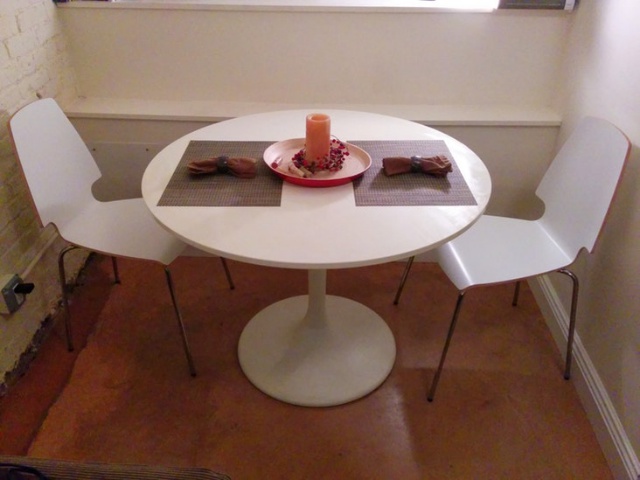}
        \end{minipage}
        \begin{minipage}{0.225\textwidth}
            \centering
            \includegraphics[width=\textwidth]{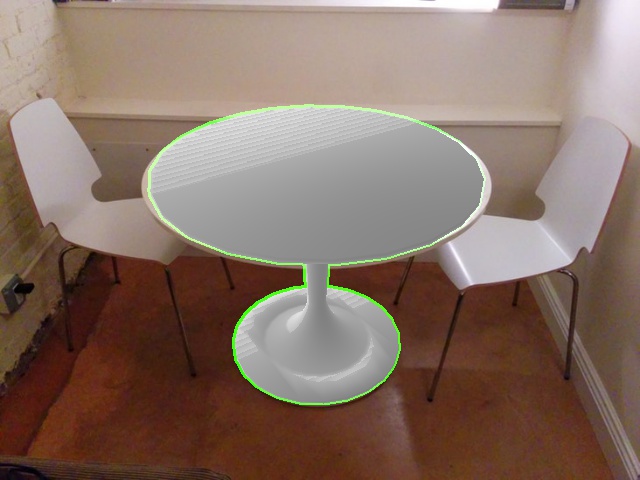}
        \end{minipage}
        \begin{minipage}{0.225\textwidth}
            \centering
            \includegraphics[width=\textwidth]{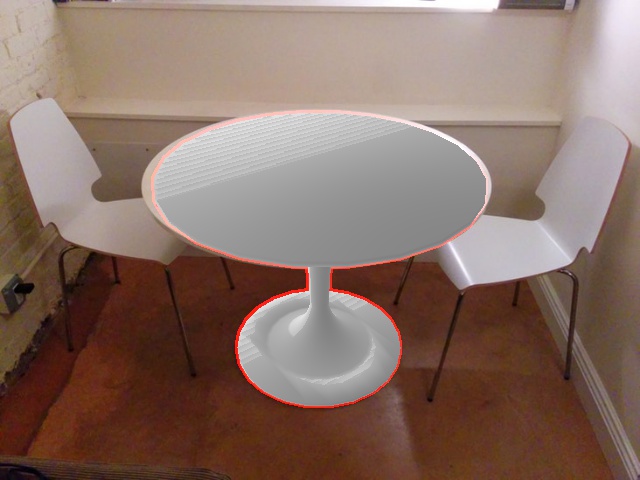}
        \end{minipage}\\[1mm]
        \begin{minipage}{0.225\textwidth}
            \centering
            \includegraphics[width=\textwidth]{figures/tables/entry_88_input.jpeg}
        \end{minipage}
        \begin{minipage}{0.225\textwidth}
            \centering
            \includegraphics[width=\textwidth]{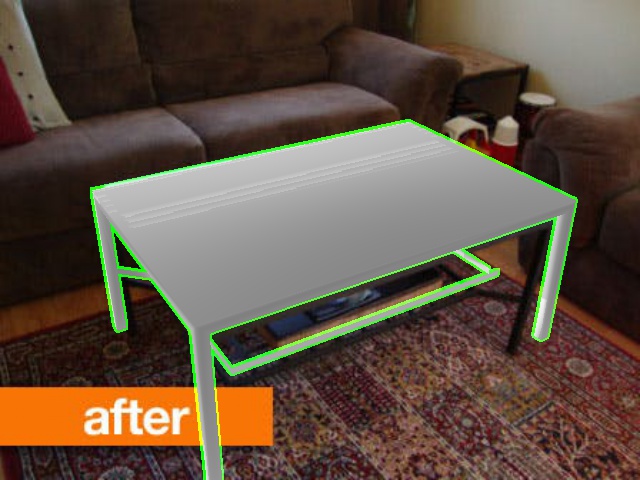}
        \end{minipage}
        \begin{minipage}{0.225\textwidth}
            \centering
            \includegraphics[width=\textwidth]{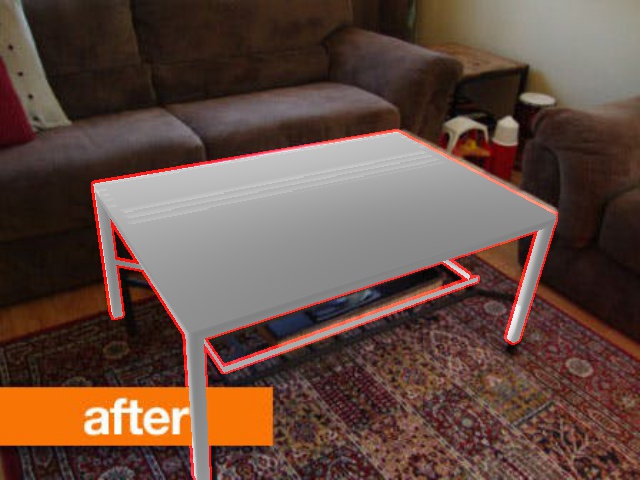}
        \end{minipage}\\[1mm]
        \begin{minipage}{0.225\textwidth}
            \centering
            \includegraphics[width=\textwidth]{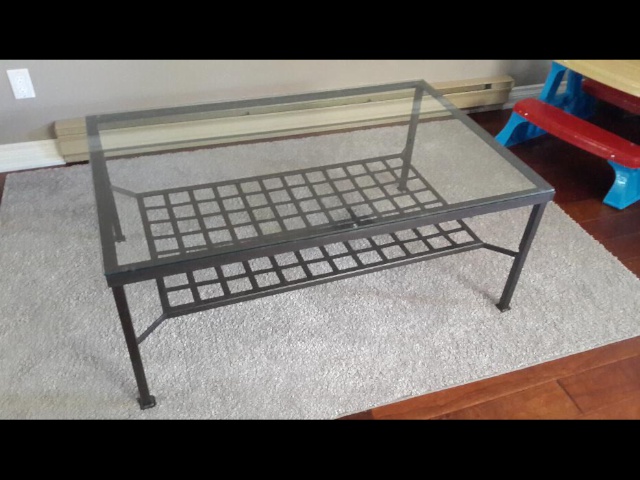}
        \end{minipage}
        \begin{minipage}{0.225\textwidth}
            \centering
            \includegraphics[width=\textwidth]{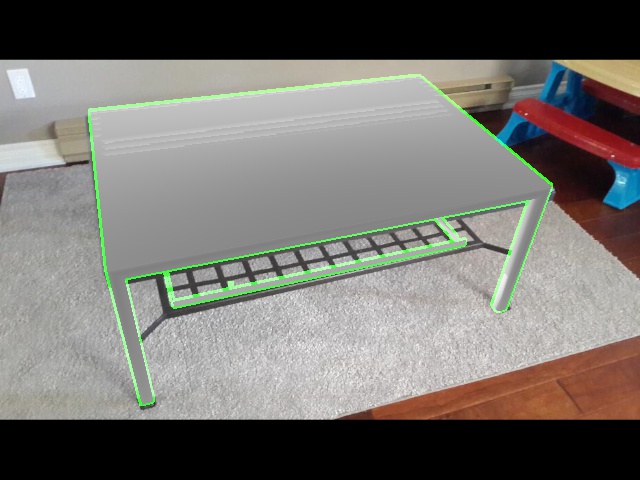}
        \end{minipage}
        \begin{minipage}{0.225\textwidth}
            \centering
            \includegraphics[width=\textwidth]{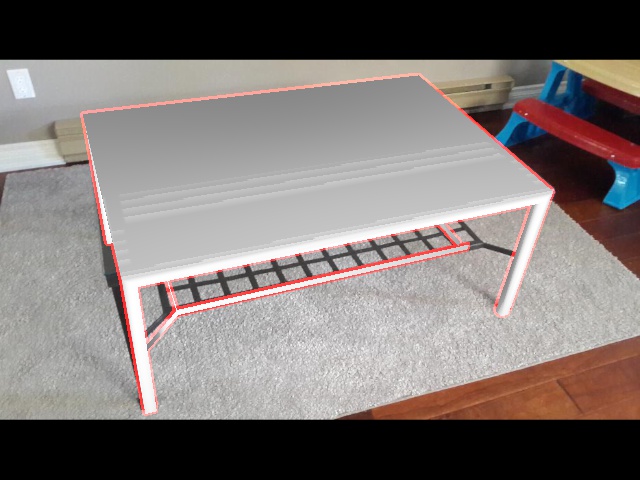}
        \end{minipage}\\[1mm]
        \begin{minipage}{0.225\textwidth}
            \centering
            \includegraphics[width=\textwidth]{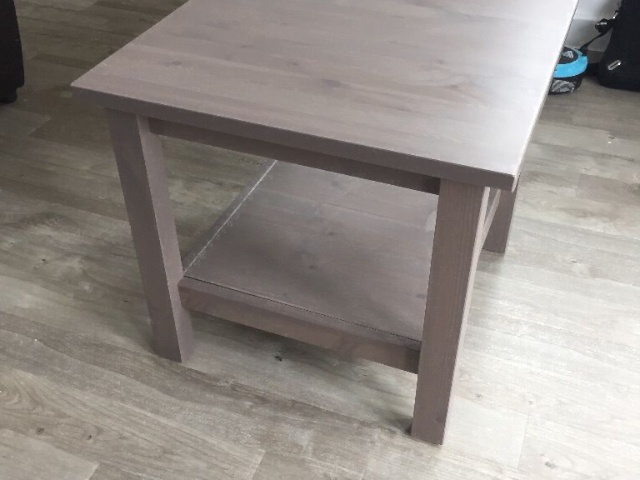}
        \end{minipage}
        \begin{minipage}{0.225\textwidth}
            \centering
            \includegraphics[width=\textwidth]{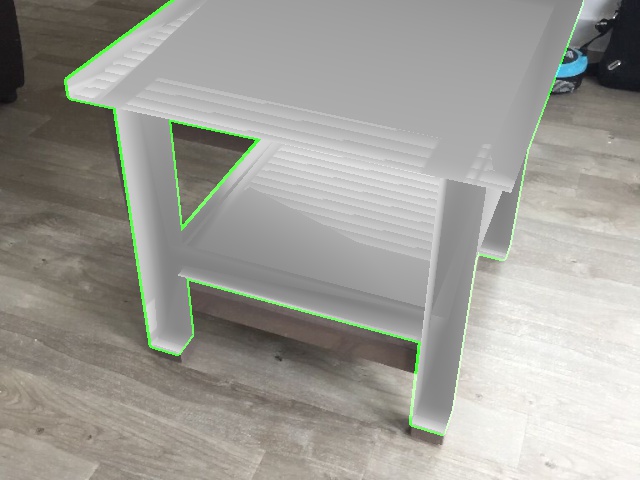}
        \end{minipage}
        \begin{minipage}{0.225\textwidth}
            \centering
            \includegraphics[width=\textwidth]{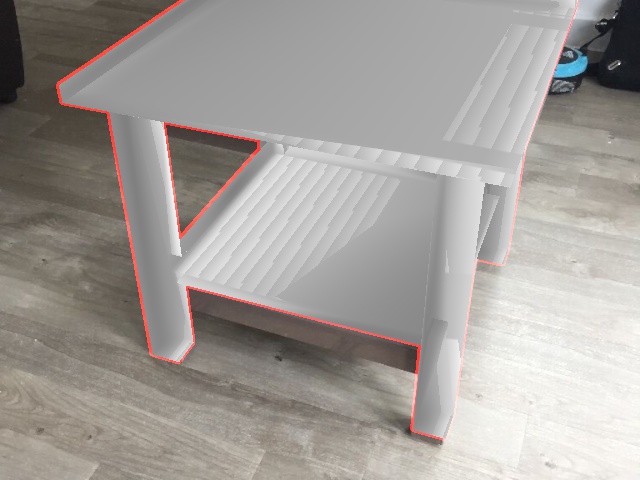}
        \end{minipage}\\[1mm]
                \begin{minipage}{0.225\textwidth}
            \centering
            \includegraphics[width=\textwidth]{figures/tables/entry_160_input.jpeg}
        \end{minipage}
        \begin{minipage}{0.225\textwidth}
            \centering
            \includegraphics[width=\textwidth]{figures/tables/entry_160_gt.jpeg}
        \end{minipage}
        \begin{minipage}{0.225\textwidth}
            \centering
            \includegraphics[width=\textwidth]{figures/tables/entry_160_pred.jpeg}
        \end{minipage}\\[1mm]
        \begin{minipage}{0.225\textwidth}
            \centering
            \includegraphics[width=\textwidth]{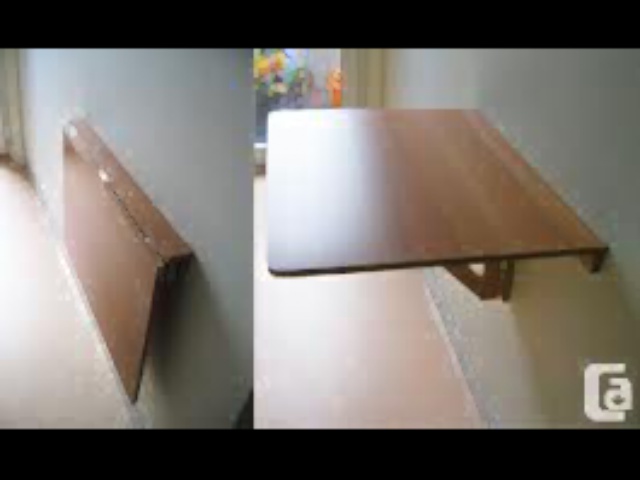}
        \end{minipage}
        \begin{minipage}{0.225\textwidth}
            \centering
            \includegraphics[width=\textwidth]{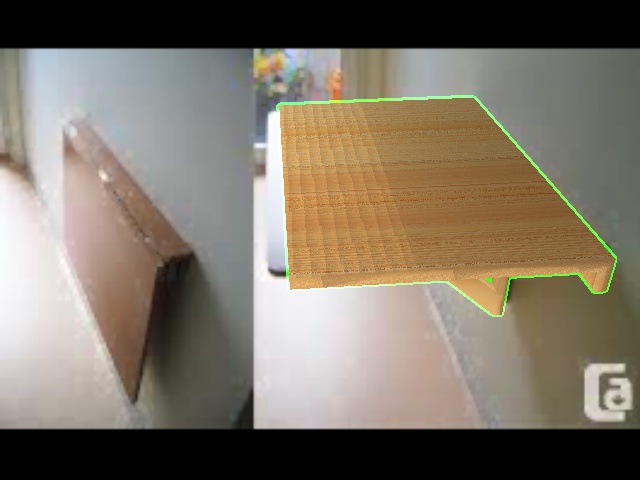}
        \end{minipage}
        \begin{minipage}{0.225\textwidth}
            \centering
            \includegraphics[width=\textwidth]{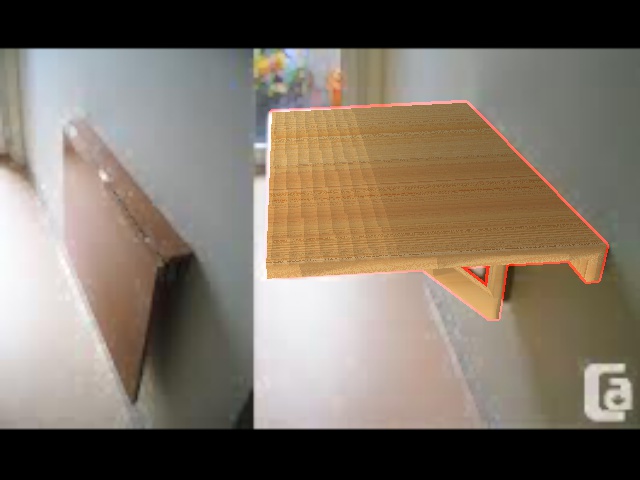}
        \end{minipage}\\[1mm]

    \caption{Qualitative results for Pix3D tables - part 2.}
    \label{pix3d-tables-q-2}
\end{figure*}

\begin{figure*}[t]
    \centering
        \begin{minipage}{0.225\textwidth}
        {\small Input image}
            \centering
            \includegraphics[width=\textwidth]{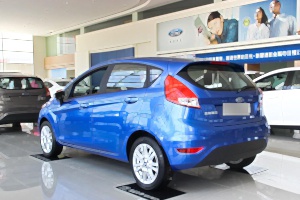}
        \end{minipage}
        \begin{minipage}{0.225\textwidth}
        {\small Ground truth}
            \centering
            \includegraphics[width=\textwidth]{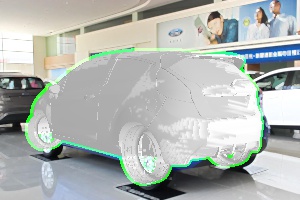}
        \end{minipage}
        \begin{minipage}{0.225\textwidth}
        {\small Our prediction}
            \centering
            \includegraphics[width=\textwidth]{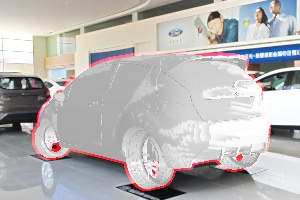}
        \end{minipage}\\[1mm]
        \begin{minipage}{0.225\textwidth}
            \centering
            \includegraphics[width=\textwidth]{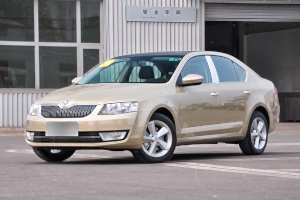}
        \end{minipage}
        \begin{minipage}{0.225\textwidth}
            \centering
            \includegraphics[width=\textwidth]{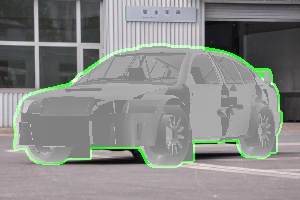}
        \end{minipage}
        \begin{minipage}{0.225\textwidth}
            \centering
            \includegraphics[width=\textwidth]{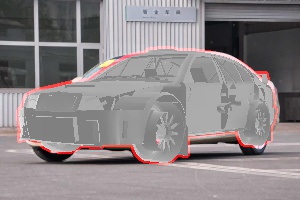}
        \end{minipage}\\[1mm]
        \begin{minipage}{0.225\textwidth}
            \centering
            \includegraphics[width=\textwidth]{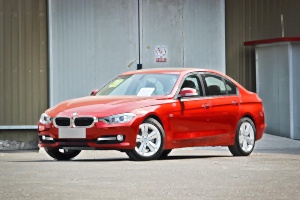}
        \end{minipage}
        \begin{minipage}{0.225\textwidth}
            \centering
            \includegraphics[width=\textwidth]{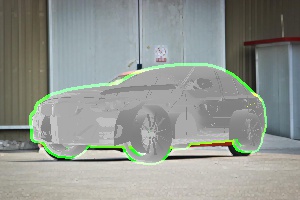}
        \end{minipage}
        \begin{minipage}{0.225\textwidth}
            \centering
            \includegraphics[width=\textwidth]{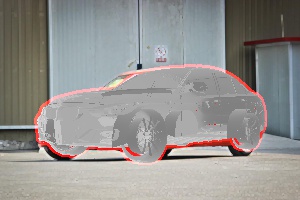}
        \end{minipage}\\[1mm]
        \begin{minipage}{0.225\textwidth}
            \centering
            \includegraphics[width=\textwidth]{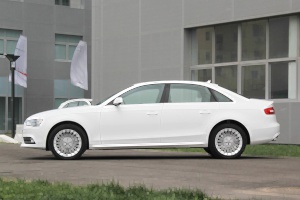}
        \end{minipage}
        \begin{minipage}{0.225\textwidth}
            \centering
            \includegraphics[width=\textwidth]{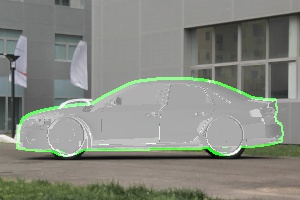}
        \end{minipage}
        \begin{minipage}{0.225\textwidth}
            \centering
            \includegraphics[width=\textwidth]{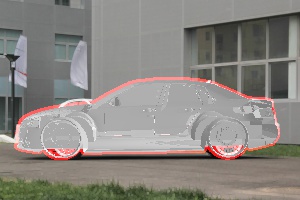}
        \end{minipage}\\[1mm]
        \begin{minipage}{0.225\textwidth}
            \centering
            \includegraphics[width=\textwidth]{figures/compcars_qualitative/entry_51_input.jpeg}
        \end{minipage}
        \begin{minipage}{0.225\textwidth}
            \centering
            \includegraphics[width=\textwidth]{figures/compcars_qualitative/entry_51_gt.jpeg}
        \end{minipage}
        \begin{minipage}{0.225\textwidth}
            \centering
            \includegraphics[width=\textwidth]{figures/compcars_qualitative/entry_51_pred.jpeg}
        \end{minipage}\\[1mm]
        \begin{minipage}{0.225\textwidth}
            \centering
            \includegraphics[width=\textwidth]{figures/compcars_qualitative/entry_53_input.jpeg}
        \end{minipage}
        \begin{minipage}{0.225\textwidth}
            \centering
            \includegraphics[width=\textwidth]{figures/compcars_qualitative/entry_53_gt.jpeg}
        \end{minipage}
        \begin{minipage}{0.225\textwidth}
            \centering
            \includegraphics[width=\textwidth]{figures/compcars_qualitative/entry_53_pred.jpeg}
        \end{minipage}\\[1mm]
        \begin{minipage}{0.225\textwidth}
            \centering
            \includegraphics[width=\textwidth]{figures/compcars_qualitative/entry_72_input.jpeg}
        \end{minipage}
        \begin{minipage}{0.225\textwidth}
            \centering
            \includegraphics[width=\textwidth]{figures/compcars_qualitative/entry_72_gt.jpeg}
        \end{minipage}
        \begin{minipage}{0.225\textwidth}
            \centering
            \includegraphics[width=\textwidth]{figures/compcars_qualitative/entry_72_pred.jpeg}
        \end{minipage}\\[1mm]
    \caption{Qualitative results for the CompCars dataset.}
    \label{compcars-q}
\end{figure*}

\begin{figure*}[t]
    \centering
        \begin{minipage}{0.225\textwidth}
        {\small Input image}
            \centering
            \includegraphics[width=\textwidth]{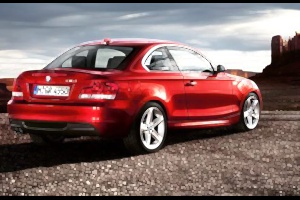}
        \end{minipage}
        \begin{minipage}{0.225\textwidth}
        {\small Ground truth}
            \centering
            \includegraphics[width=\textwidth]{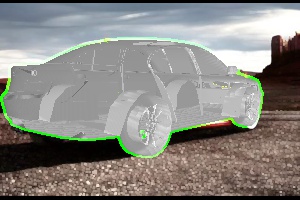}
        \end{minipage}
        \begin{minipage}{0.225\textwidth}
        {\small Our prediction}
            \centering
            \includegraphics[width=\textwidth]{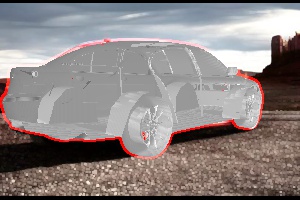}
        \end{minipage}\\[1mm]
        \begin{minipage}{0.225\textwidth}
            \centering
            \includegraphics[width=\textwidth]{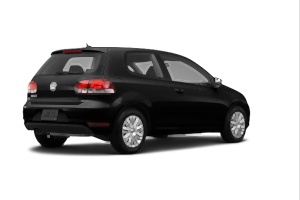}
        \end{minipage}
        \begin{minipage}{0.225\textwidth}
            \centering
            \includegraphics[width=\textwidth]{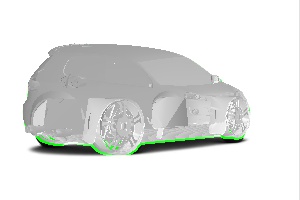}
        \end{minipage}
        \begin{minipage}{0.225\textwidth}
            \centering
            \includegraphics[width=\textwidth]{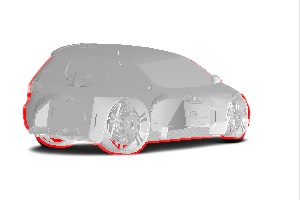}
        \end{minipage}\\[1mm]
        \begin{minipage}{0.225\textwidth}
            \centering
            \includegraphics[width=\textwidth]{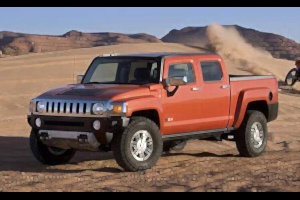}
        \end{minipage}
        \begin{minipage}{0.225\textwidth}
            \centering
            \includegraphics[width=\textwidth]{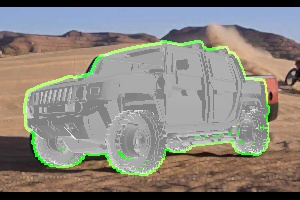}
        \end{minipage}
        \begin{minipage}{0.225\textwidth}
            \centering
            \includegraphics[width=\textwidth]{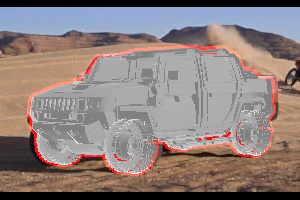}
        \end{minipage}\\[1mm]
        \begin{minipage}{0.225\textwidth}
            \centering
            \includegraphics[width=\textwidth]{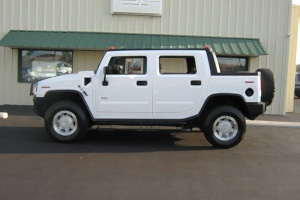}
        \end{minipage}
        \begin{minipage}{0.225\textwidth}
            \centering
            \includegraphics[width=\textwidth]{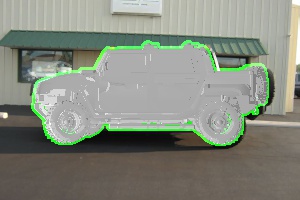}
        \end{minipage}
        \begin{minipage}{0.225\textwidth}
            \centering
            \includegraphics[width=\textwidth]{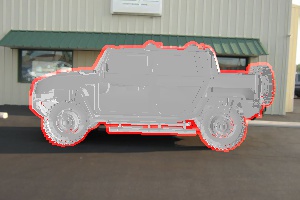}
        \end{minipage}\\[1mm]
        \begin{minipage}{0.225\textwidth}
            \centering
            \includegraphics[width=\textwidth]{figures/stanfordcars_qualitative/entry_71_input.jpeg}
        \end{minipage}
        \begin{minipage}{0.225\textwidth}
            \centering
            \includegraphics[width=\textwidth]{figures/stanfordcars_qualitative/entry_71_gt.jpeg}
        \end{minipage}
        \begin{minipage}{0.225\textwidth}
            \centering
            \includegraphics[width=\textwidth]{figures/stanfordcars_qualitative/entry_71_pred.jpeg}
        \end{minipage}\\[1mm]
        \begin{minipage}{0.225\textwidth}
            \centering
            \includegraphics[width=\textwidth]{figures/stanfordcars_qualitative/entry_163_input.jpeg}
        \end{minipage}
        \begin{minipage}{0.225\textwidth}
            \centering
            \includegraphics[width=\textwidth]{figures/stanfordcars_qualitative/entry_163_gt.jpeg}
        \end{minipage}
        \begin{minipage}{0.225\textwidth}
            \centering
            \includegraphics[width=\textwidth]{figures/stanfordcars_qualitative/entry_163_pred.jpeg}
        \end{minipage}\\[1mm]
        \begin{minipage}{0.225\textwidth}
            \centering
            \includegraphics[width=\textwidth]{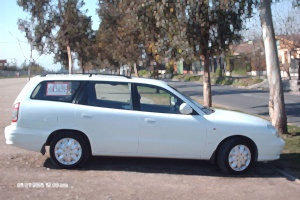}
        \end{minipage}
        \begin{minipage}{0.225\textwidth}
            \centering
            \includegraphics[width=\textwidth]{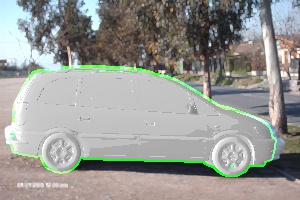}
        \end{minipage}
        \begin{minipage}{0.225\textwidth}
            \centering
            \includegraphics[width=\textwidth]{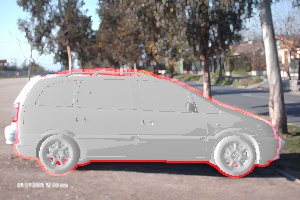}
        \end{minipage}\\[1mm]
    \caption{Qualitative results for the Stanford car dataset.}
    \label{stanford-q}
\end{figure*}

\begin{figure*}[t]
    \centering
    \small{1}
    \begin{minipage}{0.225\textwidth}
        {\small Input image}
            \centering
            \includegraphics[width=\textwidth]{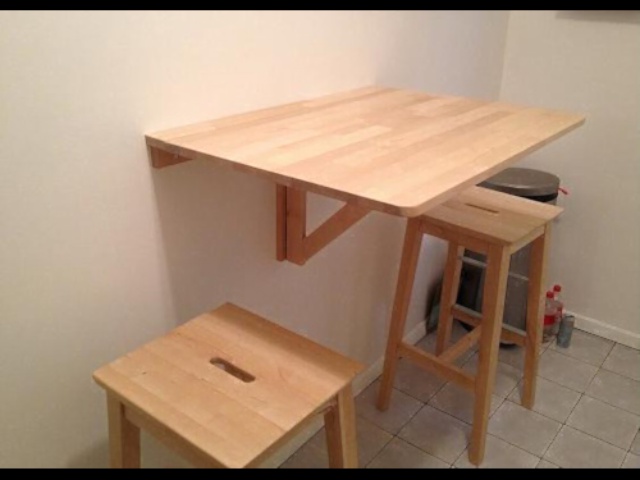}
        \end{minipage}
        \begin{minipage}{0.225\textwidth}
        {\small Ground truth}
            \centering
            \includegraphics[width=\textwidth]{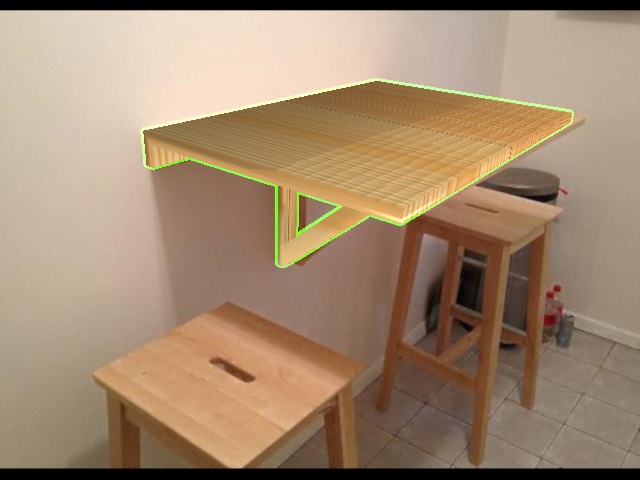}
        \end{minipage}
        \begin{minipage}{0.225\textwidth}
        {\small Our prediction}
            \centering
            \includegraphics[width=\textwidth]{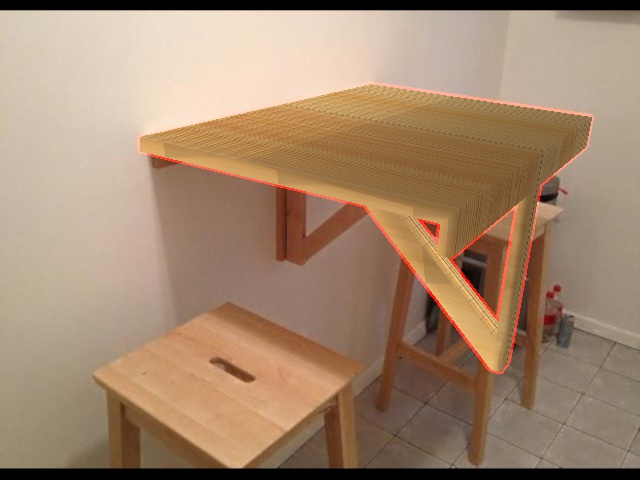}
        \end{minipage}\\[1mm]
        
        \small{2}
        \begin{minipage}{0.225\textwidth}
            \centering
            \includegraphics[width=\textwidth]{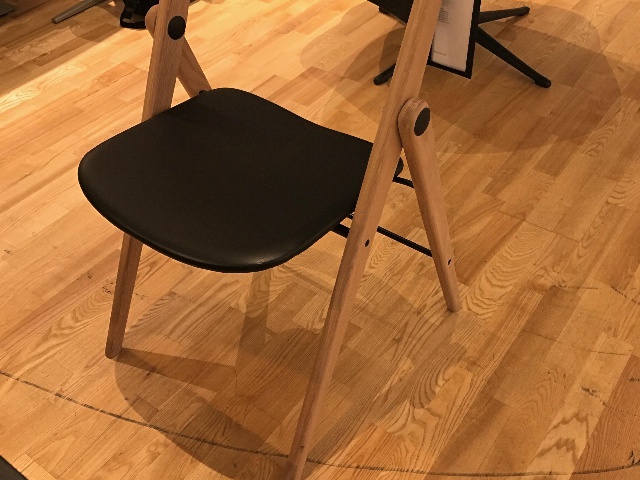}
        \end{minipage}
        \begin{minipage}{0.225\textwidth}
            \centering
            \includegraphics[width=\textwidth]{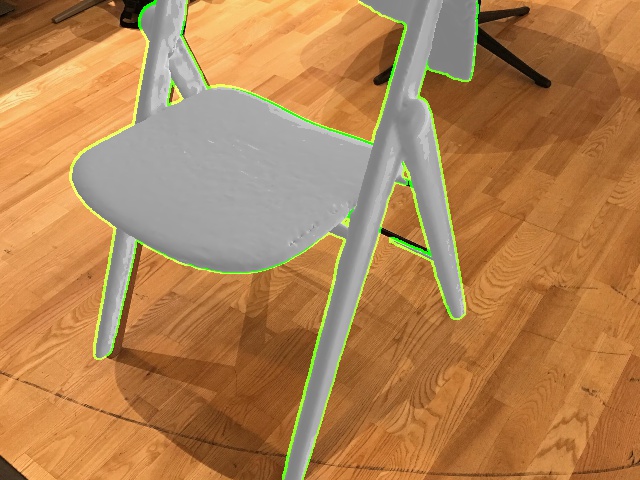}
        \end{minipage}
        \begin{minipage}{0.225\textwidth}
            \centering
            \includegraphics[width=\textwidth]{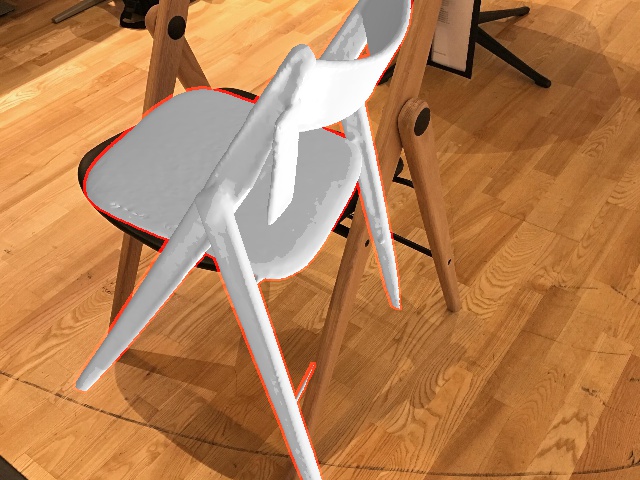}
        \end{minipage}\\[1mm]
        \small{3}
        \begin{minipage}{0.225\textwidth}
        \centering
            \includegraphics[width=\textwidth]{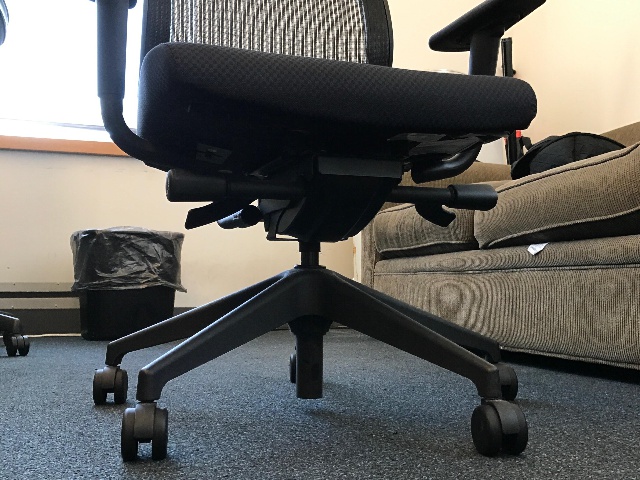}
        \end{minipage}
        \begin{minipage}{0.225\textwidth}
            \centering
            \includegraphics[width=\textwidth]{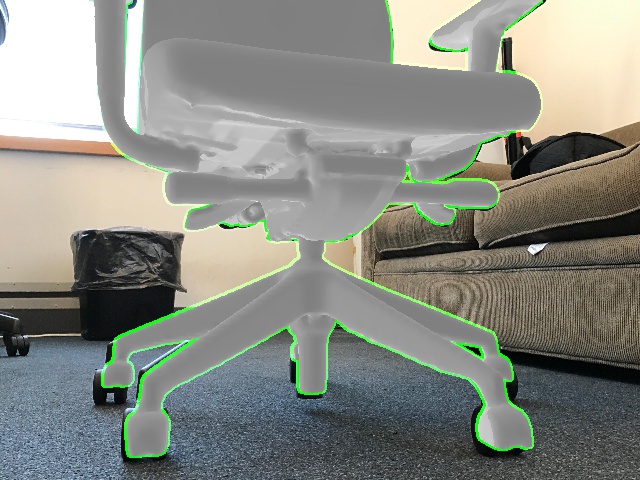}
        \end{minipage}
        \begin{minipage}{0.225\textwidth}
            \centering
            \includegraphics[width=\textwidth]{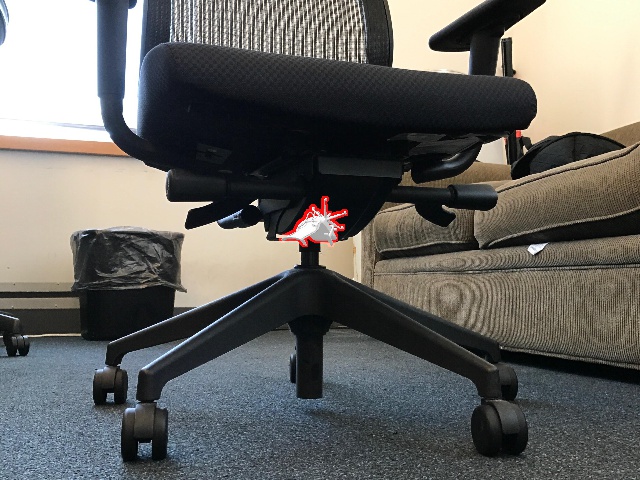}
        \end{minipage}\\[1mm]
        \small{4}
        \begin{minipage}{0.225\textwidth}
        \centering
            \includegraphics[width=\textwidth]{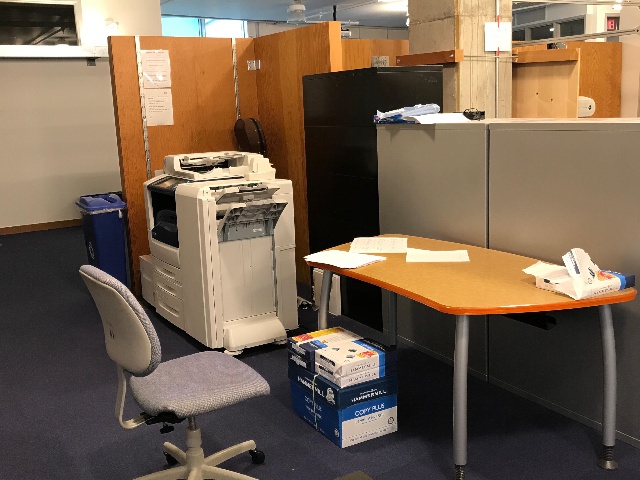}
        \end{minipage}
        \begin{minipage}{0.225\textwidth}
            \centering
            \includegraphics[width=\textwidth]{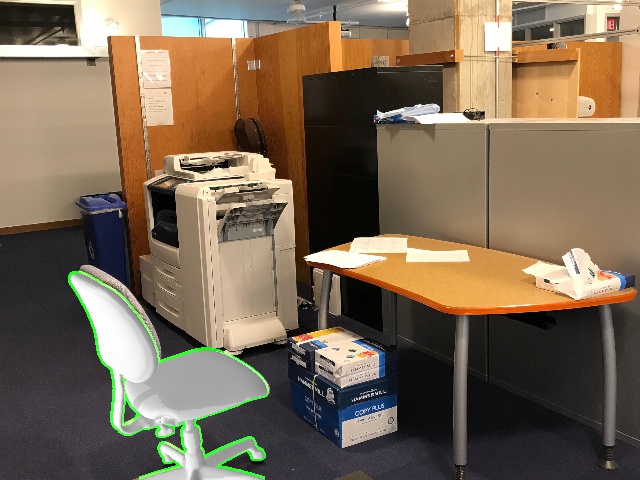}
        \end{minipage}
        \begin{minipage}{0.225\textwidth}
            \centering
            \includegraphics[width=\textwidth]{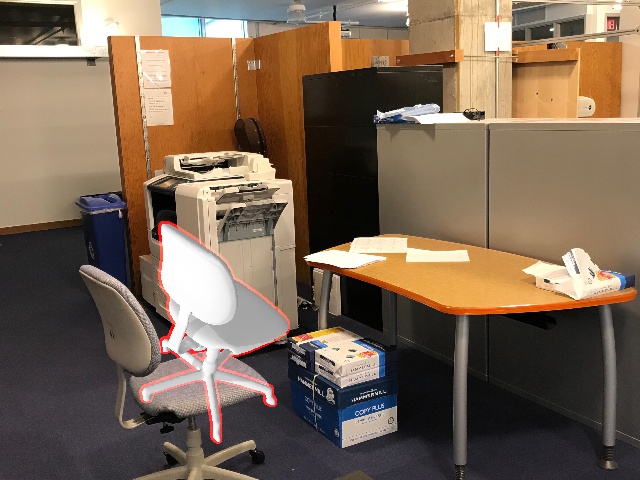}
        \end{minipage}\\[1mm]
        \small{5}
        \begin{minipage}{0.225\textwidth}
            \centering
            \includegraphics[width=\textwidth]{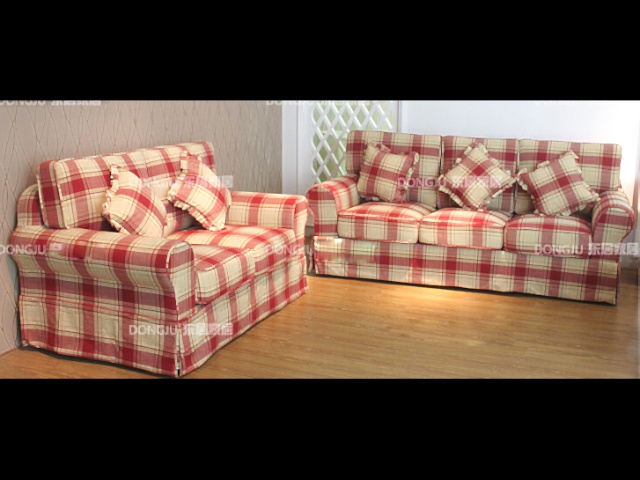}
        \end{minipage}
        \begin{minipage}{0.225\textwidth}
            \centering
            \includegraphics[width=\textwidth]{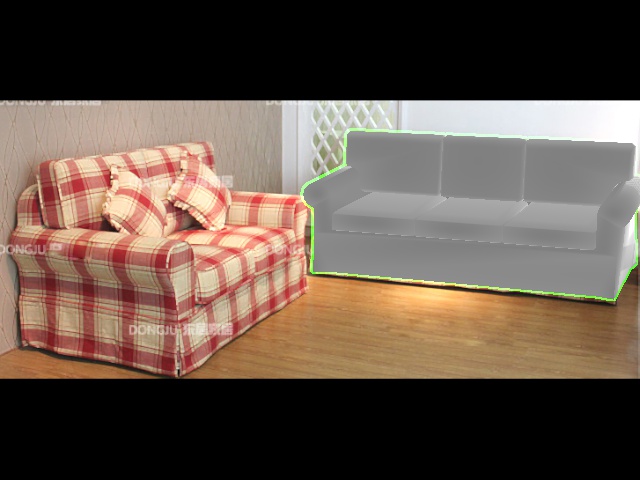}
        \end{minipage}
        \begin{minipage}{0.225\textwidth}
            \centering
            \includegraphics[width=\textwidth]{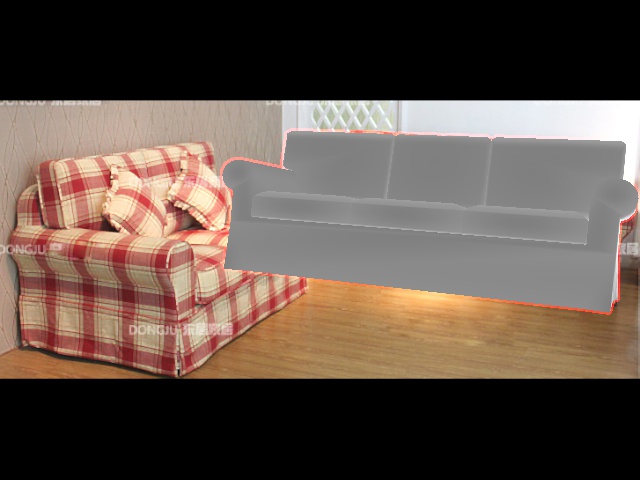}
        \end{minipage}\\[1mm]
        \small{6}
         \begin{minipage}{0.225\textwidth}
            \centering
            \includegraphics[width=\textwidth]{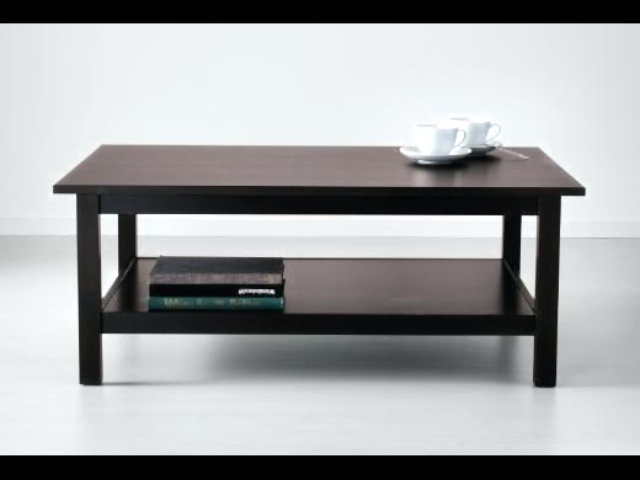}
        \end{minipage}
        \begin{minipage}{0.225\textwidth}
            \centering
            \includegraphics[width=\textwidth]{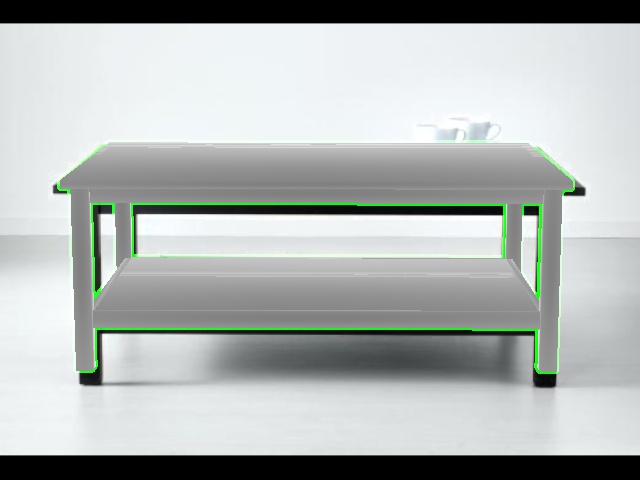}
        \end{minipage}
        \begin{minipage}{0.225\textwidth}
            \centering
            \includegraphics[width=\textwidth]{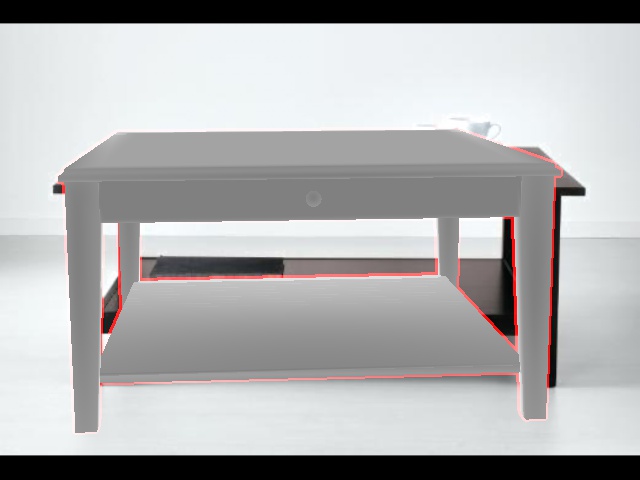}
        \end{minipage}\\[1mm]
        \small{7}
         \begin{minipage}{0.225\textwidth}
            \centering
            \includegraphics[width=\textwidth]{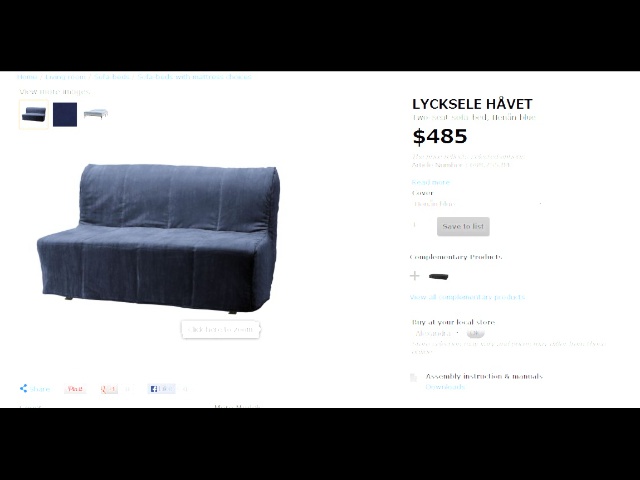}
        \end{minipage}
        \begin{minipage}{0.225\textwidth}
            \centering
            \includegraphics[width=\textwidth]{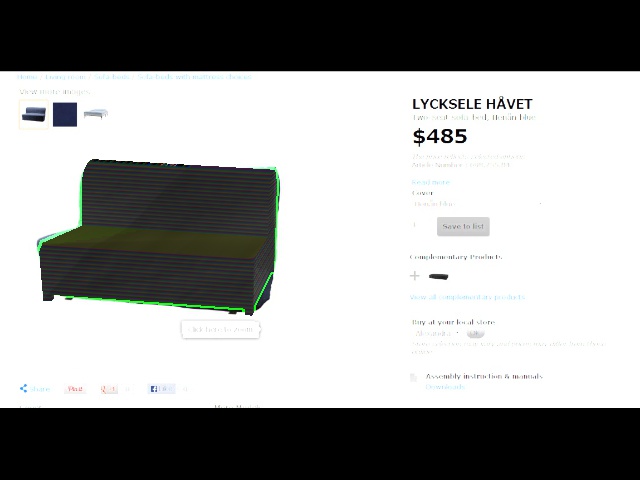}
        \end{minipage}
        \begin{minipage}{0.225\textwidth}
            \centering
            \includegraphics[width=\textwidth]{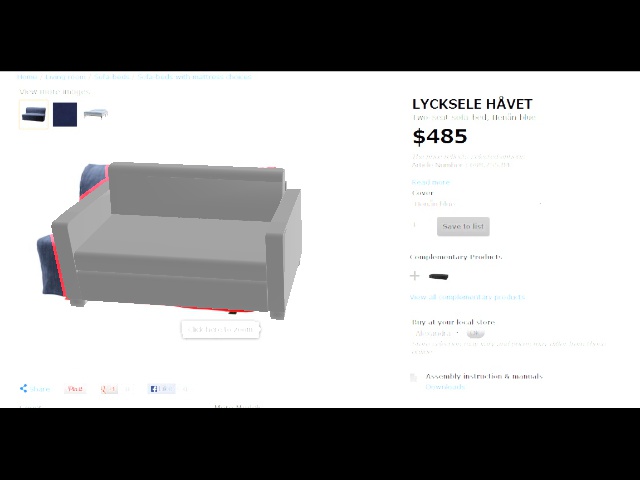}
        \end{minipage}\\[1mm]
        
        \caption{Examples of failures in the Pix3D dataset. Typical failures include symmetric objects (rows 1-2), local minima (rows 3-5) and misalignment due to the incorrect model (row 6-7). For more details please see Sec.~\textcolor{red}{5} and Fig.~\textcolor{red}{4} in the main paper.}
        \label{pix3d-q-fail}
\end{figure*}

In this section, we provide more qualitative results of our approach. 
Figures~\ref{pix3d-chair-q-1}--\ref{pix3d-tables-q-2} show additional results for the chair, bed, sofa, and table classes in the Pix3D dataset. Figures~\ref{compcars-q} and~\ref{stanford-q} show additional results for the Stanford cars and CompCars datasets, respectively. The qualitative results demonstrate the high accuracy of the alignments obtained by our approach despite variation in focal length, variability of the 3D models that have often very little texture, occlusions, and cluttered backgrounds.  Finally, Figure~\ref{pix3d-q-fail} shows additional examples of failure modes on the Pix3D dataset. 

For Pix3D, we provide good results for the chair class in Fig.~\ref{pix3d-chair-q-1} and Fig.~\ref{pix3d-chair-q-2}, for the bed class in Fig.~\ref{pix3d-beds-q-1} and Fig.~\ref{pix3d-beds-q-2}, for the sofa class in Fig.~\ref{pix3d-sofas-q-1} and Fig.~\ref{pix3d-sofas-q-2} and for the table class in Fig.~\ref{pix3d-tables-q-1} and Fig.~\ref{pix3d-tables-q-2}. We also provide qualitative results for Stanford cars in Fig.~\ref{compcars-q} and for CompCars in Fig.~\ref{stanford-q}. Please notice the quality of alignment that our approach can achieve.
We provide the failure cases for the Pix3D dataset in Fig.~\ref{pix3d-q-fail}.

\label{qualitative-results}

\end{document}